\documentclass[format=acmsmall, review=false, screen=true]{acmart}
\renewcommand\footnotetextcopyrightpermission[1]{}

\usepackage{xcolor}
\usepackage{comment}
\usepackage{caption}
\usepackage{url}
\usepackage{amsmath}
\usepackage{bm}
\usepackage{amsfonts}
\usepackage{enumitem}
\usepackage{todonotes}
\usepackage[final]{changes}
\usepackage{booktabs} 
\usepackage{adjustbox}
\usepackage{graphicx}
\usepackage{multirow}
\usepackage{subcaption}

\begin{document}
\sloppy

\title{Outcome-Oriented Predictive Process Monitoring: Review and Benchmark}

\author{Irene Teinemaa}
\affiliation{%
  \institution{University of Tartu}
  \city{Tartu}
  \country{Estonia}
  }
\email{irheta@ut.ee}

\author{Marlon Dumas}
\affiliation{%
  \institution{University of Tartu}
  \city{Tartu}
  \country{Estonia}
  }
\email{marlon.dumas@ut.ee}

\author{Marcello La Rosa}
\affiliation{%
 \institution{The University of Melbourne}
 \city{Melbourne}
 \country{Australia}
 }
\email{marcello.larosa@unimelb.edu.au}

\author{Fabrizio Maria Maggi}
\affiliation{%
  \institution{University of Tartu}
  \city{Tartu}
  \country{Estonia}
  }
\email{f.m.maggi@ut.ee}

\begin{abstract}
Predictive business process monitoring refers to the act of making predictions about the future state of ongoing cases of a business process, based on their incomplete execution traces and logs of historical (completed) traces.
Motivated by the increasingly pervasive availability of fine-grained event data about business process executions, the problem of predictive process monitoring has received substantial attention in the past years.
In particular, a considerable number of methods have been put forward to address the problem of outcome-oriented predictive process monitoring, which refers to classifying each ongoing case of a process according to a given set of possible categorical outcomes -- e.g., Will the customer complain or not? Will an order be delivered, canceled or withdrawn?
Unfortunately, different authors have used different datasets, experimental settings, evaluation measures and baselines to assess their proposals, resulting in poor comparability and an unclear picture of the relative merits and applicability of different methods.
To address this gap, this article presents a systematic review and taxonomy of outcome-oriented predictive process monitoring methods, and a comparative experimental evaluation of eleven representative methods using a benchmark covering 24~predictive process monitoring tasks based on nine real-life event logs.
\end{abstract}

%
%

%
%



\maketitle
\thispagestyle{empty}

\renewcommand{\shortauthors}{I. Teinemaa et al.}

\section{Introduction}
\label{sec:intro}



\added{A business process is a collection of inter-related events, activities, and decision points that involve a number of actors and objects, which collectively lead to an outcome that is of value to  a customer~\cite{FBPM}. 
A typical example is an order-to-cash process: a process that starts when a purchase order is received and ends when the product/service is delivered and the payment is confirmed.
An execution of a business process is called a \emph{case}. In an order-to-cash process, each purchase order gives rise to a case.}

Business process monitoring is the act of analyzing events produced by the executions of a business process at runtime, in order to understand its performance and its conformance with respect to a set of business goals~\cite{FBPM}. Traditional process monitoring techniques provide dashboards and reports showing the recent performance of a business process in terms of key performance indicators such as mean execution time, resource utilization or error rate with respect to a given notion of error.


Predictive (business) process monitoring techniques go beyond traditional ones by making predictions about the future state of the executions of a business process (\replaced{i.e. the cases}{herein called \emph{cases}}). For example, a predictive monitoring technique may seek to predict the remaining execution time of each ongoing case of a process~\cite{Rogge-SoltiW13}, the next activity that will be executed in each case~\cite{EvermannRF16}, or the final outcome of a case, with respect to a possible set of business outcomes~\cite{MetzgerFE12,maggi2014predictive,MetzgerLISFCDP15}. For instance, in an order-to-cash process \deleted{(a process going from the receipt of a purchase order to the receipt of payment of the corresponding invoice)}, the possible outcomes of a case may be that the purchase order is closed satisfactorily (i.e., the customer accepted the products and paid) or unsatisfactorily (e.g., the order was canceled or withdrawn).  Another set of possible outcomes is that the products were delivered on time (with respect to a maximum acceptable delivery time), or delivered late.



Recent years have seen the emergence of a rich field of proposed methods for predictive process monitoring in general, and predictive monitoring of (categorical) case outcomes in particular -- herein called \emph{outcome-oriented predictive process monitoring}.
Unfortunately, there is no unified approach to evaluate these methods. Indeed, different authors have used different datasets, experimental settings, evaluation measures and baselines.

This paper aims at filling this gap by (i) performing a systematic literature review of outcome-oriented predictive process monitoring methods; (ii) providing a taxonomy of existing methods; and (iii) performing a comparative experimental evaluation of eleven representative methods, using a benchmark of 24~predictive monitoring tasks based on nine real-life event logs.

The contribution of this study is a categorized collection of outcome-oriented predictive process monitoring methods and a benchmark designed to enable researchers to empirically compare new methods against existing ones in a unified setting. The benchmark is provided as an open-source framework that allows researchers to run the entire benchmark with minimal effort, and to configure and extend it with additional methods and datasets.

The rest of the paper is structured as follows. Section \ref{sec:bckg} introduces some basic concepts and definitions. Section \ref{sec:search_methodology} describes the search and selection of relevant studies. Section \ref{sec:methods} surveys the selected studies and provides a taxonomy to classify them. 
Section \ref{sec:evaluation} reports on benchmark evaluation of the selected studies while Section \ref{sec:threats} discusses threats to validity. Finally, Section \ref{sec:conclusion} summarizes the findings and outlines directions for future work.


\section{Background}
\label{sec:bckg}

The starting point of predictive process monitoring are \emph{event records} representing the execution of activities in a business process. An event record has a number of attributes. Three of these attributes are present in every event record, namely the \emph{event class} (a.k.a.\ \emph{activity name}) specifying which activity the event refers to, the \emph{timestamp} specifying when did the event occur, and the \emph{case id} indicating which case of the process generated this event. \added{For example, in an order-to-cash process, the purchase order identifier is the case id, since every event occurring in an execution of this process is associated with a purchase order.} In other words, every event represents the occurrence of an activity at a particular point in time and in the context of a given case. An event record may carry additional attributes in its payload. 
These are called \emph{event-specific attributes} (or \emph{event attributes} for short). For example, in an order-to-cash process, the amount of the invoice may be recorded as an attribute of an event referring to activity ``Create invoice''. Other attributes, namely \emph{case attributes}, belong to the case and are hence shared by all events generated by the same case. For example in an order-to-cash process, the customer identifier is likely to be a case attribute. If so, this attribute will appear in every event of every case of the order-to-cash process, and it has the same value for all events generated by a given case. In other words, the value of a case attribute is static, i.e., it does not change throughout the lifetime of a case, as opposed to attributes in the event payload, which are dynamic as they change from an event to the other.

Formally, an event record is defined as follows:
\begin{definition}[Event] An $event$ is a tuple $(a, c, t, (d_1, v_1), \ldots, (d_m, v_m))$ where $a$ is the activity name, $c$ is the case id, $t$ is the timestamp and $(d_1, v_1) \ldots, (d_m, v_m)$ (where $m \geq 0$) are the event or case attributes and their values.
\end{definition}

Herein, we use the term \emph{event} as a shorthand for \emph{event record}. The universe of all events is hereby denoted by $\mathcal{E}$.

The sequence of events generated by a given case forms a \emph{trace}. Formally:
\begin{definition}[Trace] A $trace$ is a non-empty sequence $\sigma = [e_1,\ldots,e_{n}]$ of events such that $\forall i \in [1..n], e_i \in \mathcal{E}$ and $\forall i,j \in [1..n] \; e_i{.}c = e_j{.}c$. In other words, all events in the trace refer to the same case.
\end{definition}

The universe of all possible traces is denoted by $\mathcal{S}$.

A set of \emph{completed traces} (i.e., traces recording the execution of completed cases) is called an \emph{event log}.

As a running example, we consider a simple log of a patient treatment process containing two cases (cf.\ \figurename~\ref{toyexample}). The activity name of the first event in trace $\sigma_1$ is \emph{consultation}, it refers to case $1$ and occurred at \emph{10:30AM}. The additional event attributes show that the cost of the procedure was $10$ and the activity was performed in the \emph{radiotherapy} department. These two are event attributes. Note that not all events carry every possible event attribute. For example, the first event of trace $\sigma_2$ does not have the attribute \emph{amountPaid}. In other words, the set of event attributes can differ from one event to another even within the same trace. The events in each trace also carry two case attributes: the age of the patient and the gender. The latter attributes have the same value for all events of a trace.
		\begin{figure}
			\scriptsize			
			\centering
\parbox{\textwidth}{
			\begin{tabbing}
		$\sigma_1 = $ \= [(consultation, 1, 10:30AM, (age, 33), (gender, female), (amountPaid, 10), (department, radiotherapy)), \ldots,\\
		                       \> (ultrasound, 1, 10:55AM, (age, 33), (gender, female),  (amountPaid, 15), (department, NursingWard))] \\
		$\sigma_2 = $ \= [(order blood, 2, 12:30PM, (age, 56), (gender, male), (department, GeneralLab), \ldots, \\
		                       \> (payment, 2, 2:30PM, (age, 56), (gender, male), (amountPaid, 100), (deparment, FinancialDept))]
		                       \end{tabbing}}
		\caption{Extract of an event log.}
		\label{toyexample}
	\end{figure}

An event or a case attribute can be of numeric, categorical, or of textual data type. Each data type requires different preprocessing to be usable by the classifier. With respect to the running example, possible event and case attributes and their type are presented in Table \ref{table:data_attribute_types}.

\begin{table}[hbtp]
\caption{Data attributes in the event log}
\label{table:data_attribute_types}
\begin{center}
\begin{tabular}{@{}ll@{}}
\toprule
Type & Example \\
\midrule
Case (static) \\
 \quad categorical & patient's gender \\
 \quad numeric & patient's age \\
 \quad textual & description of the application \\
Event (dynamic) \\
 \quad categorical & activity, resource \\
 \quad numeric & amount paid \\
 \quad textual & patient's medical history \\
\bottomrule
\end{tabular}
\end{center}
\end{table}



In predictive process monitoring, we aim at making predictions for traces of incomplete cases, rather than for traces of completed cases. Therefore, we make use of a function that returns the first $l$ events of a trace of a (completed) case.

\begin{definition}[Prefix function]
Given a trace $\sigma =  [e_1, \ldots, e_n]$ and a positive integer $l \leq n$, $\mathit{prefix}(\sigma, l) = [e_1, \ldots, e_l]$.
\end{definition}

Given a trace, outcome-oriented predictive process monitoring aims at predicting its \emph{class label} (expressing its outcome according to some business goal), given a set of completed cases with their known class labels.

\begin{definition}[Labeling function] A labeling function $y: \mathcal{S} \rightarrow \mathcal{Y}$ is a function that maps a trace $\sigma$ to its class label $y(\sigma) \in \mathcal{Y}$ with $\mathcal{Y}$ being
 the domain of the class labels. For outcome predictions, $\mathcal{Y}$ is a finite set of categorical outcomes. For example, for a binary outcome $\mathcal{Y} = \{0,1\}$.
\end{definition}

Predictions are made using a \emph{classifier} that takes as input a fixed number of independent variables (herein called \emph{features}) and learns a function to estimate the dependent variable (class label). This means that in order to use the data in an event log as input of a classifier, each trace in the log must be \emph{encoded} as a feature vector.

\begin{definition}[Sequence/trace encoder]
A sequence (or trace) encoder $f: \mathcal{S} \rightarrow \mathcal{X}_1 \times \cdots \times \mathcal{X}_p$ is a function that takes a (partial) trace $\sigma$ and transforms it into a feature vector in the $p$-dimensional vector space $\mathcal{X}_1 \times \cdots \times \mathcal{X}_p$ with $\mathcal{X}_j \subseteq \mathbb{R}, 1 \leq j \leq p$ being the domain of the $j$-th feature.
\end{definition}

The features extracted from a trace may encode information on activities performed during the execution of a trace and their order (herein called \emph{control-flow} features), and features that correspond to event/case attributes (herein referred to as \emph{data payload} features).

A classifier is a function that assigns a class label to a feature vector.
\begin{definition}[Classifier] A classifier $cls: \mathcal{X}_1 \times \cdots \times \mathcal{X}_p \rightarrow \mathcal{Y}$ is a function that takes an encoded $p$-dimensional sequence and estimates its class label.
\end{definition}

The construction of a classifier (a.k.a.\ classifier \emph{training}) for outcome-oriented predictive process monitoring is achieved by applying a classification algorithm over a set of prefixes of an event log. Accordingly, given a log $L$, we define its \emph{prefix log} log $L^*$ to be the event log that contains all prefixes of $L$, i.e., $L^* = \{prefix(\sigma,l) : \sigma \in L, 1 \leq l \leq |\sigma| \}$. Since the main aim of predictive process monitoring is to make predictions as early as possible (rather than when a case is about to complete), we often focus on the subset of the prefix log containing traces of up to a given length. Accordingly, we define the \emph{length-filtered prefix log} $L_k^*$ to be the subset of $L^*$ containing only prefixes of size less than or equal to $k$.

With respect to the broader literature on machine learning, we note that predictive process monitoring corresponds to a problem of \emph{early sequence classification}. In other words, given a set of labeled sequences, the goal is to build a model that for a sequence prefix predicts the label this prefix will get when completed. A survey on sequence classification presented in~\cite{XingP10} provides an overview of techniques in this field. This latter survey noted that, while there is substantial literature on the problem of sequence classification for simple symbolic sequences (e.g., sequences of events without payloads), there is a lack of proposals addressing the problem for complex symbolic sequences (i.e., sequences of events with payloads). The problem of outcome-oriented predictive process monitoring can be seen as an early classification over complex sequences where each element has a timestamp, a discrete attribute referring to an activity, and a payload made of a heterogeneous set of other attributes.


\section{Search methodology}
\label{sec:search_methodology}
In order to retrieve and select studies for our survey and benchmark, we conducted a \emph{Systematic Literature Review} (SLR) according to the approach described in \cite{kitchenham2004procedures}. We started by specifying the research questions. Next, guided by these goals, we developed relevant search strings for querying a database of academic papers. We applied inclusion and exclusion criteria to the retrieved studies in order to filter out irrelevant ones, and last, we divided all relevant studies into primary and subsumed ones based on their contribution.

\subsection{Research questions}
\label{sec:research_questions}
The purpose of this survey is to define a taxonomy of methods for \emph{outcome-oriented predictive monitoring of business processes}. The decision to focus on outcome-oriented predictive monitoring is to have a well-delimited and manageable scope, given the richness of the literature in the broader field of predictive process monitoring, and the fact that other predictive process monitoring tasks rely on entirely different techniques and evaluation measures.

In line with the selected scope, the survey focuses specifically on the following research question:

\begin{enumerate}[label=RQ0]
\item Given an event log of completed business process execution cases and the final outcome (class) of each case, how to train a model that can accurately and efficiently predict the outcome of an incomplete (partial) trace, based on the given prefix only?
\end{enumerate}

We then decomposed this overarching question into the following subquestions:
\begin{enumerate}[label=RQ\arabic*]
\item What methods exist for predictive outcome-oriented monitoring of business processes?
\item How to categorize these methods in a taxonomy?
\item What is the relative performance of these methods?
\end{enumerate}

In the following subsections, we describe our approach to identifying existing methods for predictive outcome-oriented process monitoring (RQ1). Subsequent sections address the other two research questions.

\subsection{Study retrieval}

First, we came up with relevant keywords according to the research question of predictive outcome-oriented process monitoring (RQ1) and our knowledge of the subject. We considered the following keywords relevant:

\begin{itemize}
\item ``(business) process'' --- a relevant study must take as input an event log of business process execution data;
\item ``monitoring'' --- a relevant study should concern run-time monitoring of business processes, i.e., work with partial (running) traces;
\item ``prediction'' --- a relevant study needs to estimate what will happen in the future, rather than monitor what has already happened.
\end{itemize}

We deliberately left out ``outcome'' from the set of keywords. The reason for this is that we presumed that different authors might use different words to refer to this prediction target. Therefore, in order to obtain a more exhaustive set of relevant papers, we decided to filter out studies that focus on other prediction targets (rather than the final outcome) in an a-posteriori filtering phase.

Based on these selected keywords, we constructed three search phrases: ``predictive process monitoring'', ``predictive business process monitoring'', and ``business process prediction''. We applied these search strings to the Google Scholar academic database and retrieved all studies that contained at least one of the phrases in the title, keywords, abstract, or the full text of the paper. We used Google Scholar, a well-known electronic literature database, as it encompasses all relevant databases such as ACM Digital Library and IEEE Xplore, and also allows searching within the full text of a paper.

The search was conducted in August 2017 and returned 93 papers, excluding duplicates.

\subsection{Study selection}

All the retrieved studies were matched against several inclusion and exclusion criteria to further determine their relevance to predictive outcome-oriented process monitoring. In order to be considered relevant, a study must satisfy all of the inclusion criteria and none of the exclusion criteria.

The assessment of each study was performed independently by two authors of this paper, and the results were compared to resolve inconsistencies with the mediation of a third author.

\subsubsection{Inclusion criteria}

The inclusion criteria are designed for assessing the relevance of studies in a superficial basis. Namely, these criteria are checked without working through the full text of the paper. The following inclusion criteria were applied to the retrieved studies:

\begin{enumerate}[label=IN\arabic*]
\item The study is concerned with predictions in the context of business processes (this criterion was assessed by reading title and abstract). 
\item The study is cited at least five times.
\end{enumerate}

The application of these inclusion criteria to the original set of retrieved papers resulted in eight relevant studies. We proceeded with one-hop-snowballing, i.e., we retrieved the papers that are related to (cite or are cited by) these eight studies and applied the same inclusion criteria. This procedure resulted in 545 papers, of which we retained 70 unique papers after applying the inclusion criteria.\footnote{All retrieved papers that satisfy the inclusion criteria can be found at \url{http://bit.ly/2uspLRp}}

\subsubsection{Exclusion criteria}

The list of studies that passed the inclusion criteria were further assessed according to a number of exclusion criteria. Determining if the exclusion criteria are satisfied could require a deeper analysis of the study, e.g., examining the approach and/or results sections of the paper. The applied exclusion criteria are:

\begin{enumerate}[label=EX\arabic*]
\item The study does not actually propose a predictive process monitoring \emph{method}.
\item The study does not concern \emph{outcome}-oriented prediction.
\item The technique proposed in the study is tailored to a specific labeling function.
\item The study does not take an \emph{event log} as input. 
\end{enumerate}

The EX1 criterion excludes overview papers, as well as studies that, after a more thorough examination, turned out to be focusing on some research question other than predictive process monitoring. EX2 excludes studies where the prediction target is something other than the final outcome. Common examples of other prediction targets that are considered irrelevant to this study are remaining time and next activity prediction. Using EX3, we excluded studies that are not directly about classification, i.e., that do not follow a black-box prediction of the case class. For example, studies that predict deadline violations by means of setting a threshold on the predicted remaining time, rather than by directly classifying the case as likely to violate the deadline or not. The reason for excluding such studies is that, in essence, they predict a numeric value, and are thus not applicable for predicting an arbitrarily defined case outcome. EX4 concerns studies that propose methods that do not utilize at least the following essential parts of an event log: the case identifier, the timestamp and the event classes. For instance, we excluded methods that take as input numerical time series without considering the heterogeneity in the control flow (event classes). In particular, this is the case in manufacturing processes which are of linear nature (a process chain). The reason for excluding such studies is that the challenges when predicting for a set of cases of heterogenous length are different from those when predicting for linear processes. While methods designed for heterogenous processes are usually applicable to those of linear nature, it is not so vice versa. Moreover, the linear nature of a process makes it possible to apply other, more standard methods that may achieve better performance.

The application of the exclusion criteria resulted in 14 relevant studies out of the 70 studies selected in the previous step.

\subsection{Primary and subsumed studies}

Among the papers that successfully passed both the inclusion and exclusion criteria, we determined \emph{primary} studies that constitute an original contribution for the purposes of our benchmark, and \emph{subsumed} studies that are similar to one of the primary studies and do not provide a substantial contribution with respect to it.

Specifically, a study is considered subsumed if:
\begin{itemize}
\item there exists a more recent and/or more extensive version of the study from the same authors (e.g., a conference paper is subsumed by an extended journal version), or
\item it does not propose a substantial improvement/modification over a method that is documented in an earlier paper by other authors, or
\item the main contribution of the paper is a case study or a tool implementation, rather than the predictive process monitoring method itself, and the method is described and/or evaluated more extensively in a more recent study by other authors.
\end{itemize}

This procedure resulted in seven primary and seven subsumed studies, listed in Table \ref{table:primary_and_subsumed}. In the next section we present the primary studies in detail, and classify them using a taxonomy. 

\begin{table}[hbtp]
\caption{Primary and subsumed studies}
\label{table:primary_and_subsumed}
\begin{center}
\begin{tabular}{@{}ll@{}}
\toprule
Primary study & Subsumed studies \\
\midrule
de Leoni et al.~\cite{de2016general} & de Leoni et al.~\cite{de2014general} \\
Maggi et al.~\cite{maggi2014predictive} & \\
Lakshmanan et al.~\cite{lakshmanan2010predictive} & Conforti et al.~\cite{conforti2013supporting, conforti2015recommendation}\\
di Francescomarino et al.~\cite{di2016clustering} & \\
Leontjeva et al.~\cite{leontjeva2015complex} & van der Spoel et al.~\cite{van2012process} \\
Verenich et al.~\cite{verenich2015complex} & \\
Castellanos et al.~\cite{castellanos2005predictive} & Schwegmann et al.~\cite{schwegmann2013method, schwegmann2013precep}, Ghattas et al.\cite{ghattas2014improving} \\
\bottomrule
\end{tabular}
\end{center}
\end{table}

\section{Analysis and taxonomy}
\label{sec:methods}

\begin{figure}[!h]
\centering
\includegraphics[width=0.9\textwidth]{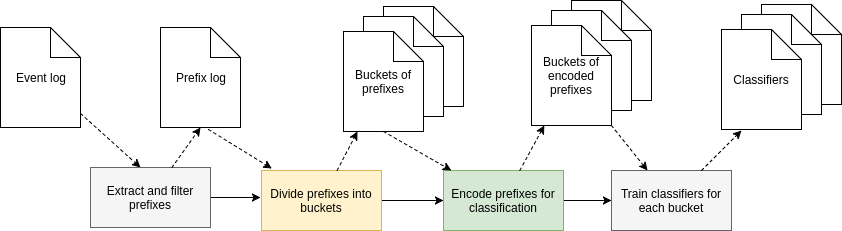}
\caption{predictive process monitoring workflow (offline phase)}
\label{fig:predictive_monitoring_benchmark_flow}
\end{figure}

In this section we present a taxonomy to classify the seven primary studies that we selected through our SLR. Effectively, with this section we aim at answering RQ1 (What methods exist?) and RQ2 (How to categorize them?) -- cf. Section \ref{sec:research_questions}. The taxonomy is framed upon a general workflow for predictive process monitoring, which we derived by studying all the methods surveyed. This workflow is divided into two phases: offline, to train a prediction model based on historical cases, and online, to make predictions on running process cases. The \emph{offline} phase, shown in Fig. \ref{fig:predictive_monitoring_benchmark_flow}, consists of four steps. First, given an event log, case prefixes are extracted and filtered (e.g., to retain only prefixes up to a certain length). Next, the identified prefixes are divided into buckets (e.g., based on process states or similarities among prefixes) and features are encoded from these buckets for classification. Finally, each bucket of encoded prefixes is used to train a classifier.


\begin{figure}[!h]
\centering
\includegraphics[width=0.9\textwidth]{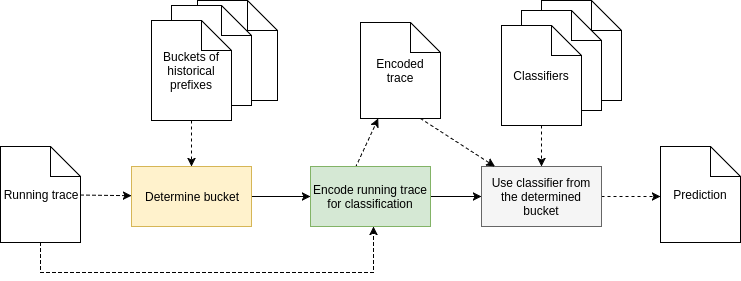}
\caption{predictive process monitoring workflow (online phase)}
\label{fig:predictive_monitoring_benchmark_flow_online}
\end{figure}

The \emph{online} phase, shown in Fig. \ref{fig:predictive_monitoring_benchmark_flow_online}, concerns the actual prediction for a running trace, by reusing the elements (buckets, classifiers) built in the offline phase. Specifically, given a running trace and a set of buckets of historical prefixes, the correct bucket is first determined. Next, this information is used to encode the features of the running trace for classification. In the last step, a prediction is extracted from the encoded trace using the correct classifier for the determined bucket.

We note that there is an exception among the surveyed methods that does not perfectly fit the presented workflow. Namely, the KNN approach proposed by Maggi et al.~\cite{maggi2014predictive} omits the offline phase. Instead, in this approach the bucket (a set of similar traces from the training set) is determined and a classifier is trained during the online phase, separately for each running case.

Table \ref{table:related_work_classification} lists the seven primary studies identified in our SLR, and shows their characteristics according to the four steps of the offline phase (prefix selection and filtering, trace bucketing, sequence encoding and classification algorithm). In the rest of this section we survey these studies based on these characteristics, and use this information to build a taxonomy that allows us to classify the studies.

\begin{table}[hbtp]
\caption{Classification of the seven primary studies according to the four steps of the offline phase.}
\label{table:related_work_classification}
\begin{center}
\begin{adjustbox}{max width=\textwidth}
\begin{tabular}{@{}lllllll@{}}
\toprule
 & Prefix extraction and & & \multicolumn{2}{c}{Sequence encoding} & \\ \cmidrule(r){4-5}
Primary study & filtering & Trace bucketing & Control flow & Data & Classification algorithm \\
\midrule
de Leoni et al.~\cite{de2016general} & all & Single & agg, last state & agg, last state  & DT \\
Maggi et al.~\cite{maggi2014predictive} & all & KNN & agg & last state & DT \\
Lakshmanan et al.~\cite{lakshmanan2010predictive} & all & State & last state & last state & DT \\
di Francescomarino et al.~\cite{di2016clustering} & prefix length 1-21,  & Cluster & agg & last state & DT, RF \\
 & with gap 3, 5, or 10 &  &  &  &  \\
Leontjeva et al.~\cite{leontjeva2015complex} & prefix length 2-20 & Prefix length & index & index & DT, RF, GBM, SVM \\
 &  &  & index & last state & RF \\
    &  &  & agg & - & RF \\
Verenich et al.~\cite{verenich2015complex} & prefix length 2-20 & Prefix length + cluster & index & index & RF \\
Castellanos et al.~\cite{castellanos2005predictive} &  all & Domain knowledge & \emph{unknown} & \emph{unknown} & DT  \\
\bottomrule
\end{tabular}
\end{adjustbox}
\end{center}
\end{table}%

\subsection{Prefix extraction and filtering}
\label{subsec:filtering}

After analyzing the identified studies, we found that all of them take as input a prefix log (as defined in Section~\ref{sec:bckg}) to train a classifier. This choice is natural given that at runtime, we need to make predictions for partial traces rather than completed ones. Using a prefix log for training ensures that our training data is comparable to the testing data. For example, for a complete trace consisting of a total of 5 events, we could consider up to 4 prefixes: the partial trace after executing the first event, the partial trace after executing the first and the second event, and so on.

Using all possible prefixes raises multiple problems. Firstly, the large number of prefixes as compared to the number of traces considerably slows down the training of the prediction models. Secondly, if the length of the original cases is very heterogenous, the longer traces produce much more prefixes than shorter ones and, therefore, the prediction model is biased towards the longer cases.
Accordingly, it is common to consider prefixes up to a certain number of events only. For example, Di Francescomarino et al.~\cite{di2016clustering} limit the maximum prefix length to 21, while Leontjeva et al.~\cite{leontjeva2015complex} use prefixes of up to 20 events only. In other words, in their training phase, these approaches take as input the length-filtered prefix log $L_k$ for $k=21$ and $k=20$.

Di Francescomarino et al.~\cite{di2016clustering} propose a second approach to filter the prefix log using so-called \emph{gaps}. Namely, instead of retaining all prefixes of up to a certain length, they retain prefixes whose length is equal to a base number (e.g., 1) plus a multiple of a gap (e.g., 1, 6, 11, 16, 21 for a gap of 5) . This approach helps to keep the prefix log sufficiently small for applications where efficiency of the calculations is a major concern.

We observe that length-based or gap-based filtering can be applied to any predictive process monitoring method. In other words, the choice of length or gap filtering is not an inherent property of a method.

\subsection{Trace bucketing}
Most of existing predictive process monitoring approaches train multiple classifiers rather than a single one. In particular, the prefix traces in the historical log are divided into several buckets and different classifiers are trained for each such buckets. At run-time, the most suitable bucket for the ongoing case is determined and the respective classifier is applied to make a prediction.
In the following, we describe the bucketing approaches that have been proposed by existing predictive process monitoring methods.

\subsubsection{Single bucket} All prefix traces are considered to be in the same bucket. A single classifier is trained on the whole prefix log and applied directly to the running cases. The single bucket approach has been used in the work by de Leoni et al.~\cite{de2016general}.

\subsubsection{KNN}
\label{subsec:knn}
In this bucketing approach, the offline training phase is skipped and the buckets are determined at run-time. Namely, for each running prefix trace, its $k$ nearest neighbors are selected from the historical prefix traces and a classifier is trained (at run-time) based on these $k$ neighbors. This means that the number of buckets (and classifiers) is not fixed, but grows with each executed event at run-time.

The KNN method for predictive process monitoring was proposed by Maggi et al.~\cite{maggi2014predictive}. Namely, they calculate the similarities between prefix traces using string-edit distance on the control flow. All instances that exceed a specified similarity threshold are considered as neighbors of the running trace. If the number of neighbors found is less than 30, the top 30 similar neighbors are selected regardless of the similarity threshold.

\subsubsection{State} In state-based approaches, a process model is derived from the event log. Then, relevant \emph{states} (or decision points) are determined from the process model and one classifier is trained for each such state. At run-time, the current state of the running case is determined, and the respective classifier is used to make a prediction for the running case.

Given an event log, Lakshmanan et al.~\cite{lakshmanan2010predictive} construct a so-called \emph{activity graph} where there is one node per possible activity (event class) in the log, and there is a directed edge from node $a_i$ to $a_j$ iff $a_j$ has occurred immediately after $a_i$ in at least one trace. This type of graph is also known as the \emph{Directly-Follows Graph} (DFG) of an event log~\cite{van2016process}. We observe that the DFG is the state-transition system obtained by mapping each trace prefix in the log to a state corresponding to the last activity appearing in the trace prefix (and hence the state of a running case is fully determined by its last activity). Alternative methods for constructing state abstractions are identified in~\cite{van2010process} (e.g., set-based, multiset-based and sequence-based state abstractions), but these have not been used for predictive process monitoring, and they are likely not to be suitable since they generate a very large number of states, which would lead to very large number of buckets. Most of these buckets would be too small to train a separate classifier.

In Lakshmanan et al.~\cite{lakshmanan2010predictive}, the edges in the DFG are annotated with transition probabilities. The transition probability from node $a_i$ to $a_j$ captures how often after performing activity $a_i$, $a_j$ is performed next. We observe that this DFG annotated with transition probabilities is a first order Markov chain.
For our purposes however, the transition probabilities are not necessary, as we aim to make a prediction for any running case regardless of its frequency. Therefore, in the rest of this paper, we will use the DFG without transition probabilities.


Lakshmanan et al.~\cite{lakshmanan2010predictive} build one classifier per \emph{decision point} --- i.e., per state in the model where the execution splits into multiple alternative branches. Given that in our problem setting, we need to be able to make a prediction for a running trace after each event, a natural extension to their approach is to build one classifier for every state in the process model.


\subsubsection{Clustering}
\label{subsec:clustering}
The cluster-based bucketer relaxes the requirement of a direct transition between the buckets of two subsequent prefixes. Conversely, the buckets (clusters) are determined by applying a clustering algorithm on the encoded prefix traces. This results in a number of clusters that do not exhibit any transitional structure. In other words, the buckets of $prefix(\sigma,l)$ and $prefix(\sigma,l+1)$ are determined independently from each other. Both of these prefixes might be assigned to the same cluster or different ones.

One classifier is trained per each resulting cluster, considering only the historical prefix traces that fall into that particular cluster. At run-time, the cluster of the running case is determined based on its similarity to each of the existing clusters and the respective classifier is applied.

A clustering-based approach is proposed by di Francescomarino et al.~\cite{di2016clustering}. They experiment with two clustering methods, DBScan (with string-edit distance) and model-based clustering (with Euclidean distance on the frequencies of performed activities), while neither achieves constantly superior performance over the other. Another clustering-based method is introduced by Verenich et al.~\cite{verenich2015complex}. In their approach, the prefixes are encoded using index-based encoding (see \ref{subsec:index}) using both control flow and data payload, and then either hierarchical agglomerative clustering (HAC) or k-medoids clustering is applied. According to their results, k-medoids clustering consistently outperforms HAC.


\subsubsection{Prefix length} In this approach, each bucket contains only the partial traces of a specific length. For example, one bucket contains traces where only the first event has been executed, another bucket contains those where first and second event have been executed, and so on. One classifier is built for each possible prefix length. The prefix length based bucketing was proposed by {Leontjeva et al.~\cite{leontjeva2015complex}. Also, Verenich et al.~\cite{verenich2015complex} bucket the prefixes according to prefix length before applying a clustering method.



\subsubsection{Domain knowledge} While the bucketing methods described so far can detect buckets through an automatic procedure, it is possible to define a bucketing function that is based on manually constructed rules. In such an approach, the input from a domain expert is needed. The resulting buckets can, for instance, refer to \emph{context categories}~\cite{ghattas2014improving} or \emph{execution stages}~\cite{castellanos2005predictive, schwegmann2013method}.

The aim of this survey and benchmark is to derive general principles by comparing methods that are applicable in arbitrary outcome-based predictive process monitoring scenarios and, thus, the methods that are based on domain knowledge about a particular dataset are left out of scope. For this reason, we do not further consider bucketing approaches based on domain knowledge.

\subsection{Sequence encoding}
In order to train a classifier, all prefix traces in the same bucket need to be represented as fixed length feature vectors. The main challenge here comes from the fact that with each executed event, additional information about the case becomes available, while each trace in a bucket (independent of the number of executed events) should still be represented with the same number of features.
This can be achieved by applying a trace abstraction technique~\cite{van2010process}, for example, considering only the last $m$ events of a trace. However, choosing an appropriate abstraction is a difficult task, where one needs to balance the trade-off between
the generality\footnote{Generality in this context means being able to apply the abstraction technique to as many prefix traces as possible; as an example, the last $m$ states abstraction is not meaningful for prefixes that are shorter than $m$ events.} and loss of information. After a trace abstraction is chosen, a set of feature extraction functions may be applied to each event data attribute of the abstracted trace. Therefore, a \emph{sequence encoding} method can be thought of as a combination of a trace abstraction technique and a set of feature extraction functions for each data attribute.

In the following subsections we describe the sequence encoding methods that have been used in the existing predictive process monitoring approaches. As described in Section \ref{sec:bckg}, a trace can contain any number of static case attributes and dynamic event attributes. Both the case and the event attributes can be of numeric, categorical, or textual type. As none of the compared methods deal with textual data, hereinafter we will focus on numeric and categorical attributes only.

\subsubsection{Static}

The encoding of case attributes is rather straightforward. As they remain the same throughout the whole case, they can simply be added to the feature vector ``as is'' without any loss of information. In order to represent all the information as a numeric vector, we assume the ``as is'' representation of a categorical attribute to be \emph{one hot encoding}. This means that each value of a categorical attribute is transformed into a bitvector $(v_1, \cdots, v_n)$, where $m$ is the number of possible levels of that attribute, $v_i = 1$ if the given value is equal to the $i$th level of the attribute, and $v_i = 0$ otherwise.

\subsubsection{Last state}

In this encoding method, only the last available snapshot of the data is used. Therefore, the size of the feature vector is proportional to the number of event attributes and is fixed throughout the execution of a case. A drawback of this approach is that it disregards all the information that has happened in the past, using only the very latest data snapshot. To alleviate this problem, this encoding can easily be extended to the last $m$ states, in which case the size of the feature vector increases $m$ times. As the size of the feature vector does not depend on the length of the trace, the last state (or, the last $m$ states) encoding can be used with buckets of traces of different lengths.

Using the last state abstraction, only one value (the last snapshot) of each data attribute is available. Therefore, no meaningful aggregation functions can be applied. Similarly to the static encoding, the numeric attributes are added to the feature vector ``as is'', while one hot encoding is applied to each categorical attribute.

Last state encoding is the most common encoding technique, having been used in the KNN approach~\cite{maggi2014predictive}, state-based bucketing~\cite{lakshmanan2010predictive}, as well as the clustering-based bucketing approach by Di Francescomarino et al.~\cite{di2016clustering}. Furthermore, De Leoni et al.~\cite{de2016general} mention the possibility of using the last and the previous (the last two) states.

\subsubsection{Aggregation}
\label{subsec:aggregation}

The last state encoding has obvious drawbacks in terms of information loss, neglecting all data that have been collected in the earlier stages of the trace. Another approach is to consider all events since the beginning of the case, but ignore the order of the events. This abstraction method paves the way to several aggregation functions that can be applied to the values that an event attribute has taken throughout the case.

In particular, the frequencies of performed activities (control flow) have been used in several existing works~\cite{leontjeva2015complex, di2016clustering}. Alternatively, boolean values have been used to express whether an activity has occurred in the trace. However, the frequency-based encoding has been shown to be superior to the boolean encoding~\cite{leontjeva2015complex}. For numerical attributes, De Leoni et al.~\cite{de2016general} proposed using general statistics, such as the average, maximum, minimum, and sum.


\subsubsection{Index}
\label{subsec:index}

While the aggregation encoding exploits information from all the performed events, it still exhibits information loss by neglecting the order of the events.
The idea of index-based encoding is to use all possible information (including the order) in the trace, generating one feature per each event attribute per each executed event (each \emph{index}). This way, a lossless encoding of the trace is achieved, which means that it is possible to completely recover the original trace based on its feature vector.
A drawback of index-based encoding is that due to the fact that the length of the feature vector increases with each executed event, this encoding can only be used in homogenous buckets where all traces have the same length.

Index-based encoding was proposed by Leontjeva et al.~\cite{leontjeva2015complex}. Additionally, in their work they combined the index-based encoding with HMM log-likelihood ratios. However, we decided not to experiment with HMMs in this study for mainly two reasons. Firstly, the HMMs did not consistently improve the basic index-based encoding in \cite{leontjeva2015complex}. Secondly, rather than being an essential part of index-based encoding, HMMs can be thought of as an aggregation function that can be applied to each event attribute, similarly to taking frequencies or numeric averages. Therefore, HMMs are not exclusive to index-based encoding, but could also be used in conjunction with the aggregation encoding. Index-based encoding is also used in the approach of Verenich et al.~\cite{verenich2015complex}.

\subsubsection*{Summary}
An overview of the encoding methods can be seen in Table \ref{table:encoding_types}. Note that the static encoding extracts different type of data from the trace (case attributes) than the other three methods (event attributes). Therefore, for obtaining a complete representation for a trace, it is reasonable to concatenate the static encoding with one of the other three encodings. In our experiments, the static encoding is included in every method, e.g., the ``last state'' method in the experiments refers to the static encoding for case attributes concatenated with the last state encoding for event attributes.

\begin{table}[hbtp]
\caption{Encoding methods}
\label{table:encoding_types}
\begin{center}
\begin{adjustbox}{max width=\textwidth}
\begin{tabular}{@{}lrrrr@{}}
\toprule
Encoding & Relevant & Trace & \multicolumn{2}{c}{Feature extraction} \\ \cmidrule(r){4-5}
name & attributes & abstraction & Numeric & Categorical \\
\midrule
Static & Case & Case attributes & as is & one-hot \\
Last state & Event & Last event & as is & one-hot \\
Aggregation & Event & All events, unordered & min, max, mean,  & frequencies or \\
 &  & (set/bag) & sum, std & occurrences \\
Index & Event & All events, ordered & as is & one-hot  \\
 &  & (sequence) & for each index & for each index \\
 \bottomrule
\end{tabular}
\end{adjustbox}
\end{center}
\end{table}

\subsection{Classification algorithm}

The existing predictive process monitoring methods have experimented with different classification algorithms. The most popular choice has been decision tree (DT), which has obvious benefits in terms of the interpretability of the results. Another popular method has been random forest~\cite{breiman2001random} (RF), which usually achieves better prediction accuracy than a single decision tree, but is harder to interpret. Additionally, Leontjeva et al.~\cite{leontjeva2015complex} experimented with support vector machines (SVM) and generalized boosted regression models (GBM), but found that their performance is inferior to RF.
Recently, gradient boosted trees~\cite{friedman2001greedy} in conjunction with existing predictive process monitoring techniques have shown promising results, often outperforming RF~\cite{rozumnyi2017nirdizati,Senderovich2017}.

\subsection{Discussion}

We have observed that the prefix filtering techniques are not inherent to any given predictive process monitoring method. Instead, these techniques are selected based on performance considerations and can be used in conjunction with any of the predictive process monitoring methods.
In a similar vein, the choice of a classification algorithm is a general problem in machine learning and is not specific to business process data.  \replaced{Indeed}{In fact}, all of the authors of the methods reviewed above claim that their method is applicable in conjunction with any classifier. \added{Therefore, we treat the prefix filtering technique and the classification algorithm employed as \emph{orthogonal} aspects to the categorization of predictive process monitoring methods. However, while excluded from the taxonomy, the specific prefix filtering technique and classification algorithm used still play an important role in obtaining good predictions, with their performance being influenced by the particular settings used.} \deleted{Therefore, we consider neither the prefix filtering technique nor the classification algorithm employed to be relevant aspects when categorizing and comparing predictive process monitoring methods.}

The considered methods also differ in terms of the event log attributes that are used for making predictions. However, it has been shown~\cite{leontjeva2015complex} that including more information (i.e., combining control flow and data payload) can drastically increase the predictive power of the models.
In order to provide a fair comparison of the different methods, it is preferable to provide the same set of attributes as input to all methods, and preferably the largest possible set of attributes. Accordingly, in the comparative evaluation below, we will encode traces using all the available case and event attributes (covering both control flow and data payload).

Based on the above, we conclude that existing outcome-oriented predictive process monitoring methods can be compared on two grounds:
\begin{itemize}
\item how the prefix traces are divided into buckets (trace bucketing)?
\item how the (event) attributes are transformed into features (sequence encoding)?
\end{itemize}


Figure \ref{fig:predictive_monitoring_taxonomy} provides a taxonomy of the relevant methods based on these two perspectives. Note that although the taxonomy is based on 7 primary studies, it contains 11 different approaches. The reason for this is that while the primary approaches tend to mix different encoding schemes, for example, use aggregation encoding for control flow and last state encoding for data payload (see Table \ref{table:related_work_classification}), the taxonomy is constructed in a modular way, so that each encoding method constitutes a separate approach. In order to provide a fair comparison of different encoding schemes, we have decided to evaluate each encoding separately, while the same encoding is applied to both control flow and data payload. Still, the different encodings (that are valid for a given bucketing method) can be easily combined, if necessary.
Similarly, the taxonomy does not contain combinations of several bucketing methods. An example of such ``double bucket'' approaches is the method by Verenich et al.~\cite{verenich2015complex}, where the prefixes are first divided into buckets based on prefix length and, in turn, clustering is applied in each bucket. We believe that comparing the performance of each bucketing method separately (rather than as a combination) provides more insights about the benefits of each method. Furthermore, the double bucket approaches divide the prefixes into many small buckets, which often leads to situations where a classifier receives too little training instances to learn meaningful patterns.

We note that the taxonomy generalizes the state-of-the-art, in the sense that even if a valid pair of bucketing and encoding method has not been used in any existing approach in the literature, it is included in the taxonomy (e.g., the state-based bucketing approach with aggregation encoding).
We also note that while the taxonomy covers the techniques proposed in the literature, all these techniques rely on applying a propositional classifier on an explicit vectorial representation of the traces. One could envisage alternative approaches that do not require an explicit feature vector as input. For instance, kernel-based SVMs have been used in the related setting of predicting the cycle time of a case~\cite{van2008cycle}.
Furthermore, one could envisage the use of data mining techniques to extract additional features from the traces (e.g., latent variables or frequent patterns). Although the taxonomy does not cover this aspect explicitly, applying such techniques is consistent with the taxonomy, since the derived features can be used in combination with any of the sequence encoding and bucketing approaches presented here. While most of the existing works on outcome-oriented predictive monitoring use the event/trace attributes ``as-is'' without an additional mining step, Leontjeva et al. used Hidden Markov Models for extracting additional features in combination with index-based encoding~\cite{leontjeva2015complex}. Similarly, Teinemaa et al. applied different natural language processing techniques to extract features from textual data~\cite{teinemaa2016predictive}. Further on, although not yet applied to outcome-oriented predictive process monitoring tasks, different pattern mining techniques could be applied to extract useful patterns from the sequences, occurrences of which could then be used as features in the feature vectors of the traces. Such techniques have been used in the domain of early time series/sequence classification~\cite{xing2008mining,ghalwash2013extraction,ghalwash2012early,he2015early,lin2015reliable} and for predicting numerical measures (e.g., remaining time) for business processes~\cite{folino2014mining}.

\begin{figure}[hbtp]
\centering
\includegraphics[width=0.9\textwidth]{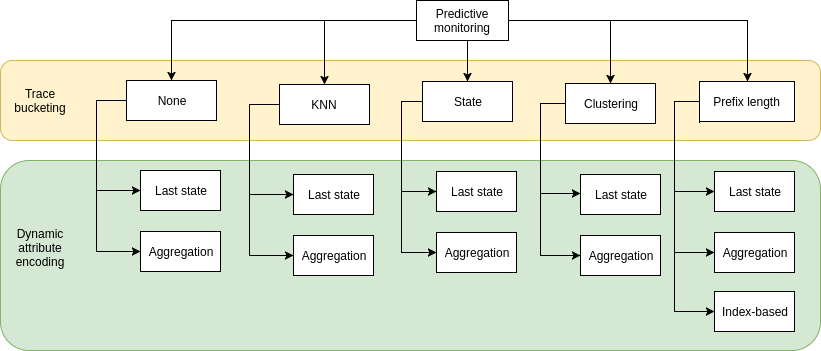}
\caption{Taxonomy of methods for predictive monitoring of business process outcome.}
\label{fig:predictive_monitoring_taxonomy}
\end{figure}

\newcommand{\lnext}{\ensuremath{\mathbf{X}}}
\newcommand{\lwnext}{\ensuremath{\mathbf{\bar{X}}}}
\newcommand{\luntil}{\ensuremath{\mathbf{U}}}
\newcommand{\lsince}{\ensuremath{\mathbf{S}}}
\newcommand{\lzero}{\ensuremath{\mathbf{Z}}}
\newcommand{\ltrigger}{\ensuremath{\mathbf{T}}}
\newcommand{\lrelease}{\ensuremath{\mathbf{R}}}
\newcommand{\lwuntil}{\ensuremath{\mathbf{W}}}
\newcommand{\lglobally}{\ensuremath{\mathbf{G}}}
\newcommand{\lfuture}{\ensuremath{\mathbf{F}}}
\newcommand{\tnext}{\ensuremath{\mathbf{X}_{[t_1,t_2]}}}
\newcommand{\twnext}{\ensuremath{\mathbf{\bar{X}_I}}}
\newcommand{\tuntil}{\ensuremath{\mathbf{U}_{[t_1,t_2]}}}
\newcommand{\tsince}{\ensuremath{\mathbf{S}_{[t_1,t_2]}}}
\newcommand{\trelease}{\ensuremath{\mathbf{R}_{[t_1,t_2]}}}
\newcommand{\tglobally}{\ensuremath{\mathbf{G}_{[t_1,t_2]}}}
\newcommand{\lonce}{\ensuremath{\mathbf{O}}}
\newcommand{\tonce}{\ensuremath{\mathbf{O}_{[t_1,t_2]}}}
\newcommand{\lyesterday}{\ensuremath{\mathbf{Y}}}
\newcommand{\tyesterday}{\ensuremath{\mathbf{Y}_{[t_1,t_2]}}}
\newcommand{\lhistorically}{\ensuremath{\mathbf{H}}}
\newcommand{\thistorically}{\ensuremath{\mathbf{H}_{[t_1,t_2]}}}
\newcommand{\tfuture}{\ensuremath{\mathbf{F}_{[t_1,t_2]}}}
\newcommand{\nfuture}{\ensuremath{\mathbf{F}_{[0,t_1]}}}
\newcommand{\true}{\ensuremath{\mbox{true}}}
\newcommand{\false}{\ensuremath{\mbox{false}}}
\newcommand{\n}{\ensuremath{\figitem{N}}}
\newcommand{\old}{\ensuremath{\figitem{O}}}
\newcommand{\R}{\ensuremath{\mathbf{R}_+}}

\section{Benchmark}
\label{sec:evaluation}

After conducting our survey, we proceeded with benchmarking the 11 approaches (shown in Figure \ref{fig:predictive_monitoring_taxonomy}) using different evaluation criteria (prediction accuracy, earliness and computation time), to address RQ3 (What is the relative performance of these methods?) -- cf.\ Section \ref{sec:research_questions}.

To perform our benchmark, we implemented an open-source, tunable and extensible predictive process monitoring framework in Python.\footnote{The code is available at \url{https://github.com/irhete/predictive-monitoring-benchmark}} All experiments were run using Python 3.6 and the scikit-learn library~\cite{scikit-learn} on a single core of a Intel(R) Xeon(R) CPU E5-2660 v2 @ 2.20GHz with 64GB of RAM.

In the rest of this section we first introduce the evaluation datasets, then describe the evaluation procedure and conclude with discussing the results of the experiments.

\subsection{Datasets}
\label{sec:datasets}

The benchmark is based on nine real-life event logs, out of which eight are publicly available and one is a private dataset. The public logs
are accessible from the 4TU Centre for Research Data.\footnote{\url{https://data.4tu.nl/repository/collection:event_logs_real}} The private log \emph{Insurance} originates from a claims handling process at an Australian insurance company.
The criterion for selecting the public event logs for the evaluation was that the log must contain both case attributes (static) and event attributes (dynamic). Based on this, we discarded the logs from years 2013-2014. We also discarded the BPIC 2016 dataset because it is a click-dataset of a Web service, rather than an event log of a business process.

In case of some logs, we applied several labeling functions $y$. In other words, the outcome of a case is defined in several ways depending on the goals and needs of the process owner. Each such notion of the outcome constitutes a separate predictive process monitoring task with slightly different input datasets. In total, we formulated a total of 24 different outcome prediction tasks based on the nine original event logs. In the following paragraphs we describe the original logs, the applied labeling functions, and the resulting predictive monitoring tasks in more detail.

{\bfseries BPIC2011.} This event log contains cases from the Gynaecology department of a Dutch Academic Hospital. Each case assembles the medical history of a given patient, where the applied procedures and treatments are recorded as activities. Similarly to previous work~\cite{leontjeva2015complex, di2016clustering}, we use four different labeling functions based on LTL rules~\cite{pnueli1977temporal}.
Specifically, we define the class label for a case $\sigma$ according to whether an LTL rule $\varphi$ is violated or satisfied by each trace $\sigma$.
\begin{equation*}
 y(\sigma) =
 \begin{cases}
   1  & \text{if } \varphi \text{ violated in } \sigma \\
   0  & \text{otherwise}
 \end{cases}
\end{equation*}
\tablename~\ref{operators} introduces the semantics of the LTL operators. 
\begin{table}[h!]
\caption{\footnotesize{\sc{LTL Operators Semantics}}}
\centering
\scalebox{0.7}{
\begin{tabular}{c | l}
\textbf{\textit{operator}}  & \textbf{\textit{semantics}} \\ \hline
$\lnext \varphi$ & \textit{$\varphi$ has to hold in the next position of a path.}\\ \hline
$\lglobally \varphi$ & \textit{$\varphi$ has to hold always in the subsequent positions of a path.}\\ \hline
$\lfuture \varphi$ & \textit{$\varphi$ has to hold eventually (somewhere) in the subsequent positions of a path.}\\ \hline
\multirow{2}{*}{$\varphi$ $\luntil\psi$} & \textit{$\varphi$ has to hold in a path at least until $\psi$ holds. $\psi$ must hold in the current or} \\
 & \textit{in a future position.}\\ \hline
\end{tabular}
}
\label{operators}
\end{table}

The four LTL rules used to formulate the four prediction tasks on the BPIC 2011 log are as follows:

\begin{itemize}
  \item \emph{bpic2011\_1}: $\varphi = \lfuture(``tumor~marker~CA-19.9") \vee \lfuture(``{ca-125}~using~meia")$,
  \item \emph{bpic2011\_2}: $\varphi = \\ \lglobally(``CEA-tumor~marker~using~meia" \rightarrow \lfuture(``squamous~cell~carcinoma~using~eia"))$,
  \item \emph{bpic2011\_3}: $\varphi = \\ (\neg``histological~examination-biopsies~nno") \luntil (``squamous~cell~carcinoma~using~eia")$, and
  \item \emph{bpic2011\_4}: $\varphi = \lfuture(``histological~examination-big~resectiep")$.
\end{itemize}

For example, the $\varphi$ for \emph{bpic2011\_1} expresses the rule that at least one of the activities ``tumor~marker~CA-19.9'' or ``{ca-125}~using~meia'' must happen eventually during a case. Evidently, the class label of a case becomes known and irreversible when one of these two events has been executed. In order to avoid bias introduced by this phenomenon during the evaluation phase, all the cases are cut exactly before either of these events happens. Similarly, the cases are cut before the occurrence of ``histological~examination-biopsies~nno'' in the \emph{bpic2011\_3} dataset and before ``histological~examination-big~resectiep'' in \emph{bpic2011\_4}. However, no cutting is performed in the \emph{bpic2011\_2} dataset, because the $\varphi$ states that a ``CEA-tumor~marker~using~meia'' event must \emph{always} be followed by a ``squamous~cell~carcinoma~using~eia'' event sometime in the future. Therefore, even if one occurrence of ``CEA-tumor~marker~using~meia'' has successfully been followed by a ``squamous~cell~carcinoma~using~eia'' ($\varphi$ is satisfied), another occurrence of ``CEA-tumor~marker~using~meia'' will cause the $\varphi$ to be violated again and, thus, the class label is not irreversibly known until the case completes.

{\bfseries BPIC2015.} This dataset assembles event logs from 5 Dutch municipalities, pertaining to the building permit application process. We treat the datasets from each municipality as separate event logs and apply a single labeling function to each one. Similarly to BPIC 2011, the labeling function is based on the satisfaction/violation of an LTL rule $\varphi$. The prediction tasks for each of the 5 municipalities are denoted as \emph{bpic2015\_i}, where $i = 1 \ldots 5$ indicates the number of the municipality. The LTL rule used in the labeling functions is as follows:

\begin{itemize}
\item \emph{bpic2015\_i}: $\varphi = \lglobally(``send~confirmation~receipt" \rightarrow \lfuture(``retrieve~missing~data")) $.
\end{itemize}

No trace cutting can be performed here, because, similarly to \emph{bpic2011\_2}, the  final satisfaction/violation of $\varphi$ is not known until the case completes.

{\bfseries Production.} This log contains data from a manufacturing process. Each trace records information about the activities, workers and/or machines involved in producing an item. The labeling (\emph{production}) is based on whether or not the number of rejected work orders is larger than zero.

{\bfseries Insurance.} This is the only private log we use in the experiments. It comprises of cases from an Australian insurance claims handling process. We apply two labeling functions:

\begin{itemize}
\item \emph{insurance\_1}: $y$ is based on whether a specific ``key'' activity is performed during the case or not.
\item \emph{insurance\_2}: $y$ is based on the time taken for handling the case, dividing them into slow and fast cases.
\end{itemize}

{\bfseries Sepsis cases.} This log records trajectories of patients with symptoms of the life-threatening sepsis condition in a Dutch hospital. Each case logs events since the patient's registration in the emergency room until her discharge from the hospital. Among others, laboratory tests together with their results are recorded as events. Moreover, the reason of the discharge is available in the data in an obfuscated format.

We created three different labelings for this log:

\begin{itemize}
\item \emph{sepsis\_1}: the patient returns to the emergency room within 28 days from the discharge,
\item \emph{sepsis\_2}: the patient is (eventually) admitted to intensive care,
\item \emph{sepsis\_3}: the patient is discharged from the hospital on the basis of something other than \emph{Release A} (i.e., the most common release type).
\end{itemize}

{\bfseries BPIC2012.} This dataset, originally published in relation to the Business Process Intelligence Challenge (BPIC) in 2012, contains the execution history of a loan application process in a Dutch financial institution. Each case in this log records the events related to a particular loan application. For classification purposes, we defined some labelings based on the final outcome of a case, i.e., whether the application is accepted, rejected, or canceled. Intuitively, this could be thought of as a multi-class classification problem. However, to remain consistent with previous work on outcome-oriented predictive process monitoring, we approach it as three separate binary classification tasks. In the experiments, these tasks are referred to as \emph{bpic2012\_1}, \emph{bpic2012\_2}, and \emph{bpic2012\_3}.

{\bfseries BPIC2017.} This event log originates from the same financial institution as the BPIC2012 one. However, the data collection has been improved, resulting in a richer and cleaner dataset. As in the previous case, the event log records execution traces of a loan application process. Similarly to BPIC2012, we define three separate labelings based on the outcome of the application, referred to as \emph{bpic2017\_1}, \emph{bpic2017\_2}, and \emph{bpic2017\_3}.

{\bfseries Hospital billing.} This dataset comes from an ERP system of a hospital. Each case is an execution of a billing procedure for medical services. We created two labelings for this log:

\begin{itemize}
\item \emph{hospital\_1}: the billing package not was eventually closed,
\item \emph{hospital\_2}: the case is reopened.
\end{itemize}

{\bfseries Traffic fines.} This log comes from an Italian local police force. The dataset contains events about notifications sent about a fine, as well as (partial) repayments. Additional information related to the case and to the individual events include, for instance, the reason, the total amount, and the amount of repayments for each fine. We created the labeling (\emph{traffic}) based on whether the fine is repaid in full or is sent for credit collection.

The resulting 24 datasets exhibit different characteristics which can be seen in Table \ref{table:dataset_stats}. The smallest log is \emph{production} which contains 220 cases, while the largest one is \emph{traffic} with 129615 cases. The most heterogenous in terms of case length are the \emph{bpic2011} labelled datasets, where the longest case consists of 1814 events. On the other hand, the most homogenous is the \emph{traffic} log, where case length varies from 2 to 20 events. The class labels are the most imbalanced in the \emph{hospital\_2} dataset, where only 5\% of cases are labeled as \emph{positive} ones (class label = 1). Conversely, in \emph{bpic2012\_1}, \emph{bpic2017\_3}, and \emph{traffic}, the classes are almost balanced. In terms of event classes, the most homogenous are the insurance datasets, with only 9 distinct activity classes. The most heterogenous are the \emph{bpic2015} datasets, reaching 396 event classes in \emph{bpic2015\_2}. The datasets also differ in terms of the number of static and dynamic attributes. The \emph{insurance} logs contain the most dynamic attributes (22), while the \emph{sepsis} datasets contain the largest number of static attributes (24).

\begin{table}[hbtp]
\caption{Statistics of the datasets used in the experiments.}
\label{table:dataset_stats}
\begin{center}
\begin{adjustbox}{max width=\textwidth}
\begin{tabular}{@{}lcccccccccccc@{}}
\toprule  &  & min  & med  & max  & trunc  & \# variants & pos class  & \# event & \# static  & \# dynamic  & \# static  & \# dynamic \\ 
 dataset & \# traces &  length &  length &  length &  length & (after trunc) & ratio & classes &  attr-s & attr-s &  cat levels & cat levels\\ \midrule
bpic2011\_1 & 1140 & 1 & 25.0 & 1814 & 36 & 815 & 0.4 & 193 & 6 & 14 & 961 & 290 \\
bpic2011\_2 & 1140 & 1 & 54.5 & 1814 & 40 & 977 & 0.78 & 251 & 6 & 14 & 994 & 370 \\
bpic2011\_3 & 1121 & 1 & 21.0 & 1368 & 31 & 793 & 0.23 & 190 & 6 & 14 & 886 & 283 \\
bpic2011\_4 & 1140 & 1 & 44.0 & 1432 & 40 & 977 & 0.28 & 231 & 6 & 14 & 993 & 338 \\
bpic2015\_1 & 696 & 2 & 42.0 & 101 & 40 & 677 & 0.23 & 380 & 17 & 12 & 19 & 433 \\
bpic2015\_2 & 753 & 1 & 55.0 & 132 & 40 & 752 & 0.19 & 396 & 17 & 12 & 7 & 429 \\
bpic2015\_3 & 1328 & 3 & 42.0 & 124 & 40 & 1280 & 0.2 & 380 & 18 & 12 & 18 & 428 \\
bpic2015\_4 & 577 & 1 & 42.0 & 82 & 40 & 576 & 0.16 & 319 & 15 & 12 & 9 & 347 \\
bpic2015\_5 & 1051 & 5 & 50.0 & 134 & 40 & 1048 & 0.31 & 376 & 18 & 12 & 8 & 420 \\
production & 220 & 1 & 9.0 & 78 & 23 & 203 & 0.53 & 26 & 3 & 15 & 37 & 79 \\
insurance\_1 & 1065 & 6 & 12.0 & 100 & 8 & 785 & 0.16 & 9 & 0 & 22 & 0 & 207 \\
insurance\_2 & 1065 & 6 & 12.0 & 100 & 13 & 924 & 0.26 & 9 & 0 & 22 & 0 & 207 \\
sepsis\_1 & 754 & 5 & 14.0 & 185 & 30 & 684 & 0.14 & 14 & 24 & 13 & 195 & 38 \\
sepsis\_2 & 782 & 4 & 13.0 & 60 & 13 & 656 & 0.14 & 15 & 24 & 13 & 200 & 40 \\
sepsis\_3 & 782 & 4 & 13.0 & 185 & 22 & 709 & 0.14 & 15 & 24 & 13 & 200 & 40 \\
bpic2012\_1 & 4685 & 15 & 35.0 & 175 & 40 & 3578 & 0.48 & 36 & 1 & 10 & 0 & 99 \\
bpic2012\_2 & 4685 & 15 & 35.0 & 175 & 40 & 3578 & 0.17 & 36 & 1 & 10 & 0 & 99 \\
bpic2012\_3 & 4685 & 15 & 35.0 & 175 & 40 & 3578 & 0.35 & 36 & 1 & 10 & 0 & 99 \\
bpic2017\_1 & 31413 & 10 & 35.0 & 180 & 20 & 2087 & 0.41 & 26 & 3 & 20 & 13 & 194 \\
bpic2017\_2 & 31413 & 10 & 35.0 & 180 & 20 & 2087 & 0.12 & 26 & 3 & 20 & 13 & 194 \\
bpic2017\_3 & 31413 & 10 & 35.0 & 180 & 20 & 2087 & 0.47 & 26 & 3 & 20 & 13 & 194 \\
traffic & 129615 & 2 & 4.0 & 20 & 10 & 185 & 0.46 & 10 & 4 & 14 & 54 & 173 \\
hospital\_1 & 77525 & 2 & 6.0 & 217 & 6 & 246 & 0.1 & 18 & 1 & 21 & 23 & 1756 \\
hospital\_2 & 77525 & 2 & 6.0 & 217 & 8 & 358 & 0.05 & 17 & 1 & 21 & 23 & 1755 \\
\bottomrule
\end{tabular}
\end{adjustbox}
\end{center}
\end{table}

While most of the data attributes can be readily included in the train and test datasets, timestamps should be preprocessed in order to derive meaningful features. In our experiments, we use the following features extracted from the timestamp: month, weekday, hour, duration from the previous event in the given case, duration from the start of the case, and the position of the event in the case.
Additionally, some recent works have shown that adding features extracted from collections of cases (inter-case features) increases the accuracy of the predictive models, particularly when predicting deadline violations~\cite{conforti2015recommendation,Senderovich2017}. For example, waiting times are highly dependent on the number of ongoing cases of a process (the so-called ``Work-In-Process''). In turn, waiting times may affect the outcome of a case, particularly if the outcome is defined with respect to a deadline or with respect to customer satisfaction.
Accordingly, we extract an inter-case feature reflecting the number of cases that are ``open'' at the time of executing a given event. All the abovementioned features are used as numeric dynamic (event) attributes.

Not surprisingly, recent work along this latter direction has shown that adding features extracted from collections of cases (inter-case features) increases the accuracy of the predictive models, particularly when predicting deadline violations~\cite{conforti2015recommendation,Senderovich2017}.


Each of the categorical attributes has a fixed number of possible values, called \emph{levels}. For some attributes, the number of distinct levels can be very large, with some of the levels appearing in only a few cases. In order to avoid exploding the dimensionality of the input dataset, we filter only the category levels that appear in at least 10 samples.
This filtering is applied to each categorical attribute except the event class (activity), where we use all category levels.

\subsection{Experimental set-up}

In this subsection, we start with describing the employed evaluation measures. We then proceed with describing our approach to splitting the event logs into train and test datasets and optimizing the hyperparameters of the compared methods.

\subsubsection{Research questions and evaluation measures}


In a predictive process monitoring use case, the quality of the predictions is typically measured with respect to two main dimensions based on the following desiderata: A good prediction should be \emph{accurate} and it should be made in the \emph{early} stages of the process.
A prediction that is often inaccurate is a useless prediction, as it cannot be relied on when making decisions. Therefore, accuracy is, in a sense, the most important quality of a prediction. The earlier an accurate prediction is made, the more useful it is in practice, as it leaves more time to act upon the prediction. Based on this rationale, we formulate the first subquestion (RQ3.1) as follows: How do the existing outcome-oriented predictive business process monitoring techniques compare in terms of accuracy and earliness of the predictions?

Different metrics can be used to measure the accuracy of predictions. Rather than returning a hard prediction (a binary number) on the expected case outcome, the classifiers usually output a real-valued score, reflecting how likely it is that the case will end in one way or the other. A good classifier will give higher scores to cases that will end with a positive outcome, and lower values to those ending with a negative one. Based on this intuition, we use the \emph{area under the ROC curve (AUC)} metric that expresses the probability that a given classifier will rank a positive case higher than a negative one. A major advantage of the AUC metric over the commonly used \emph{accuracy} (the proportion of correctly classified instances) or \emph{F-score} (the harmonic mean of precision and recall), is that the AUC remains unbiased even in case of a highly imbalanced distribution of class labels~\cite{bradley1997use}. \added{Furthermore, AUC is a threshold-independent measure, as it operates on the ranking of the scores rather than the binary class values. Still, relying on a single evaluation criterion may provide a biased viewpoint of the results; therefore, we report the F-scores (on the default threshold of 0.5) additionally to AUC.}

From the literature, two different approaches emerge for measuring the earliness of the predictions. One way~\cite{leontjeva2015complex} is to evaluate the models separately for each prefix length. In each step, the prediction model is applied to a subset of prefixes of exactly the given length. The improvement of prediction accuracy as the prefix length increases provides an implicit notion of earliness. In particular, the smaller the prefix length when an acceptable level of accuracy is reached, the better the method in terms of earliness. If needed, earliness can be defined explicitly as a metric --- the smallest prefix length where the model achieves a specified accuracy threshold.
Another option is to keep monitoring each case until the classifier gives an outcome prediction with a sufficiently high confidence, and then we measure the earliness as the average prefix length when such a prediction is made~\cite{di2016clustering, teinemaa2016predictive}. The latter approach is mostly relevant in failure prediction scenarios, when the purpose is to raise an alarm when the estimated risk becomes higher than a pre-specified threshold (for a survey of failure prediction models, see~\cite{salfner2010survey}). However, even when the predictions come with a high confidence score, they might not necessarily be accurate.
In the benchmark, we employ the first approach to measuring earliness, evaluating the models on each prefix length, because it provides a more straightforward representation of earliness that is relevant for all business processes. Also, it assures that the ``early predictions'' have reached a suitable level of accuracy.

In order to be applicable in practice a prediction should be produced \emph{efficiently}, i.e., the execution times should be suitable for a given application. To address this, we formulate the second subquestion (RQ3.2) as follows: How do the existing outcome-oriented predictive business process monitoring techniques compare in terms of execution times? When measuring the execution times of the methods, we distinguish the time taken in the offline and the online modes. The \emph{offline time} is the total time needed to construct the classifier from the historic traces available in an event log. Namely, it includes the time for constructing the prefix log, bucketing and encoding the prefix traces, and training the classifier. \added{Note that we do not add the time spent on model selection to the offline time measurements. The reason for this is that we consider the extent of hyperparameter optimization to be largely dependent on the requirements and the constraints of a given project. Specifically, if obtaining a final model with minimal amount of time is critical in a project, one can settle for a smaller
number of iterations for hyperparameter optimization, while if the accuracy of the final model is of greater importance than the time for obtaining the model itself, more time can be spent on model selection. Still, a rough estimate of the total time needed for optimizing the hyperparameters can be obtained by multiplying the time taken for building the final model with the number of optimization rounds to be performed.} In the online phase, it is essential that a prediction is produced almost instantaneously, as the predictions are usually needed in real time. Accordingly, we define the \emph{online time} as the average time for processing one incoming event (incl.\ bucketing, encoding, and predicting based on this new event).

The execution times are affected by mainly two factors. Firstly, since each prefix of a trace constitutes one sample, the lengths of the traces have a direct effect on the number of (training) samples. It is natural that the more samples are used for training, the better accuracy the predictive monitoring system could yield. At the same time, using more samples increases the execution times of the system. In applications where the efficiency of the predictions is of critical importance, reducing the number of training samples can yield a reasonable tradeoff, bringing down the execution times to a suitable level, while accepting lower accuracy. One way to reduce the number of samples is gap-based filtering~\cite{di2016clustering}, where a prefix is added to the training set only after each $g$ events in the trace. This leads us to the third subquestion (RQ3.3): To what extent does gap-based filtering improve the execution times of the predictions?

The second factor that affects the execution times is the number and the diversity of attributes that need to be processed. In particular, the number of unique values (levels) in the categorical attribute domains has a direct effect on the length of the feature vector constructed for each sample, since each level corresponds to a feature in the vector (this holds for one hot encoding, as well as using occurrences or frequencies). The dimensionality of the vector can be controlled by filtering of the levels, for instance, by using only the most frequent levels for each categorical attribute. However, such filtering may negatively impact the accuracy of the predictions. In the fourth subquestion (RQ3.4), we aim to answer the following: To what extent does filtering the levels of categorical attributes based on their frequencies improve the execution times of the predictions?

\subsubsection{Train-test split} In order to simulate the real-life situation where prediction models are trained using historic data and applied to ongoing cases, we employ a temporal split to divide the event log into train and test cases. Namely, the cases are ordered according to the start time and the first 80\% are used for \added{selecting the best model parameters and} training the \added{final} model, while the remaining 20\% are used to evaluate the performance \added{of the final model}.
\added{Specifically, splitting is done on the level of completed traces, so that different prefixes of the same trace remain in the same chunk (either all in the train set or all in the test set).}
In other words, the classifier is \added{optimized and} trained with all cases that started before a given date, and the testing is done only on cases that start afterwards.
Note that, using this approach, some events in the training cases
could still overlap with the test period. In order to avoid that, we cut the
training cases so that events that overlap with the test period are discarded.

\subsubsection{Classifier learning and bucketing parameters}
We selected four classification algorithms for the experiments: random forest (RF), gradient boosted trees (XGBoost), logistic regression (logit), and support vector machines (SVM). We chose logistic regression because of its simplicity and wide application in various machine learning applications. SVM and RF have been used in existing outcome-oriented predictive monitoring studies, whereas RF has shown to outperform many other methods (such as decision trees) in both predictive monitoring scenarios~\cite{leontjeva2015complex} and in more general empirical studies~\cite{fernandez2014we}. We also included the XGBoost classifier which has recently gained attention and showed promising results when applied to business process data~\cite{rozumnyi2017nirdizati,Senderovich2017}. Furthermore, a recent empirical study on the performance of classification algorithms across 165 datasets has shown that RF and boosted trees generally outperform other classifier learning techniques~\cite{Olson2017}.
For the clustering-based bucketing approach (cf.\ Section~\ref{subsec:clustering}), we use the k-means clustering algorithm, which is one of the most widely used clustering methods in general.

The classification algorithms as well as some of the bucketing methods (clustering and KNN), require one to specify a number of parameters.
In order to achieve good performance with each of the techniques, we optimize the hyperparameters using the Tree-structured Parzen Estimator (TPE) algorithm~\cite{bergstra2011algorithms}, separately for each combination of a dataset, a bucketing method, and a sequence encoding method. For each combination of parameter values (i.e., a configuration) we performed 3-fold cross validation within the whole set of prefix traces $L^*$ extracted from the training set, and we selected the configuration that led to the highest mean AUC calculated across the three folds.
In the case of the prefix length based bucketing method, an optimal configuration was chosen for each prefix length separately (i.e., for each combination of a dataset, a bucketing method, an encoding approach and a prefix length).
Table~\ref{table:hyperparameters} presents the bounds and the sampling distributions for each of the parameters, given as input to the optimizer.
In the case of RF and XGBoost, we found via exploratory testing that the results are almost unaffected by the \emph{number of estimators (i.e., trees)} trained per model. Therefore, we use a fixed value of $n\_estimators = 500$ throughout the experiments.

\begin{table}[hbtp]
\caption{Hyperparameters and distributions used in optimization via TPE.}
\label{table:hyperparameters}
\begin{center}
\begin{tabular}{@{}llll@{}}
\toprule
Classifier & Parameter & Distribution & Values \\
\midrule
RF & Max features & Uniform & $x \in [0, 1]$ \\
\midrule
\multirow{5}{*}{XGBoost} & Learning rate & Uniform & $x \in [0, 1]$ \\
& Subsample & Uniform & $x \in [0.5, 1]$ \\
& Max tree depth & Uniform integer & $x \in [4, 30]$ \\
& Colsample bytree & Uniform & $x \in [0.5, 1]$ \\
& Min child weight & Uniform integer & $x \in [1, 6]$ \\
\midrule
Logit & Inverse of regularization strength (C) & Uniform integer & $2^x, x \in [-15, 15]$ \\
\midrule
\multirow{2}{*}{SVM} & Penalty parameter of the error term (C) & Uniform integer & $2^x, x \in [-15, 15]$ \\
& Kernel coefficient (gamma) & Uniform integer & $2^x, x \in [-15, 15]$ \\
\midrule
K-means & Number of clusters & Uniform integer & $x \in [2, 50]$ \\
\midrule
KNN & Number of neighbors & Uniform integer & $x \in [2, 50]$ \\
\bottomrule
\end{tabular}
\end{center}
\end{table}

Both k-means and KNN require us to map each trace prefix into a feature vector in order to compute the Euclidean distance between pairs of prefixes. To this end, we applied the aggregation encoding approach, meaning that we map each trace to a vector that tells us how many times each possible activity appears in the trace. In order to keep consistent with the original methods, we decided to use only the control flow information for the clustering and the determining of the nearest neighbors.

In the case of the state-based bucketing, we need to specify a function that maps each trace prefix to a state. To this end, we used the last-activity encoding, meaning that one state is defined per possible activity and a trace prefix is mapped to the state corresponding to the last activity in the prefix. Note that the number of buckets produced by this approach is equal to the number of unique activities in the dataset (see Table~\ref{table:dataset_stats}). The reason for this choice is because this approach leads to reasonably large buckets. We also experimented with the multiset state abstraction approach, but it led to too many buckets, some of small size, so that in general there were not enough samples per bucket to train a classifier with sufficient accuracy.

When using a state-based or a clustering-based bucketing method, it may happen that a given bucket contains too few trace prefixes to train a meaningful classifier. Accordingly, we set a minimum bucket size threshold. If the number of trace prefixes in a bucket is less than the threshold, we do not build a classifier for that bucket but instead, any trace prefix falling in that bucket is mapped to the label (i.e., the outcome) that is predominant in that bucket, with a likelihood score equal to the ratio of trace prefixes in the bucket that have the predominant label. To be consistent with the choice of the parameter K in the KNN approach proposed in~\cite{maggi2014predictive}, we fixed the minimum bucket size threshold to 30. Similarly, when all of the training instances in a bucket belong to the same class, no classifier is trained for this bucket and, instead, the test instances falling to this bucket are simply assigned the same class (i.e., the assigned prediction score is either 0 or 1).

In case of logit and SVM, the features are standardized by subtracting the mean and scaling to unit variance before given as input to the classifier.

\subsubsection{Filtering and feature encoding parameters}

As discussed in Section~\ref{subsec:filtering}, training a classifier over the entire prefix log $L^*$ (all prefixes of all traces) can be time-consuming. Furthermore, we are only interested in making predictions for earlier events rather than making predictions towards the end of a trace. Additionally, we observe that the distributions of the lengths of the traces can be different within the classes corresponding to different outcomes (see Figures~\ref{fig:case_length_hist_3cols_1}-\ref{fig:case_length_hist_3cols_2} in Appendix). When all instances of long prefixes belong to the same class, predicting the outcome for these (or longer) prefixes becomes trivial.
Accordingly, during both the training and the evaluating phases, we vary the prefix length from 1 to the point where 90\% of the minority class have finished (or until the end of the given trace, if it ends earlier than this point). For computational reasons, we set the upper limit of the prefix lengths to 40, except for the \emph{bpic2017} datasets where we further reduced the limit to 20.
We argue that setting a limit to the maximum prefix length is a reasonable design choice, as the aim of predictive process monitoring is to predict as early as possible and, therefore, we are more interested in predictions made for shorter prefixes.
When answering RQ3.3, we additionally apply the gap-based filtering to the training set with $g \in \{3, 5\}$. For instance, in case of $g=5$, only prefixes of lengths 1, 6, 11, 16, 21, 26, 31, and 36 are included in the training set.


In Section~\ref{subsec:aggregation}, we noted that the aggregation encoding requires us to specify an aggregation function for each event attribute. For activities and resource attributes we use the count (frequency) aggregation function (i.e., how many times a given activity has been executed, or how many activities has a given resource executed).
The same principle is applied to any other event attribute of a categorical type.
For each numeric event attribute, we include two numeric features in the feature vector: the mean and the standard deviation.
Furthermore, to answer RQ3.4, we filter each of the categorical attribute domains by using only the top $\{10, 25, 50, 75, 90\}$ percent of the most frequent levels from each attribute.

For index-based encoding (Section~\ref{subsec:index}), we focus on the basic index-encoding technique without the HMM extension proposed in~\cite{leontjeva2015complex}. The reason is that the results reported in \cite{leontjeva2015complex} do not show that HMM provides any visible improvement, and instead this encoding adds complexity to the training phase.

\subsection{Results: accuracy and earliness}

Table \ref{table:avg_results_xgboost} reports the overall AUC \added{and F-score} for each dataset and method using XGBoost, while Tables \ref{table:avg_results_rf}, \ref{table:avg_results_logit} and \ref{table:avg_results_svm} in the Appendix report the same results for RF, logit, and SVM. The overall \replaced{metric}{AUC} values \added{(AUR or F-score)} are obtained by first calculating the \deleted{AUC} scores separately for each prefix length (using only prefixes of a given length) and then by taking the weighted average of the obtained scores, where the weights are assigned according to the number of prefixes used for the calculation of a given \deleted{AUC} score. This weighting assures that the overall \replaced{metrics are}{AUC is} influenced equally by each prefix in the evaluation set, instead of being biased towards longer prefixes (i.e., where many cases have already finished). The best-performing classifiers are XGBoost, which achieves the highest AUC \added{in 15 out of 24 datasets and the highest F-score in 11 datasets, and RF, which achieves the best AUC in 11 datasets and the top F-score in 14. Logit achieves the highest AUC in 7 and the highest F-score in 6 datasets. SVM in general does not reach the same level of accuracy as the other classifiers, the only exceptions being \emph{bpic2012\_3}, \emph{traffic}, and \emph{hospital\_2} (and only in terms of AUC)}.

\begin{table}[hbtp]
\caption{Overall AUC \added{(F-score)} for {\bfseries XGBoost}}
\vspace{-\baselineskip}
\label{table:avg_results_xgboost}
\begin{center}
\begin{adjustbox}{max width=\textwidth}
\begin{tabular}{@{}lcccccc@{}}
\toprule
 & {\bfseries bpic2011\_1} & {\bfseries bpic2011\_2} & {\bfseries bpic2011\_3} & {\bfseries bpic2011\_4} & {\bfseries insurance\_1} & {\bfseries insurance\_2}
\\ \midrule
single\_laststate & $0.85$ $(0.73)$ & $0.91$ $(0.82)$ & $0.94$ $(0.78)$ & $\bm{0.89}$ $(\bm{0.8})$ & $0.86$ $(0.36)$ & $0.83$ $(0.44)$ \\
single\_agg & $0.94$ $(0.86)$ & $\bm{0.98}$ $(\bm{0.95})$ & $\bm{0.98}$ $(\bm{0.94})$ & $0.86$ $(0.78)$ & $0.9$ $(0.5)$ & $0.8$ $(0.51)$ \\
knn\_laststate & $0.87$ $(0.86)$ & $0.91$ $(0.93)$ & $0.88$ $(0.81)$ & $0.71$ $(0.64)$ & $0.85$ $(0.49)$ & $0.78$ $(0.49)$ \\
knn\_agg & $0.87$ $(0.85)$ & $0.91$ $(0.93)$ & $0.88$ $(0.82)$ & $0.72$ $(0.64)$ & $0.84$ $(0.52)$ & $0.78$ $(0.5)$ \\
state\_laststate & $0.87$ $(0.73)$ & $0.91$ $(0.84)$ & $0.93$ $(0.8)$ & $0.87$ $(0.77)$ & $0.89$ $(0.55)$ & $\bm{0.84}$ $(0.59)$ \\
state\_agg & $0.94$ $(0.84)$ & $0.95$ $(0.91)$ & $0.97$ $(0.89)$ & $0.85$ $(0.75)$ & $0.89$ $(0.59)$ & $0.83$ $(\bm{0.6})$ \\
cluster\_laststate & $0.89$ $(0.74)$ & $0.91$ $(0.86)$ & $0.97$ $(0.9)$ & $\bm{0.89}$ $(\bm{0.8})$ & $0.87$ $(0.38)$ & $0.81$ $(0.45)$ \\
cluster\_agg & $\bm{0.95}$ $(0.84)$ & $0.97$ $(0.94)$ & $0.97$ $(0.9)$ & $0.84$ $(0.75)$ & $\bm{0.91}$ $(0.57)$ & $0.8$ $(0.45)$ \\
prefix\_index & $0.93$ $(0.79)$ & $0.94$ $(0.82)$ & $0.97$ $(0.8)$ & $0.85$ $(0.74)$ & $0.89$ $(0.55)$ & $0.8$ $(0.55)$ \\
prefix\_laststate & $0.89$ $(0.76)$ & $0.94$ $(0.86)$ & $0.95$ $(0.74)$ & $0.88$ $(0.78)$ & $0.87$ $(0.42)$ & $0.83$ $(0.53)$ \\
prefix\_agg & $0.94$ $(\bm{0.87})$ & $\bm{0.98}$ $(0.94)$ & $\bm{0.98}$ $(0.85)$ & $0.86$ $(0.77)$ & $0.9$ $(\bm{0.6})$ & $0.83$ $(\bm{0.6})$ \\
\midrule
 & {\bfseries bpic2015\_1} & {\bfseries bpic2015\_2} & {\bfseries bpic2015\_3} & {\bfseries bpic2015\_4} & {\bfseries bpic2015\_5} & {\bfseries production}
\\ \midrule
single\_laststate & $0.81$ $(0.42)$ & $0.83$ $(0.34)$ & $0.78$ $(0.45)$ & $0.8$ $(0.41)$ & $0.83$ $(0.67)$ & $0.62$ $(0.57)$ \\
single\_agg & $\bm{0.89}$ $(\bm{0.62})$ & $\bm{0.92}$ $(\bm{0.75})$ & $0.9$ $(\bm{0.75})$ & $0.85$ $(0.62)$ & $0.87$ $(\bm{0.77})$ & $0.7$ $(\bm{0.59})$ \\
knn\_laststate & $0.8$ $(0.37)$ & $0.87$ $(0.64)$ & $0.83$ $(0.6)$ & $0.81$ $(0.55)$ & $0.86$ $(0.72)$ & $0.62$ $(0.56)$ \\
knn\_agg & $0.79$ $(0.39)$ & $0.87$ $(0.67)$ & $0.84$ $(0.61)$ & $0.8$ $(0.56)$ & $0.86$ $(0.72)$ & $0.62$ $(0.55)$ \\
state\_laststate & $0.77$ $(0.46)$ & $0.85$ $(0.56)$ & $0.85$ $(0.56)$ & $0.86$ $(0.46)$ & $0.85$ $(0.66)$ & $0.62$ $(0.51)$ \\
state\_agg & $0.8$ $(0.54)$ & $0.88$ $(0.67)$ & $0.87$ $(0.61)$ & $\bm{0.88}$ $(0.63)$ & $0.87$ $(0.72)$ & $0.68$ $(0.56)$ \\
cluster\_laststate & $0.7$ $(0.39)$ & $0.85$ $(0.46)$ & $0.86$ $(0.66)$ & $0.87$ $(0.61)$ & $0.87$ $(0.71)$ & $0.68$ $(0.57)$ \\
cluster\_agg & $0.88$ $(0.58)$ & $\bm{0.92}$ $(0.73)$ & $0.9$ $(0.72)$ & $0.87$ $(\bm{0.64})$ & $0.88$ $(0.76)$ & $0.71$ $(\bm{0.59})$ \\
prefix\_index & $0.8$ $(0.46)$ & $0.83$ $(0.39)$ & $0.88$ $(0.63)$ & $0.86$ $(0.5)$ & $0.85$ $(0.64)$ & $0.68$ $(0.57)$ \\
prefix\_laststate & $0.75$ $(0.32)$ & $0.82$ $(0.28)$ & $0.76$ $(0.4)$ & $0.82$ $(0.4)$ & $0.83$ $(0.62)$ & $0.68$ $(0.56)$ \\
prefix\_agg & $0.84$ $(0.6)$ & $0.88$ $(0.7)$ & $\bm{0.91}$ $(0.71)$ & $0.87$ $(0.6)$ & $\bm{0.89}$ $(0.75)$ & $\bm{0.73}$ $(0.57)$ \\
\midrule
 & {\bfseries sepsis\_1} & {\bfseries sepsis\_2} & {\bfseries sepsis\_3} & {\bfseries bpic2012\_1} & {\bfseries bpic2012\_2} & {\bfseries bpic2012\_3}
\\ \midrule
single\_laststate & $0.4$ $(\bm{0.08})$ & $0.84$ $(\bm{0.48})$ & $0.65$ $(0.24)$ & $0.68$ $(0.61)$ & $0.59$ $(0.09)$ & $\bm{0.7}$ $(0.38)$ \\
single\_agg & $0.33$ $(0.0)$ & $\bm{0.85}$ $(0.42)$ & $0.72$ $(\bm{0.35})$ & $\bm{0.7}$ $(0.59)$ & $0.57$ $(0.16)$ & $0.69$ $(0.36)$ \\
knn\_laststate & $\bm{0.49}$ $(0.07)$ & $0.61$ $(0.01)$ & $0.61$ $(0.23)$ & $0.57$ $(\bm{0.72})$ & $0.55$ $(\bm{0.34})$ & $0.59$ $(0.45)$ \\
knn\_agg & $0.45$ $(0.05)$ & $0.68$ $(0.05)$ & $0.59$ $(0.15)$ & $0.63$ $(0.61)$ & $0.59$ $(0.05)$ & $0.63$ $(\bm{0.48})$ \\
state\_laststate & $0.39$ $(0.0)$ & $0.83$ $(0.42)$ & $0.71$ $(0.13)$ & $0.68$ $(0.62)$ & $0.61$ $(0.12)$ & $\bm{0.7}$ $(0.35)$ \\
state\_agg & $0.42$ $(0.0)$ & $0.83$ $(0.43)$ & $0.71$ $(0.2)$ & $\bm{0.7}$ $(0.59)$ & $0.6$ $(0.13)$ & $\bm{0.7}$ $(0.33)$ \\
cluster\_laststate & $0.48$ $(0.04)$ & $0.81$ $(0.42)$ & $0.72$ $(0.11)$ & $0.65$ $(0.6)$ & $0.59$ $(0.12)$ & $0.69$ $(0.3)$ \\
cluster\_agg & $0.46$ $(0.0)$ & $0.82$ $(0.44)$ & $0.7$ $(0.28)$ & $0.67$ $(0.6)$ & $0.6$ $(0.17)$ & $\bm{0.7}$ $(0.29)$ \\
prefix\_index & $0.44$ $(0.07)$ & $0.79$ $(0.36)$ & $\bm{0.73}$ $(0.15)$ & $0.68$ $(0.61)$ & $\bm{0.62}$ $(0.13)$ & $0.69$ $(0.35)$ \\
prefix\_laststate & $0.47$ $(0.07)$ & $0.82$ $(0.38)$ & $0.72$ $(0.06)$ & $0.66$ $(0.61)$ & $0.59$ $(0.1)$ & $0.69$ $(0.35)$ \\
prefix\_agg & $0.48$ $(\bm{0.08})$ & $0.8$ $(0.4)$ & $0.71$ $(0.21)$ & $0.68$ $(0.61)$ & $0.59$ $(0.13)$ & $\bm{0.7}$ $(0.35)$ \\
\midrule
 & {\bfseries bpic2017\_1} & {\bfseries bpic2017\_2} & {\bfseries bpic2017\_3} & {\bfseries traffic} & {\bfseries hospital\_1} & {\bfseries hospital\_2}
\\ \midrule
single\_laststate & $0.81$ $(0.66)$ & $\bm{0.81}$ $(0.42)$ & $0.79$ $(0.73)$ & $0.66$ $(0.67)$ & $0.89$ $(\bm{0.66})$ & $0.73$ $(0.11)$ \\
single\_agg & $\bm{0.84}$ $(0.71)$ & $\bm{0.81}$ $(0.45)$ & $0.79$ $(\bm{0.76})$ & $0.66$ $(0.67)$ & $\bm{0.9}$ $(0.63)$ & $\bm{0.76}$ $(0.08)$ \\
knn\_laststate & $0.76$ $(0.66)$ & $0.6$ $(0.04)$ & $0.62$ $(0.53)$ & $0.63$ $(0.69)$ & $0.78$ $(0.37)$ & $0.56$ $(0.06)$ \\
knn\_agg & $0.74$ $(0.59)$ & $0.56$ $(0.0)$ & $0.62$ $(0.52)$ & $0.59$ $(\bm{0.7})$ & $0.75$ $(0.47)$ & $0.55$ $(0.01)$ \\
state\_laststate & $0.83$ $(0.7)$ & $0.79$ $(0.45)$ & $0.78$ $(0.72)$ & $0.66$ $(0.67)$ & $\bm{0.9}$ $(0.65)$ & $0.74$ $(0.11)$ \\
state\_agg & $0.83$ $(0.7)$ & $0.79$ $(\bm{0.46})$ & $0.79$ $(0.73)$ & $\bm{0.67}$ $(0.66)$ & $\bm{0.9}$ $(0.65)$ & $0.69$ $(0.11)$ \\
cluster\_laststate & $\bm{0.84}$ $(0.69)$ & $0.8$ $(0.39)$ & $0.78$ $(0.72)$ & $0.66$ $(0.67)$ & $0.89$ $(0.64)$ & $0.68$ $(\bm{0.12})$ \\
cluster\_agg & $\bm{0.84}$ $(0.7)$ & $0.79$ $(0.44)$ & $0.79$ $(0.73)$ & $\bm{0.67}$ $(0.66)$ & $0.88$ $(0.64)$ & $0.69$ $(0.09)$ \\
prefix\_index & $0.83$ $(\bm{0.72})$ & $0.8$ $(0.45)$ & $\bm{0.8}$ $(0.73)$ & $\bm{0.67}$ $(0.66)$ & $0.87$ $(0.64)$ & $0.69$ $(0.11)$ \\
prefix\_laststate & $0.82$ $(0.7)$ & $0.76$ $(0.44)$ & $0.79$ $(0.72)$ & $0.66$ $(0.66)$ & $0.86$ $(0.63)$ & $0.74$ $(0.08)$ \\
prefix\_agg & $\bm{0.84}$ $(0.71)$ & $0.77$ $(0.45)$ & $0.79$ $(0.73)$ & $\bm{0.67}$ $(0.66)$ & $0.87$ $(0.64)$ & $0.74$ $(0.1)$ \\
\bottomrule
\end{tabular}
\end{adjustbox}
\end{center}
\end{table}

\added{In order to further assess the relative performance of the classifiers, we applied the Nemenyi test (as proposed in~\cite{demvsar2006statistical}) as a means for statistical comparison of classifiers over multiple datasets. In this setting, we compared the best AUC scores obtained by each classifier for a given dataset, i.e.\ we selected the best combination of bucketing and encoding technique for each dataset and classification algorithm. The resulting critical difference diagram (Figure~\ref{fig:cd_classifiers}), obtained using a 0.05 significance level, confirms that XGBoost is on average the best performing classifier, achieving an average rank of around 1.8. However, the difference between XGBoost, RF, and logit is not statistically significant (indicated by the horizontal line connecting these three classifiers). On the other hand, SVM performs significantly worse than XGBoost and RF.} 
In the following we analyze the results obtained by XGBoost in detail.

\begin{figure}[hbtp]
\centering
\includegraphics[width=1\textwidth]{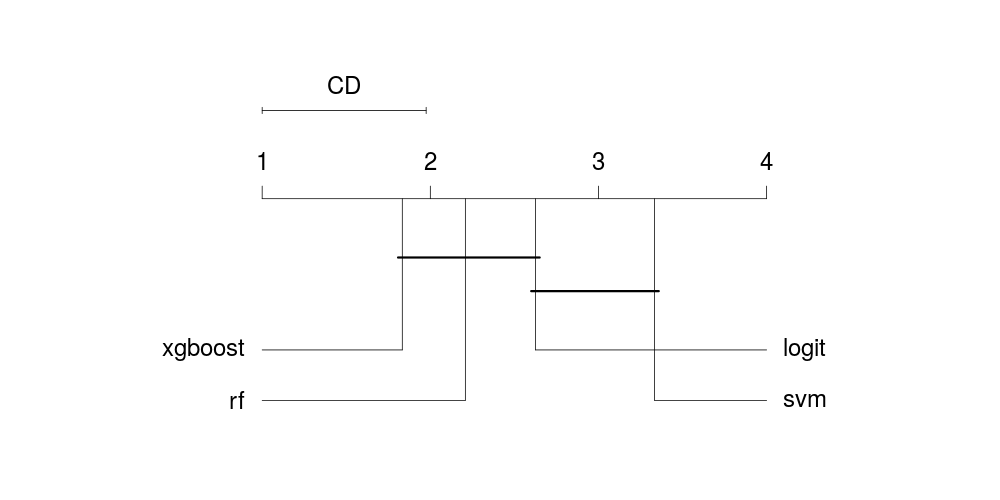}
\caption{Comparison of all classifiers against each other with the Nemenyi test. The classifiers are compared in terms of the best AUC achieved in each of the 24 datasets. Groups of classifiers that are not significantly different (at $p<.05$) are connected.}
\label{fig:cd_classifiers}
\end{figure}

Concerning the bucketing and encoding methods, we can see in Table \ref{table:avg_results_xgboost} that \added{single\_agg achieves the best AUC in 10 out of 24 datasets (and the best F-score in 9 datasets), followed by prefix\_agg, which is best in 8 datasets (4 in terms of F-score). They are followed by cluster\_agg, state\_agg, and prefix\_index, which obtain the best AUC in 6, 5, and 4 datasets, respectively. With a few exceptions, which are discussed separately below, the last state encodings in general perform worse than their aggregation encoding counterparts and KNN performs worse that the other bucketing methods.}

\added{The critical difference diagram in Figure~\ref{fig:cd_xgboost} shows that in terms of the average rank, prefix\_agg slightly outperforms single\_agg, while both are closely followed by cluster\_agg and state\_agg. Despite a larger gap in the average ranks, the differences from prefix\_agg to prefix\_index, cluster\_laststate, state\_laststate, and single\_laststate are not statistically significant either. On the other hand, prefix\_laststate, knn\_laststate, and knn\_agg are found to perform significantly worse than prefix\_agg.}

\begin{figure}[hbtp]
\centering
\includegraphics[width=1\textwidth]{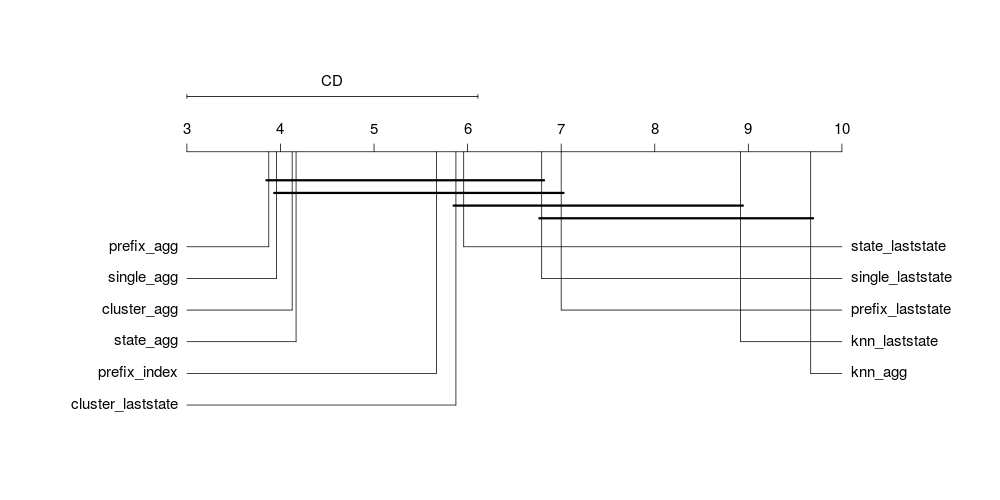}
\caption{Comparison of the bucketing/encoding combinations with the Nemenyi test. The methods are compared in terms of AUC achieved in each of the 24 datasets using the {\bfseries XGBoost} classifier. Groups of methods that are not significantly different (at $p<.05$) are connected.}
\label{fig:cd_xgboost}
\end{figure}

Figures \ref{fig:results_selected_xgboost_1} and \ref{fig:results_selected_xgboost_2} present the prediction accuracy in terms of AUC \added{for the six best performing methods (according to Figure~\ref{fig:cd_xgboost})}, evaluated over different prefix lengths\footnote{For a comparison of all the twelve methods, see Figures \ref{fig:results_all_xgboost_1} and \ref{fig:results_all_xgboost_2} in Appendix.}. Each evaluation point includes prefix traces of exactly the given length. In other words, traces that are altogether shorter than the required prefix are left out of the calculation. Therefore, the number of cases used for evaluation is monotonically decreasing when increasing prefix length. In most of the datasets, we see that starting from a specific prefix length the methods with aggregation encoding achieve perfect prediction accuracy ($AUC=1$). It is natural that the prediction task becomes trivial when cases are close to completion, especially if the labeling function is related to the control flow or to the data payload present in the event log. However, there are a few exceptions from this rule, namely, in the \emph{bpic2012} and \emph{sepsis} datasets, the results seem to decline on larger prefix sizes. To investigate this phenomenon, we recalculated the AUC scores on the longer traces only, i.e., traces that have a length larger than or equal to the maximum considered trace length (see Figure~\ref{fig:results_long_traces} in Appendix). This analysis confirmed \added{(with the exception of \emph{sepsis\_1}, which we discuss separately later in this section)} that the phenomenon is caused by the fact that the datasets contain some short traces for which it appears to be easy to predict the outcome. These short traces are not included in the later evaluation points, as they have already finished by that time. Therefore, we are left with longer traces only, which appear to be more challenging for the classifier, dragging down the total AUC score on larger prefix lengths. 

\begin{figure}[hbtp]
\centering
\includegraphics[width=1\textwidth]{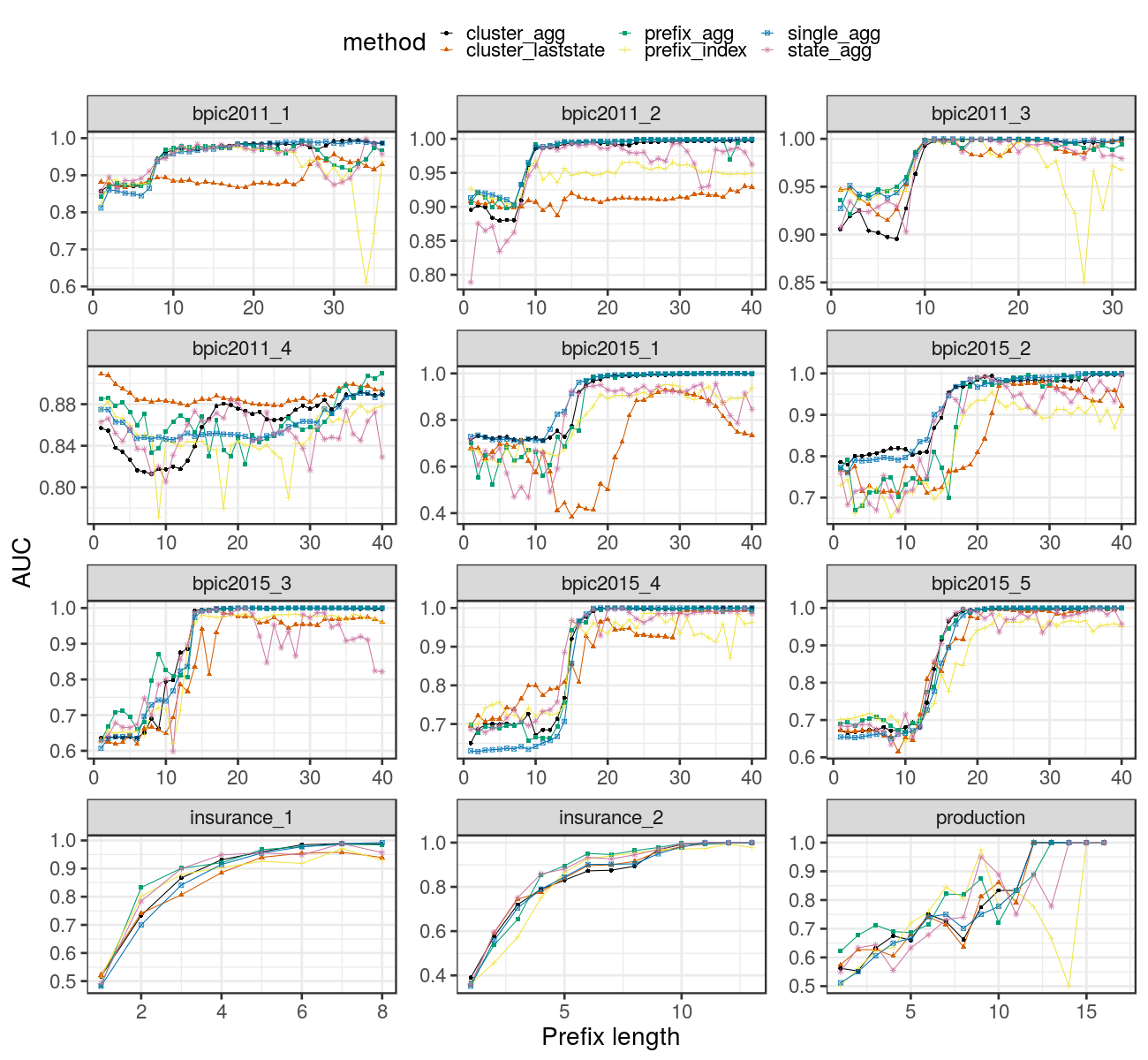}
\caption{AUC across different prefix lengths using {\bfseries XGBoost} \deleted{(1)}}
\label{fig:results_selected_xgboost_1}
\end{figure}

\begin{figure}[hbtp]
\centering
\includegraphics[width=1\textwidth]{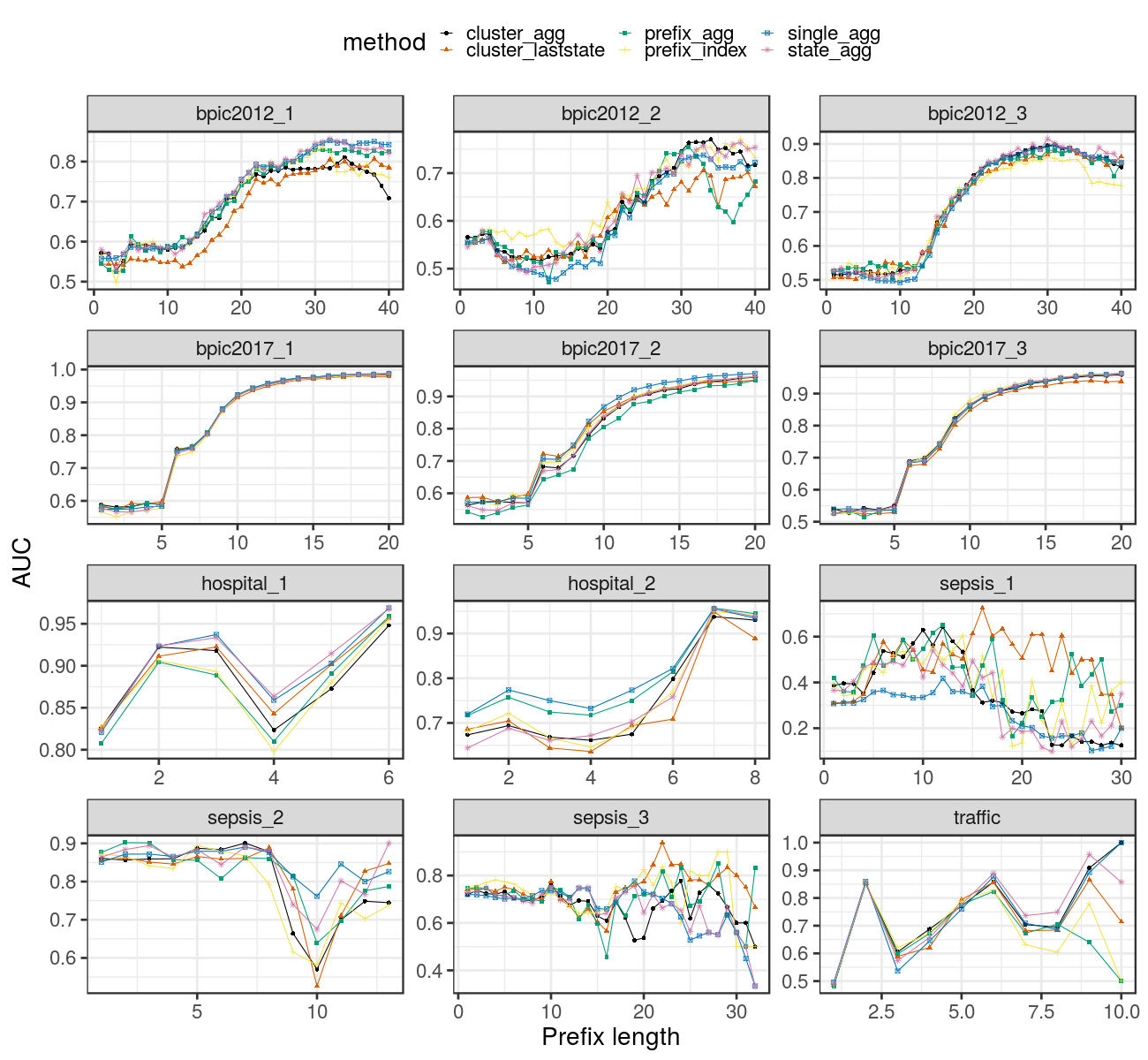}
\caption{AUC across different prefix lengths using {\bfseries XGBoost} (\replaced{continued}{2})}
\label{fig:results_selected_xgboost_2}
\end{figure}

\added{From the results presented above, we see that} the choice of the bucketing method seems to have a smaller effect on the results than the sequence encoding. Namely, the best results are usually achieved using the aggregation encoding with either \added{the single bucket, clustering, prefix length based, or state-based bucketing. In general these methods achieve very comparable results. Still, it appears that in event logs with many trace variants (relative to the total number of traces), such as \emph{insurace}, \emph{production}, \emph{bpic2012}, and \emph{bpic2015\_4} (see Table~\ref{table:dataset_stats}) it may be preferred to use a multiclassifier instead of a single bucket. However, this only holds if each bucket receives enough data to learn relevant patterns. For instance, when the number of categorical attribute levels is very high (as in the \emph{bpic2011}, \emph{bpic2015}, and \emph{hospital} datasets), a single classifier is usually able to produce better predictions. Similarly, when the classes are very imbalanced (\emph{hospital}, \emph{bpic2017\_2}, \emph{sepsis\_2}), it is likely that some buckets receive too limited information about the minority class and, therefore, a single classifier is recommended over a multiclassifier. The effect of having too many buckets can easily be seen in case of state-based bucketing when the number of different event classes (and therefore, the number of buckets) is very large (see \emph{bpic2011} and \emph{bpic2015} in Figure~\ref{fig:results_selected_xgboost_1} and the counts of training prefix traces in each bucket in Figures~\ref{fig:bucket_size_density_plots_1}-\ref{fig:bucket_size_density_plots_2} in Appendix). As a result, each classifier receives a small number of traces for training, resulting in a very spiky performance across different prefix lengths.}
The same phenomenon can be seen in case of prefix\_agg, which usually achieves very good performance, but at times can produce unexpectedly inaccurate results (like in the longer prefixes of \emph{bpic2011\_1} and \emph{bpic2012\_2}). \added{On the contrary, single\_agg and cluster\_agg in general produce stable and reliable results on all datasets and across all prefix sizes}. The optimal number of clusters in case of cluster\_agg with XGBoost was often found to be small, i.e. between 2-7 (see Table \ref{table:optimal_params_n_clusters} in Appendix), which explains why these two methods behave similarly. In some cases where the optimized number of clusters was higher,  e.g., \emph{bpic2012\_1} and \emph{hospital\_2}, the accuracy of cluster\_agg drops compared to single\_agg.

We can see from Figure \ref{fig:results_selected_xgboost_1} that in several cases (e.g., \emph{bpic2011}, \emph{bpic2015}, \emph{bpic2012\_1}, and \emph{sepsis\_2}), all the methods achieve a similar AUC on shorter prefixes, but then quickly grow apart as the size of the prefix increases. In particular, the aggregation encoding seems to be able to carry along relevant information from the earlier prefixes, while the last state encoding entails more limited information that is often insufficient for making accurate predictions. \added{Comparing the overall AUC in Table~\ref{table:avg_results_xgboost}, the last state encodings outperform the other methods in only three datasets. One such exceptional case is \emph{bpic2011\_4}, where single\_laststate and cluster\_laststate considerably outperform their aggregation encoding counterparts. A deeper investigation of this case revealed that this is due to overfitting in the presence of a concept drift in the dataset. In particular, the aggregation encodings yielded a more complex classifier (e.g. the optimized maximum tree depth is 15 in case of single\_agg and 6 in case of single\_laststate), memorizing the training data completely. However, a concept drift occurs in the relationship between some data attributes and the class labeling, which affects the aggregated features more than the ``as-is'' features (see Figure~\ref{fig:bpic2011_f4_boxplots} in Appendix). Another exception is \emph{sepsis\_1}, where the best results are achieved by knn\_laststate. In this dataset, all the methods consistently yield an AUC less than 0.5 (i.e. worse than random). Further investigation revealed that this phenomenon is also due to a concept drift in the dataset, which makes it impossible to learn useful patterns in case of a temporal train-test split. 
For instance, Figure~\ref{fig:sepsis_cases_1_boxplots} in Appendix illustrates that in the training set the values of \emph{CRP} are larger in positive instances, while in the test set the \emph{CRP} values are larger in negative instances. 
The third exceptional dataset is \emph{insurance\_2}, where state\_laststate slightly outperforms other techniques. We did not find any peculiarities in this dataset that would explain this phenomenon, however, the differences in scores (between, e.g.\ state\_laststate and state\_agg) are much smaller compared to the previous two datasets.}

We can also observe that the index-based encoding, although lossless, in general does not outperform the lossy encoding schemes\added{, reaching the highest overall AUC only in 4 datasets: \emph{bpic2012\_2}, \emph{bpic2017\_3}, \emph{sepsis\_3}, and \emph{traffic}. In these logs the number of levels in dynamic categorical attributes is not very high (see Table~\ref{table:dataset_stats}), which helps to keep the size of the feature vector in reasonable bounds. Still, even in these cases the difference in AUC compared to the other methods (such as prefix\_agg) is marginal.} 
In fact, in some datasets (e.g., \emph{hospital\_2} and \emph{sepsis\_2}) index-based encoding performs even worse than the last state encoding. This suggests that in the given datasets, the order of events is not as relevant for determining the final outcome of the case. Instead, combining the knowledge from all events performed so far provides much more signal. Alternatively, it may be that the order of events (i.e., the control-flow) does matter in some cases, but the classifiers considered in this study (including XGBoost) are not able to infer high-level control-flow features by themselves, which would explain why we see that even the simple aggregation-based methods outperform index-based encoding. This phenomenon deserves a separate in-depth study.

These observations, further supported by the fact that KNN does not appear among the top performing methods, lead to the conclusion that it is preferable to build few classifiers (or even just a single one), with a larger number of traces as input. XGBoost seems to be a classifier sophisticated enough to derive the ``bucketing patterns'' by itself when necessary. Another advantage of the single\_agg method over cluster\_agg is the simplicity of a single bucket. In fact, no additional preprocessing step for bucketing the prefix traces is needed. On the other hand, clustering (regardless of the clustering algorithm) comes with a set of parameters, such as the number of clusters in k-means, that need to be tuned for optimal performance. Therefore, the time and effort needed from the user of the system for setting up the prediction framework can be considerably higher in case of cluster\_agg, which makes single\_agg the overall preferred choice of method in terms of accuracy and earliness. This discussion concludes the answer to RQ3.1.

\subsection{Results: time performance}
The time measurements for all of the methods and classifiers, calculated as averages over 5 identical runs using the final (optimal) parameters, are presented in Tables \ref{table:performance_results_xgboost_1} and \ref{table:performance_results_xgboost_2} (XGBoost), Tables \ref{table:performance_results_rf_1} and \ref{table:performance_results_rf_2} (RF), Tables \ref{table:performance_results_logit_1} and \ref{table:performance_results_logit_2} (logit), and Tables \ref{table:performance_results_svm_1} and \ref{table:performance_results_svm_2} (SVM). In the offline phase, the fastest of the four classifiers is logit. The ordering of the others differs between the small (\emph{production}, \emph{bpic2011}, \emph{bpic2015}, \emph{insurance}, and \emph{sepsis}) and the large (\emph{bpic2017}, \emph{traffic}, \emph{hospital}) datasets. In the former group, the second fastest classifier is SVM, usually followed by RF and, then, XGBoost. Conversely, in the larger datasets, XGBoost appears to scale better than the others, while SVM tends to be the slowest of the three. In terms of online time, logit, SVM, and XGBoost yield comparable performance, while RF is usually slower than the others.
In the following, we will, again, analyse deeper the results obtained with the XGBoost classifier.

Recall that the KNN method (almost) skips the offline phase, since all the classifiers are built at runtime. The offline time for KNN still includes the time for constructing the prefix log and setting up the matrix of encoded historical prefix traces, which is later used for finding the nearest neighbors for running traces. Therefore, the offline times in case of the KNN approaches are almost negligible. The offline phase for the other methods (i.e., excluding KNN) takes between 3 seconds on the smallest dataset (\emph{production}) to 6 hours and 30 minutes on \emph{hospital\_1}. There is no clear winner between the last state encoding and the corresponding aggregation encoding counterparts, which indicates that the time for applying the aggregation functions is small compared to the time taken for training the classifier.
The most time in the offline phase is, in general, taken by index-based encoding that constructs the sequences of events for each trace.

In terms of bucketing, the fastest approach in the offline phase is usually state-based bucketing, followed by either the prefix length or the clustering based method, while the slowest is single bucket. This indicates that the time taken to train multiple (``small'') classifiers, each trained with only a subset of the original data, is smaller than training a few (``large'') classifiers using a larger portion of the data.

In general, all methods are able to process an event in less than 100 milliseconds during the online phase (the times in Tables \ref{table:performance_results_xgboost_1} and \ref{table:performance_results_xgboost_2} are in milliseconds per processed event in a partial trace). Exceptions are \emph{hospital\_1} and \emph{hospital\_2}, where processing an event takes around 0.3-0.4 seconds. The online execution times are very comparable across all the methods, except for KNN and prefix\_index. While prefix\_index often takes double the time of other methods, the patterns for KNN are less straightforward. Namely, in some datasets (\emph{bpic2012}, \emph{sepsis}, \emph{production}, \emph{insurance}, and \emph{traffic}), the KNN approaches take considerably more time than the other techniques, which can be explained by the fact that these approaches train a classifier at runtime. However, somewhat surprisingly, in other datasets (\emph{hospital} and \emph{bpic2011} datasets) the KNN approaches yield the best execution times even at runtime. A possible explanation for this is that in cases where all the selected nearest neighbors are of the same class, no classifier is trained and the class of the neighbors is immediately returned as the prediction. However, note that the overall AUC in these cases is 7-21 percentage points lower than that of the best method (\ref{table:avg_results_xgboost}). In the offline phase, the overhead of applying aggregation functions becomes more evident, with the last state encoding almost always outperforming the aggregation encoding methods by a few milliseconds. The fastest method in the online phase tends to be prefix\_laststate, which outperforms the others in 17 out of 24 datasets. It is followed by knn\_laststate, state\_laststate, and single\_laststate. 

\begin{table}[hbtp!]
\caption{Execution times for {\bfseries XGBoost}\deleted{ (1)}}
\label{table:performance_results_xgboost_1}
\begin{center}
\begin{adjustbox}{max width=\textwidth}
\begin{tabular}{@{}lcccccccc@{}}
\toprule
 & \multicolumn{2}{c}{{\bfseries bpic2011\_1}} & \multicolumn{2}{c}{{\bfseries bpic2011\_2}} & \multicolumn{2}{c}{{\bfseries bpic2011\_3}} \\ \cmidrule(lr){2-3} \cmidrule(lr){4-5} \cmidrule(lr){6-7}
method & offline\_total (s) & online\_avg (ms) & offline\_total (s) & online\_avg (ms) & offline\_total (s) & online\_avg (ms) \\ \midrule
single\_laststate & $418.35 \pm 0.56$ & $69 \pm 98$ & $581.68 \pm 1.09$ & $62 \pm 96$ & $217.69 \pm 1.38$ & $71 \pm 96$ \\
single\_agg & $317.18 \pm 0.58$ & $69 \pm 99$ & $342.3 \pm 2.02$ & $62 \pm 97$ & $271.33 \pm 0.54$ & $71 \pm 97$ \\
knn\_laststate & $5.9 \pm 0.31$ & $\bm{44 \pm 65}$ & $9.82 \pm 0.66$ & $\bm{37 \pm 59}$ & $4.14 \pm 0.06$ & $\bm{48 \pm 72}$ \\
knn\_agg & $6.63 \pm 0.12$ & $52 \pm 76$ & $9.79 \pm 0.44$ & $46 \pm 72$ & $4.57 \pm 0.06$ & $61 \pm 91$ \\
state\_laststate & $142.53 \pm 0.31$ & $52 \pm 72$ & $181.87 \pm 0.92$ & $48 \pm 74$ & $86.78 \pm 0.52$ & $53 \pm 70$ \\
state\_agg & $183.67 \pm 0.79$ & $61 \pm 84$ & $169.98 \pm 0.47$ & $58 \pm 90$ & $119.84 \pm 0.2$ & $62 \pm 82$ \\
cluster\_laststate & $211.88 \pm 1.0$ & $66 \pm 112$ & $592.92 \pm 4.84$ & $67 \pm 112$ & $93.83 \pm 0.48$ & $57 \pm 98$ \\
cluster\_agg & $341.4 \pm 1.89$ & $70 \pm 113$ & $381.61 \pm 1.87$ & $62 \pm 111$ & $94.45 \pm 0.71$ & $72 \pm 111$ \\
prefix\_index & $763.8 \pm 20.33$ & $114 \pm 68$ & $1405.87 \pm 88.64$ & $126 \pm 62$ & $428.6 \pm 19.49$ & $113 \pm 69$ \\
prefix\_laststate & $290.64 \pm 1.44$ & $57 \pm 86$ & $264.71 \pm 4.58$ & $50 \pm 81$ & $108.84 \pm 0.33$ & $59 \pm 84$ \\
prefix\_agg & $172.8 \pm 12.28$ & $56 \pm 82$ & $274.29 \pm 9.08$ & $53 \pm 82$ & $125.61 \pm 7.42$ & $58 \pm 80$ \\
\bottomrule
 & \multicolumn{2}{c}{{\bfseries bpic2011\_4}} & \multicolumn{2}{c}{{\bfseries bpic2015\_1}} & \multicolumn{2}{c}{{\bfseries bpic2015\_2}} \\ \cmidrule(lr){2-3} \cmidrule(lr){4-5} \cmidrule(lr){6-7}
method & offline\_total (s) & online\_avg (ms) & offline\_total (s) & online\_avg (ms) & offline\_total (s) & online\_avg (ms) \\ \midrule
single\_laststate & $418.65 \pm 17.42$ & $62 \pm 97$ & $263.56 \pm 0.38$ & $20 \pm 30$ & $134.87 \pm 0.59$ & $18 \pm 29$ \\
single\_agg & $319.27 \pm 10.75$ & $62 \pm 98$ & $105.58 \pm 0.39$ & $22 \pm 34$ & $282.97 \pm 8.14$ & $20 \pm 32$ \\
knn\_laststate & $8.99 \pm 0.23$ & $\bm{42 \pm 67}$ & $8.44 \pm 0.04$ & $31 \pm 51$ & $11.46 \pm 0.63$ & $34 \pm 59$ \\
knn\_agg & $9.47 \pm 0.07$ & $57 \pm 86$ & $9.45 \pm 0.06$ & $45 \pm 72$ & $11.95 \pm 0.12$ & $38 \pm 66$ \\
state\_laststate & $156.45 \pm 1.1$ & $48 \pm 73$ & $25.2 \pm 0.08$ & $25 \pm 44$ & $30.28 \pm 0.2$ & $27 \pm 45$ \\
state\_agg & $301.91 \pm 7.54$ & $58 \pm 90$ & $53.39 \pm 0.34$ & $29 \pm 48$ & $64.72 \pm 0.08$ & $31 \pm 49$ \\
cluster\_laststate & $274.97 \pm 8.03$ & $69 \pm 112$ & $46.71 \pm 0.58$ & $37 \pm 58$ & $52.8 \pm 1.6$ & $36 \pm 60$ \\
cluster\_agg & $252.81 \pm 6.36$ & $69 \pm 112$ & $62.25 \pm 0.48$ & $29 \pm 46$ & $135.33 \pm 1.21$ & $28 \pm 45$ \\
prefix\_index & $794.59 \pm 35.34$ & $111 \pm 59$ & $396.03 \pm 4.79$ & $51 \pm 10$ & $442.17 \pm 8.47$ & $48 \pm 13$ \\
prefix\_laststate & $344.73 \pm 13.5$ & $48 \pm 80$ & $62.17 \pm 0.26$ & $\bm{7 \pm 9}$ & $45.27 \pm 0.27$ & $\bm{6 \pm 7}$ \\
prefix\_agg & $441.82 \pm 1.01$ & $63 \pm 98$ & $57.14 \pm 0.16$ & $10 \pm 10$ & $74.98 \pm 1.04$ & $8 \pm 8$ \\
\bottomrule
 & \multicolumn{2}{c}{{\bfseries bpic2015\_3}} & \multicolumn{2}{c}{{\bfseries bpic2015\_4}} & \multicolumn{2}{c}{{\bfseries bpic2015\_5}} \\ \cmidrule(lr){2-3} \cmidrule(lr){4-5} \cmidrule(lr){6-7}
method & offline\_total (s) & online\_avg (ms) & offline\_total (s) & online\_avg (ms) & offline\_total (s) & online\_avg (ms) \\ \midrule
single\_laststate & $150.3 \pm 0.44$ & $21 \pm 33$ & $67.21 \pm 0.09$ & $18 \pm 27$ & $83.78 \pm 0.31$ & $17 \pm 26$ \\
single\_agg & $627.3 \pm 3.48$ & $21 \pm 35$ & $191.22 \pm 0.21$ & $19 \pm 29$ & $428.23 \pm 1.24$ & $17 \pm 28$ \\
knn\_laststate & $18.9 \pm 0.68$ & $37 \pm 61$ & $7.38 \pm 0.08$ & $29 \pm 50$ & $17.22 \pm 0.14$ & $36 \pm 60$ \\
knn\_agg & $19.57 \pm 0.84$ & $41 \pm 69$ & $7.5 \pm 0.35$ & $34 \pm 58$ & $19.67 \pm 0.53$ & $37 \pm 87$ \\
state\_laststate & $45.06 \pm 0.03$ & $29 \pm 49$ & $18.5 \pm 0.06$ & $25 \pm 42$ & $37.97 \pm 0.08$ & $23 \pm 38$ \\
state\_agg & $86.47 \pm 0.06$ & $33 \pm 53$ & $30.76 \pm 0.07$ & $29 \pm 45$ & $73.2 \pm 0.46$ & $26 \pm 41$ \\
cluster\_laststate & $102.82 \pm 0.26$ & $35 \pm 60$ & $34.94 \pm 0.19$ & $34 \pm 62$ & $72.76 \pm 0.26$ & $35 \pm 60$ \\
cluster\_agg & $181.85 \pm 1.13$ & $28 \pm 45$ & $61.06 \pm 0.84$ & $26 \pm 40$ & $74.79 \pm 0.7$ & $20 \pm 34$ \\
prefix\_index & $2155.36 \pm 80.83$ & $55 \pm 15$ & $261.21 \pm 0.5$ & $41 \pm 8$ & $550.87 \pm 4.23$ & $48 \pm 12$ \\
prefix\_laststate & $98.37 \pm 0.7$ & $\bm{7 \pm 9}$ & $37.59 \pm 0.06$ & $\bm{6 \pm 7}$ & $109.77 \pm 0.1$ & $\bm{6 \pm 8}$ \\
prefix\_agg & $108.91 \pm 2.83$ & $9 \pm 9$ & $40.83 \pm 0.38$ & $8 \pm 8$ & $92.61 \pm 0.88$ & $8 \pm 9$ \\
\bottomrule
 & \multicolumn{2}{c}{{\bfseries production}} & \multicolumn{2}{c}{{\bfseries insurance\_1}} & \multicolumn{2}{c}{{\bfseries insurance\_2}} \\ \cmidrule(lr){2-3} \cmidrule(lr){4-5} \cmidrule(lr){6-7}
method & offline\_total (s) & online\_avg (ms) & offline\_total (s) & online\_avg (ms) & offline\_total (s) & online\_avg (ms) \\ \midrule
single\_laststate & $7.28 \pm 0.12$ & $25 \pm 21$ & $36.0 \pm 0.1$ & $38 \pm 31$ & $15.52 \pm 0.09$ & $33 \pm 30$ \\
single\_agg & $4.3 \pm 0.13$ & $28 \pm 25$ & $12.81 \pm 0.12$ & $40 \pm 33$ & $78.17 \pm 0.62$ & $34 \pm 32$ \\
knn\_laststate & $1.09 \pm 0.01$ & $51 \pm 49$ & $0.66 \pm 0.01$ & $49 \pm 44$ & $0.96 \pm 0.01$ & $48 \pm 36$ \\
knn\_agg & $0.76 \pm 0.0$ & $62 \pm 60$ & $0.66 \pm 0.0$ & $57 \pm 53$ & $1.06 \pm 0.01$ & $69 \pm 50$ \\
state\_laststate & $2.69 \pm 0.11$ & $\bm{23 \pm 19}$ & $11.45 \pm 0.1$ & $\bm{31 \pm 22}$ & $15.22 \pm 0.06$ & $\bm{26 \pm 21}$ \\
state\_agg & $4.37 \pm 0.04$ & $30 \pm 26$ & $17.67 \pm 0.11$ & $40 \pm 31$ & $30.62 \pm 0.17$ & $35 \pm 30$ \\
cluster\_laststate & $6.94 \pm 0.06$ & $30 \pm 30$ & $32.75 \pm 0.11$ & $39 \pm 37$ & $42.42 \pm 0.15$ & $37 \pm 34$ \\
cluster\_agg & $6.35 \pm 0.07$ & $35 \pm 31$ & $30.06 \pm 0.2$ & $52 \pm 43$ & $47.2 \pm 0.14$ & $45 \pm 43$ \\
prefix\_index & $13.81 \pm 0.04$ & $49 \pm 10$ & $58.79 \pm 0.58$ & $90 \pm 4$ & $80.56 \pm 0.12$ & $91 \pm 4$ \\
prefix\_laststate & $5.62 \pm 0.35$ & $28 \pm 23$ & $9.44 \pm 0.03$ & $\bm{31 \pm 21}$ & $22.3 \pm 0.07$ & $\bm{26 \pm 21}$ \\
prefix\_agg & $5.89 \pm 0.01$ & $28 \pm 23$ & $17.17 \pm 0.05$ & $35 \pm 24$ & $24.66 \pm 0.07$ & $30 \pm 24$ \\
\bottomrule
\end{tabular}
\end{adjustbox}
\end{center}
\end{table}

\begin{table}[hbtp!]
\caption{Execution times for {\bfseries XGBoost} (\replaced{continued}{2})}
\label{table:performance_results_xgboost_2}
\begin{center}
\begin{adjustbox}{max width=\textwidth}
\begin{tabular}{@{}lcccccccc@{}}
\toprule
 & \multicolumn{2}{c}{{\bfseries sepsis\_1}} & \multicolumn{2}{c}{{\bfseries sepsis\_2}} & \multicolumn{2}{c}{{\bfseries sepsis\_3}} \\ \cmidrule(lr){2-3} \cmidrule(lr){4-5} \cmidrule(lr){6-7}
method & offline\_total (s) & online\_avg (ms) & offline\_total (s) & online\_avg (ms) & offline\_total (s) & online\_avg (ms) \\ \midrule
single\_laststate & $40.05 \pm 0.15$ & $27 \pm 31$ & $18.68 \pm 0.03$ & $33 \pm 33$ & $81.33 \pm 0.08$ & $29 \pm 31$ \\ 
single\_agg & $39.65 \pm 0.2$ & $29 \pm 33$ & $21.86 \pm 0.15$ & $36 \pm 35$ & $54.24 \pm 0.18$ & $31 \pm 34$ \\ 
knn\_laststate & $2.85 \pm 0.04$ & $54 \pm 58$ & $1.04 \pm 0.04$ & $64 \pm 62$ & $2.08 \pm 0.03$ & $57 \pm 60$ \\ 
knn\_agg & $2.91 \pm 0.04$ & $61 \pm 66$ & $1.07 \pm 0.03$ & $83 \pm 78$ & $1.95 \pm 0.05$ & $68 \pm 70$ \\ 
state\_laststate & $25.72 \pm 0.22$ & $29 \pm 33$ & $15.5 \pm 0.07$ & $35 \pm 36$ & $41.82 \pm 0.17$ & $31 \pm 33$ \\ 
state\_agg & $28.85 \pm 0.13$ & $32 \pm 36$ & $24.5 \pm 0.06$ & $39 \pm 39$ & $61.92 \pm 0.82$ & $34 \pm 37$ \\ 
cluster\_laststate & $27.8 \pm 0.1$ & $\bm{26 \pm 32}$ & $17.28 \pm 0.2$ & $37 \pm 36$ & $59.4 \pm 0.33$ & $32 \pm 35$ \\ 
cluster\_agg & $21.22 \pm 0.14$ & $28 \pm 33$ & $19.71 \pm 0.07$ & $39 \pm 39$ & $64.08 \pm 0.39$ & $33 \pm 37$ \\ 
prefix\_index & $93.4 \pm 1.62$ & $37 \pm 24$ & $23.02 \pm 0.16$ & $41 \pm 26$ & $43.06 \pm 0.29$ & $38 \pm 25$ \\ 
prefix\_laststate & $21.86 \pm 0.12$ & $\bm{26 \pm 29}$ & $23.41 \pm 0.12$ & $\bm{31 \pm 29}$ & $28.9 \pm 0.08$ & $\bm{27 \pm 29}$ \\ 
prefix\_agg & $26.75 \pm 0.15$ & $28 \pm 31$ & $20.73 \pm 0.05$ & $34 \pm 33$ & $28.3 \pm 0.19$ & $29 \pm 31$ \\ 
\bottomrule
 & \multicolumn{2}{c}{{\bfseries bpic2012\_1}} & \multicolumn{2}{c}{{\bfseries bpic2012\_2}} & \multicolumn{2}{c}{{\bfseries bpic2012\_3}} \\ \cmidrule(lr){2-3} \cmidrule(lr){4-5} \cmidrule(lr){6-7}
method & offline\_total (s) & online\_avg (ms) & offline\_total (s) & online\_avg (ms) & offline\_total (s) & online\_avg (ms) \\ \midrule
single\_laststate & $402.09 \pm 3.02$ & $7 \pm 11$ & $181.94 \pm 6.17$ & $7 \pm 10$ & $454.27 \pm 1.31$ & $7 \pm 11$ \\ 
single\_agg & $290.33 \pm 0.99$ & $8 \pm 12$ & $268.54 \pm 2.51$ & $8 \pm 12$ & $268.29 \pm 1.14$ & $8 \pm 12$ \\ 
knn\_laststate & $29.8 \pm 0.58$ & $82 \pm 93$ & $28.27 \pm 0.26$ & $117 \pm 132$ & $34.41 \pm 0.03$ & $156 \pm 171$ \\ 
knn\_agg & $29.86 \pm 0.59$ & $143 \pm 159$ & $30.07 \pm 0.59$ & $403 \pm 434$ & $29.65 \pm 0.55$ & $38 \pm 52$ \\ 
state\_laststate & $234.38 \pm 0.68$ & $8 \pm 10$ & $251.72 \pm 0.8$ & $8 \pm 10$ & $132.18 \pm 0.61$ & $8 \pm 10$ \\ 
state\_agg & $205.54 \pm 4.88$ & $10 \pm 12$ & $640.61 \pm 8.42$ & $10 \pm 12$ & $293.0 \pm 3.62$ & $9 \pm 12$ \\ 
cluster\_laststate & $718.57 \pm 15.83$ & $9 \pm 14$ & $533.58 \pm 2.9$ & $9 \pm 13$ & $141.21 \pm 1.96$ & $10 \pm 15$ \\ 
cluster\_agg & $200.12 \pm 0.45$ & $9 \pm 13$ & $741.83 \pm 19.09$ & $10 \pm 16$ & $264.94 \pm 6.29$ & $10 \pm 16$ \\ 
prefix\_index & $2637.76 \pm 3.59$ & $36 \pm 12$ & $5857.46 \pm 19.9$ & $36 \pm 12$ & $2815.82 \pm 14.87$ & $34 \pm 11$ \\ 
prefix\_laststate & $313.24 \pm 3.13$ & $\bm{4 \pm 3}$ & $253.69 \pm 1.02$ & $\bm{4 \pm 3}$ & $240.8 \pm 1.81$ & $\bm{4 \pm 3}$ \\ 
prefix\_agg & $562.56 \pm 7.46$ & $5 \pm 5$ & $409.94 \pm 4.45$ & $5 \pm 5$ & $223.96 \pm 10.6$ & $5 \pm 5$ \\ 
\bottomrule
 & \multicolumn{2}{c}{{\bfseries bpic2017\_1}} & \multicolumn{2}{c}{{\bfseries bpic2017\_2}} & \multicolumn{2}{c}{{\bfseries bpic2017\_3}} \\ \cmidrule(lr){2-3} \cmidrule(lr){4-5} \cmidrule(lr){6-7}
method & offline\_total (s) & online\_avg (ms) & offline\_total (s) & online\_avg (ms) & offline\_total (s) & online\_avg (ms) \\ \midrule
single\_laststate & $1845.39 \pm 36.22$ & $19 \pm 23$ & $2116.09 \pm 41.53$ & $18 \pm 22$ & $2587.32 \pm 50.78$ & $19 \pm 23$ \\ 
single\_agg & $4569.55 \pm 89.68$ & $19 \pm 24$ & $7042.71 \pm 138.21$ & $21 \pm 26$ & $2021.16 \pm 39.67$ & $20 \pm 25$ \\ 
knn\_laststate & $124.61 \pm 2.45$ & $1476 \pm 1389$ & $125.47 \pm 2.46$ & $1474 \pm 1386$ & $125.41 \pm 2.46$ & $1477 \pm 1390$ \\ 
knn\_agg & $134.2 \pm 2.63$ & $1601 \pm 1504$ & $125.39 \pm 2.46$ & $1488 \pm 1398$ & $125.58 \pm 2.46$ & $1480 \pm 1393$ \\ 
state\_laststate & $1568.31 \pm 30.78$ & $18 \pm 20$ & $2661.92 \pm 66.32$ & $19 \pm 23$ & $2771.55 \pm 54.39$ & $18 \pm 20$ \\ 
state\_agg & $2357.57 \pm 46.27$ & $20 \pm 22$ & $4387.99 \pm 194.51$ & $22 \pm 25$ & $3051.07 \pm 59.88$ & $20 \pm 22$ \\ 
cluster\_laststate & $780.47 \pm 15.32$ & $17 \pm 21$ & $2894.35 \pm 56.8$ & $19 \pm 23$ & $2032.37 \pm 39.89$ & $16 \pm 20$ \\ 
cluster\_agg & $2556.04 \pm 50.16$ & $16 \pm 20$ & $4233.3 \pm 83.08$ & $20 \pm 24$ & $1800.81 \pm 35.34$ & $17 \pm 20$ \\ 
prefix\_index & $19581.63 \pm 384.29$ & $72 \pm 9$ & $15822.79 \pm 310.52$ & $81 \pm 7$ & $17384.94 \pm 341.18$ & $78 \pm 13$ \\ 
prefix\_laststate & $2745.56 \pm 53.88$ & $\bm{14 \pm 15}$ & $1863.41 \pm 36.57$ & $\bm{17 \pm 18}$ & $1660.26 \pm 32.58$ & $\bm{15 \pm 15}$ \\ 
prefix\_agg & $4751.78 \pm 93.25$ & $18 \pm 24$ & $2712.28 \pm 53.23$ & $\bm{17 \pm 17}$ & $2680.83 \pm 52.61$ & $16 \pm 17$ \\ 
\bottomrule
 & \multicolumn{2}{c}{{\bfseries traffic}} & \multicolumn{2}{c}{{\bfseries hospital\_1}} & \multicolumn{2}{c}{{\bfseries hospital\_2}} \\ \cmidrule(lr){2-3} \cmidrule(lr){4-5} \cmidrule(lr){6-7}
method & offline\_total (s) & online\_avg (ms) & offline\_total (s) & online\_avg (ms) & offline\_total (s) & online\_avg (ms) \\ \midrule
single\_laststate & $3169.91 \pm 76.52$ & $62 \pm 34$ & $23191.09 \pm 455.13$ & $380 \pm 249$ & $5303.26 \pm 104.08$ & $410 \pm 270$ \\ 
single\_agg & $4018.67 \pm 82.0$ & $71 \pm 40$ & $5999.77 \pm 6.31$ & $401 \pm 264$ & $8634.78 \pm 169.46$ & $426 \pm 281$ \\ 
knn\_laststate & $400.63 \pm 7.64$ & $96 \pm 60$ & $117.9 \pm 2.31$ & $\bm{54 \pm 37}$ & $320.52 \pm 6.29$ & $\bm{104 \pm 90}$ \\ 
knn\_agg & $444.14 \pm 8.93$ & $158 \pm 101$ & $213.95 \pm 4.2$ & $79 \pm 51$ & $315.68 \pm 6.2$ & $114 \pm 93$ \\ 
state\_laststate & $1088.97 \pm 18.94$ & $74 \pm 43$ & $10056.83 \pm 197.37$ & $312 \pm 257$ & $7321.59 \pm 143.69$ & $293 \pm 235$ \\ 
state\_agg & $828.18 \pm 14.85$ & $75 \pm 41$ & $16417.43 \pm 322.19$ & $392 \pm 238$ & $16783.7 \pm 329.38$ & $363 \pm 212$ \\ 
cluster\_laststate & $2152.68 \pm 3.26$ & $64 \pm 37$ & $11463.75 \pm 224.98$ & $296 \pm 270$ & $4704.15 \pm 92.32$ & $302 \pm 282$ \\ 
cluster\_agg & $1572.57 \pm 3.52$ & $69 \pm 40$ & $3297.8 \pm 64.72$ & $339 \pm 254$ & $9174.11 \pm 180.04$ & $353 \pm 264$ \\ 
prefix\_index & $2895.03 \pm 56.82$ & $102 \pm 13$ & $16114.53 \pm 316.25$ & $930 \pm 136$ & $21000.14 \pm 412.13$ & $960 \pm 220$ \\ 
prefix\_laststate & $2963.04 \pm 3.57$ & $\bm{59 \pm 32}$ & $6756.85 \pm 132.6$ & $380 \pm 265$ & $9208.33 \pm 180.71$ & $323 \pm 219$ \\ 
prefix\_agg & $1669.98 \pm 4.9$ & $62 \pm 34$ & $11395.98 \pm 223.65$ & $399 \pm 241$ & $7993.41 \pm 156.87$ & $353 \pm 204$ \\ 
\bottomrule
\end{tabular}
\end{adjustbox}
\end{center}
\end{table}

In terms of online execution times, the observed patterns are in line with those of other classifiers. However, there are some differences in the offline phase. Namely, in case of RF, the single classifiers perform relatively better as compared to bucketing methods. Furthermore, the difference between the encoding methods becomes more evident, with the last state encodings usually outperforming their aggregation encoding counterparts. The index-based encoding is still the slowest of the techniques. In case of logit, all the methods achieve comparable offline times, except for index-based encoding and the clustering based bucketings, which are slower than the others. In case of SVM, the single\_laststate method tends to be much slower than other techniques. This discussion concludes the answer to RQ3.2.

\subsection{Results: gap-based filtering}
In order to investigate the effects of gap-based filtering on the execution times and the accuracy, we selected 4 methods based on their performance in the above subsections: single\_agg, single\_laststate, prefix\_index, and prefix\_agg. The first three of these methods were shown to take the most time in the offline phase, i.e., they have the most potential to benefit from a filtering technique. Also, single\_agg and prefix\_agg achieved the highest overall AUC scores, which makes them the most attractive candidates to apply in practice. Furthermore, we selected 6 datasets which are representative in terms of their sizes (i.e., number of traces), consist of relatively long traces on average, and did not yield a very high accuracy very early in the trace.

\added{Figures~\ref{fig:offline_time_gaps_xgboost}--\ref{fig:online_time_gaps_xgboost} plot the performance of the classifiers over different gap sizes, i.e. on the $x$-axis, $g=1$ corresponds to no filtering (using prefixes obtained after every event), $g=3$ to using prefixes obtained after every 3rd event, and $g=5$ to prefixes after every 5th event.} In Figure~\ref{fig:offline_time_gaps_xgboost} we can see that using $g=3$ yields an improvement of about 2-3 times in the offline execution times, while using $g=5$, the improvement is usually around 3-4 times, as compared to no filtering ($g=1$). For instance, in case of single\_agg on the \emph{bpic2017\_2} dataset with $g=5$, this means that the offline phase takes about 30 minutes instead of 2 hours. At the same time, the overall AUC remains at the same level, sometimes even increasing when a filtering is applied (Figure~\ref{fig:aucs_gaps_xgboost}). On the other hand, the gap-based filtering only has a marginal (positive) effect on the online execution times, which usually remain on the same level as without filtering (Figure~\ref{fig:online_time_gaps_xgboost}).
This concludes the answer to RQ3.3.

\begin{figure}[hbtp]
\centering
\includegraphics[width=1\textwidth]{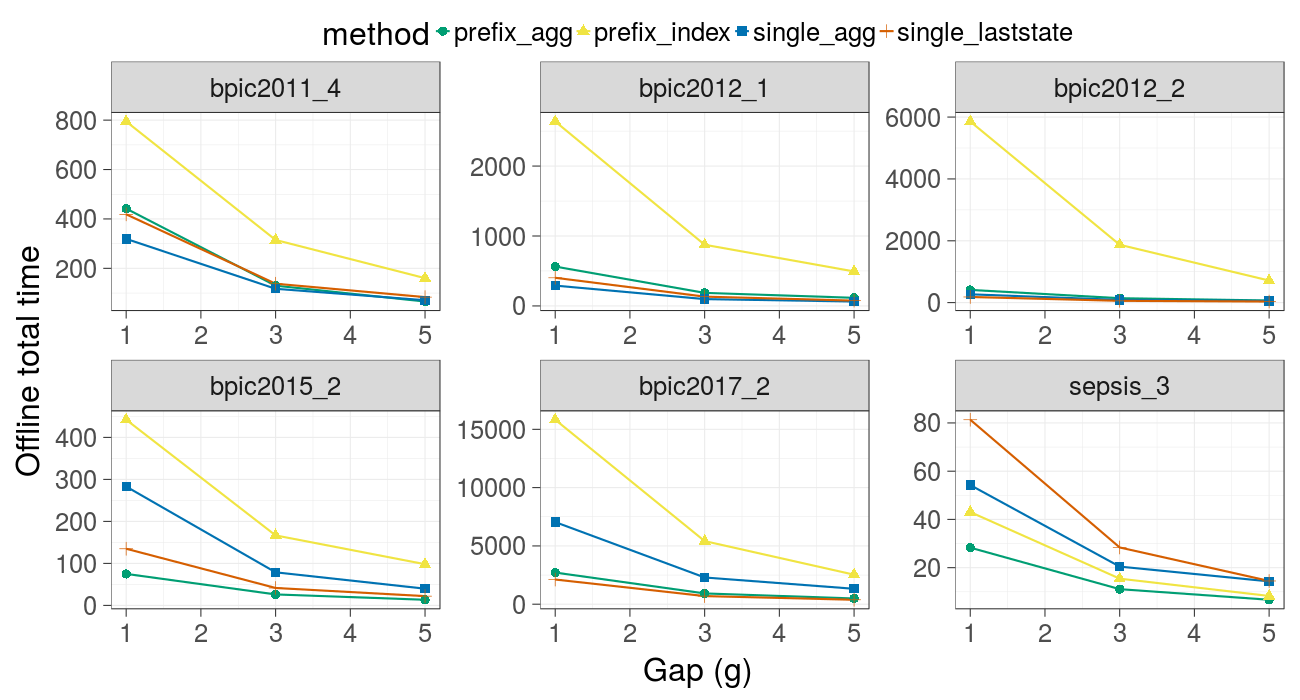}
\caption{Offline times across different gaps (XGBoost)}
\label{fig:offline_time_gaps_xgboost}
\end{figure}

\begin{figure}[hbtp]
\centering
\includegraphics[width=1\textwidth]{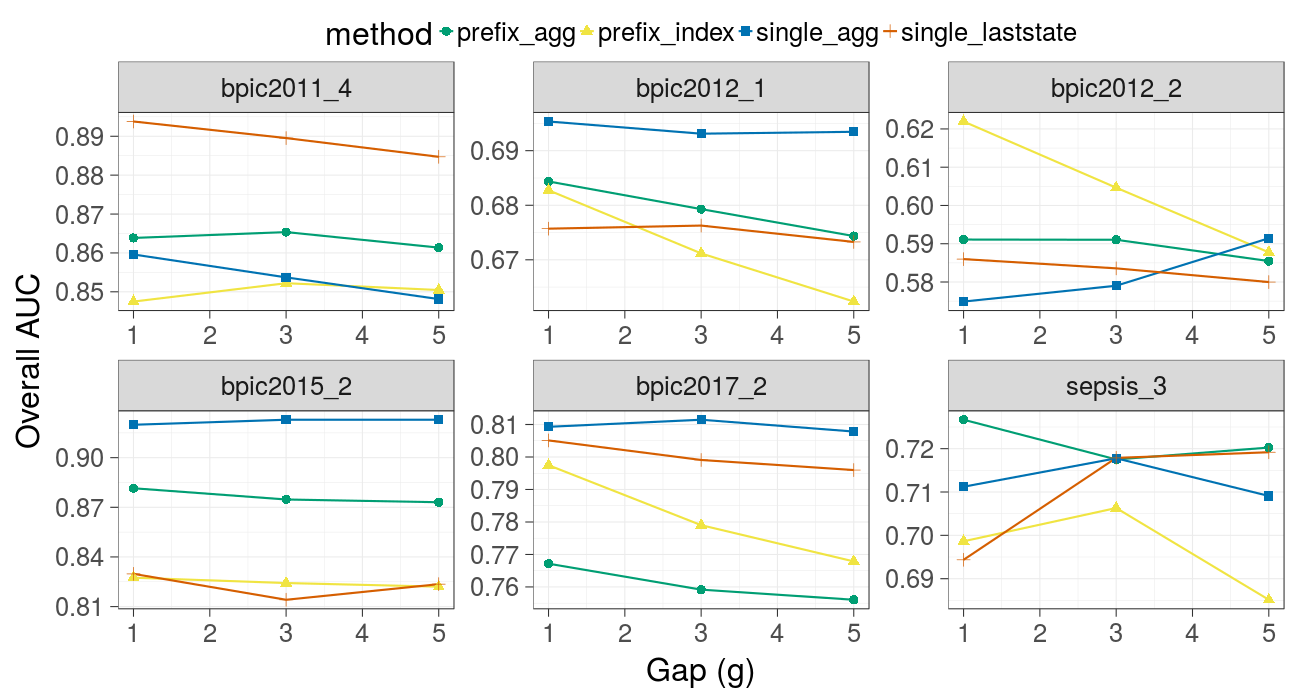}
\caption{AUC across different gaps (XGBoost)}
\label{fig:aucs_gaps_xgboost}
\end{figure}

\begin{figure}[hbtp]
\centering
\includegraphics[width=1\textwidth]{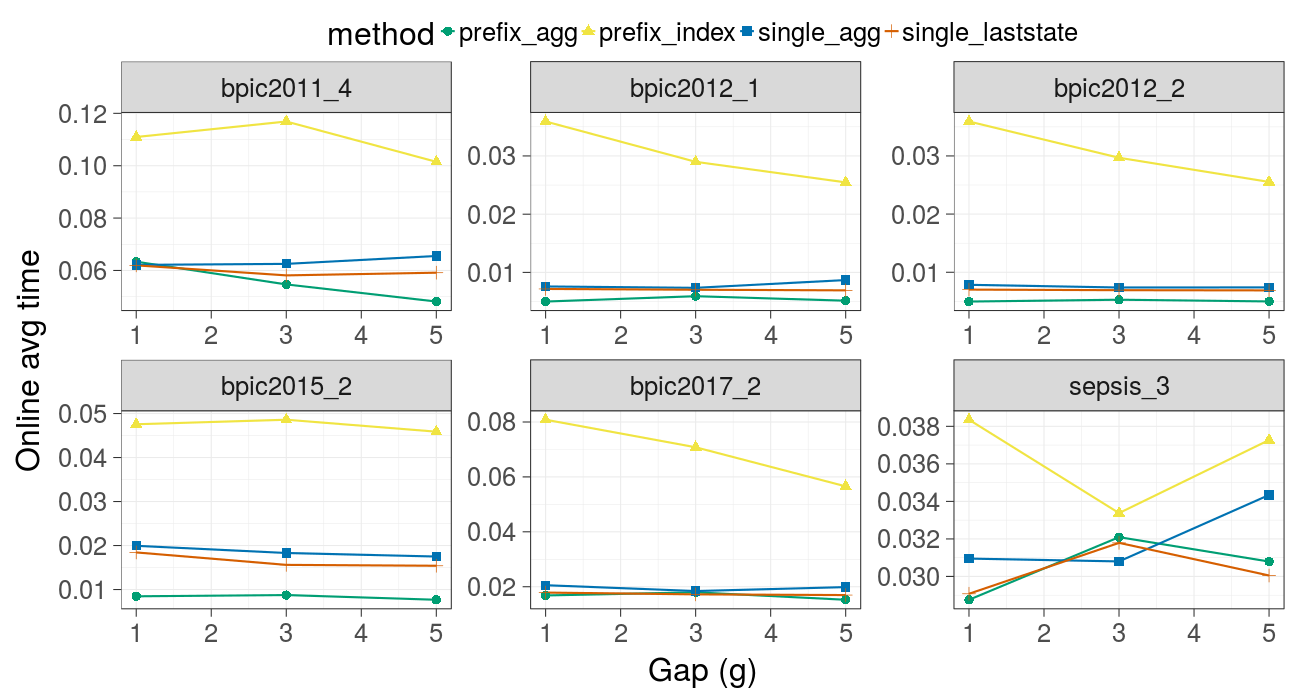}
\caption{Online times across different gaps (XGBoost)}
\label{fig:online_time_gaps_xgboost}
\end{figure}

\subsection{Results: categorical domain filtering}
To answer RQ3.4, we proceed with the 4 methods as discussed in the previous subsection. To better investigate the effect of filtering the categorical attribute levels, we distinguish between the static and the dynamic categorical attributes.
For investigating the effects of dynamic categorical domain filtering, we selected 9 datasets that contain a considerable number of levels in the dynamic categorical attributes.

Both the offline (Figure~\ref{fig:offline_time_dynamic_levels_xgboost}) and the online (Figure~\ref{fig:online_time_dynamic_levels_xgboost}) execution times tend to increase linearly when the proportion of levels is increased. As expected, the prefix\_index method benefits the most from the filtering, since the size of the feature vector increases more rapidly than in the other methods when more levels are added (the vector contains one feature per level per event). Although the overall AUC is negatively affected by the filtering of levels (see Figure~\ref{fig:aucs_dynamic_levels_xgboost}), reasonable tradeoffs can still be found. For instance, when using 50\% of the levels in case of single\_agg on the \emph{hospital\_2} dataset, the AUC is almost unaffected, while the training time has decreased by more than 30 minutes and the online execution times have decreased by a half.

\begin{figure}[hbtp]
\centering
\includegraphics[width=1\textwidth]{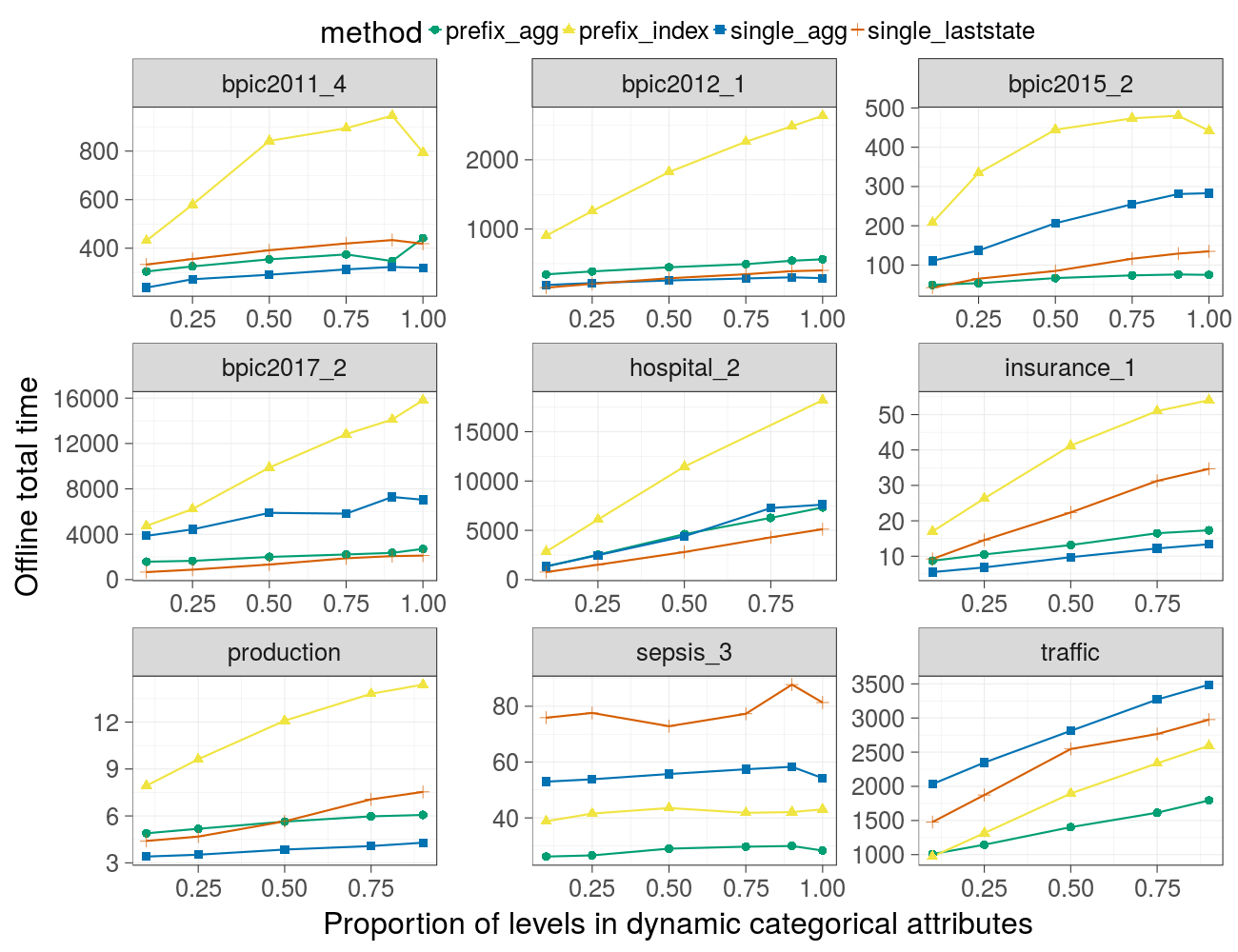}
\caption{Offline times across different filtering proportions of {\bfseries dynamic}  categorical attribute levels (XGBoost)}
\label{fig:offline_time_dynamic_levels_xgboost}
\end{figure}

\begin{figure}[hbtp]
\centering
\includegraphics[width=1\textwidth]{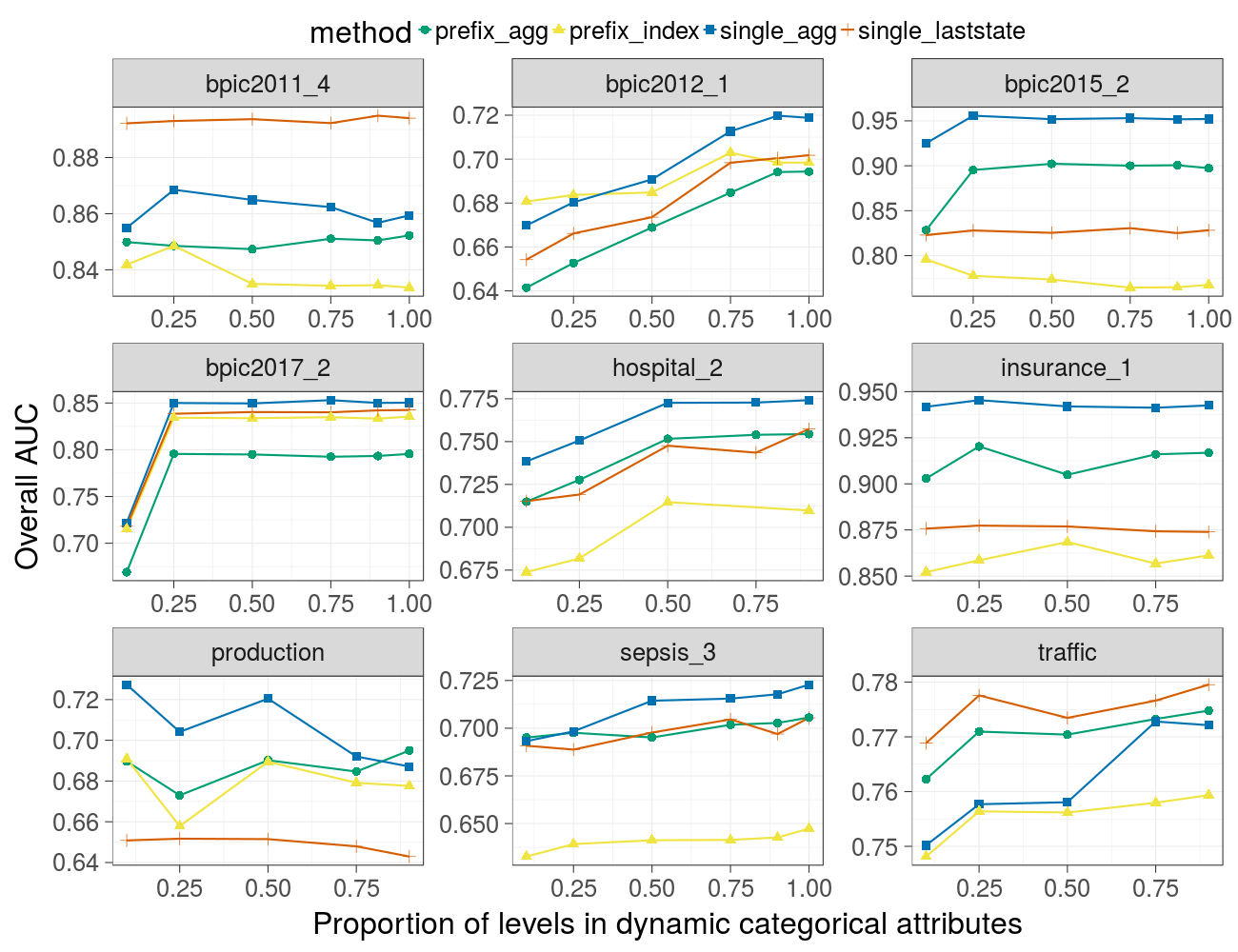}
\caption{AUC across different filtering proportions of {\bfseries dynamic} categorical attribute levels (XGBoost)}
\label{fig:aucs_dynamic_levels_xgboost}
\end{figure}

\begin{figure}[hbtp]
\centering
\includegraphics[width=1\textwidth]{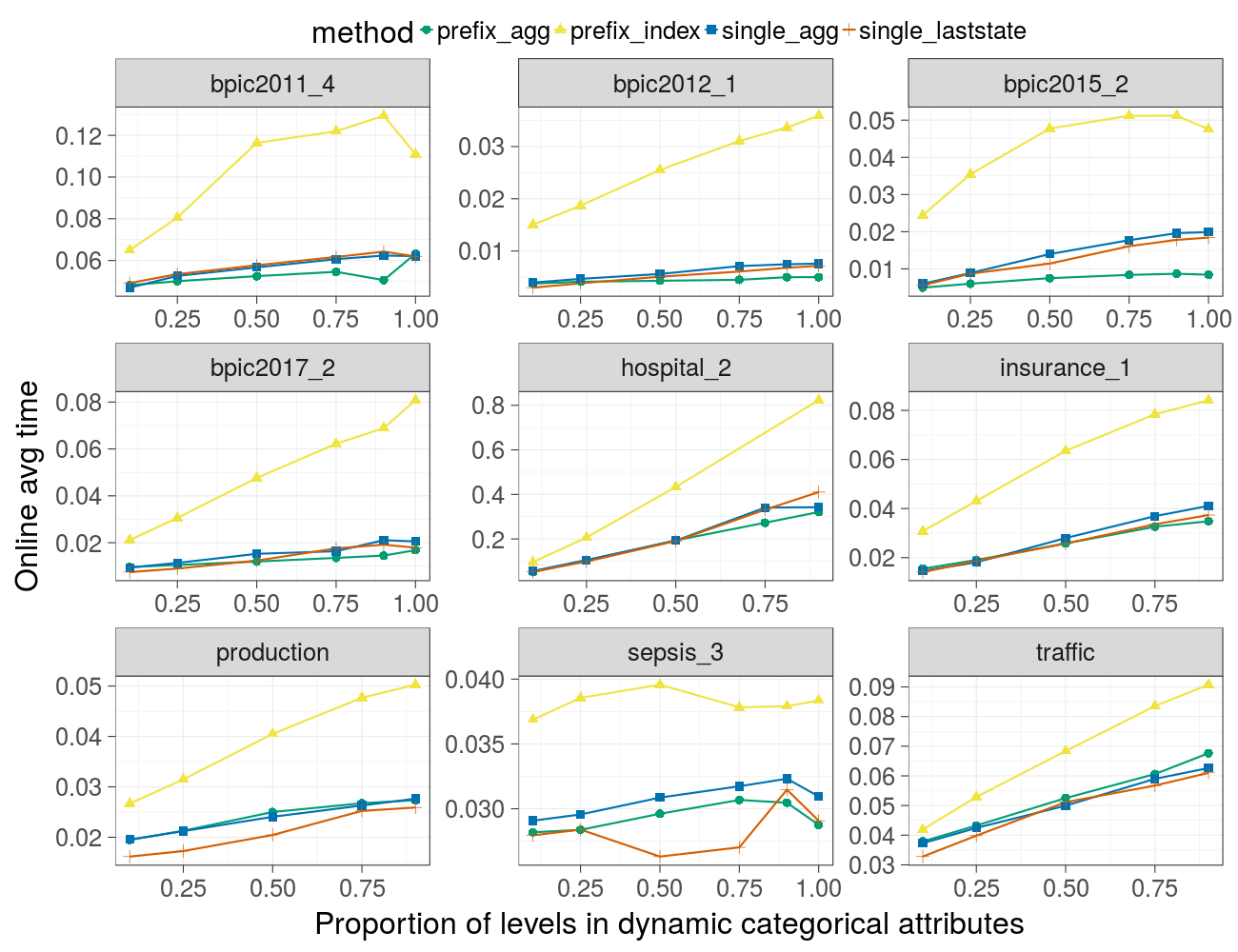}
\caption{Online times across different filtering proportions of {\bfseries dynamic}  categorical attribute levels (XGBoost)}
\label{fig:online_time_dynamic_levels_xgboost}
\end{figure}

We performed similar experiments by filtering the static categorical attribute domains, selecting 6 datasets that contain a considerable number of levels in these attributes. However, the improvement in execution times were marginal compared to those obtained when using dynamic attribute filtering (see Figures~\ref{fig:offline_time_static_levels_xgboost}-\ref{fig:aucs_static_levels_xgboost} in Appendix). This is natural, since the static attributes have a smaller effect on the size of the feature vector (each level occurs in the vector only once).
This concludes the answer to RQ3.4.

\section{Threats to validity}
\label{sec:threats}

One of the threats to the validity of this study relates to the potential selection bias in the literature review. To minimize this, we described our systematic literature review procedure on a level of detail that is sufficient to replicate the search. However, in time the search and ranking algorithms of the used academic database (Google Scholar) might be updated and return different results. Another potential source of bias is the subjectivity when applying inclusion and exclusion criteria, as well as when determining the primary and subsumed studies. In order to alleviate this issue, all the included papers were collected in a publicly available spreadsheet, together with decisions and reasons about excluding them from the study. Moreover, each paper was independently assessed against the inclusion and exclusion criteria by two authors, and inconsistencies were resolved with the mediation of a third author.  

Another threat to validity is related to the comprehensiveness of the conducted experiments. In particular, only one clustering method was tested, a single state abstraction was used when building the transition systems for state-based bucketing, and four classification algorithms were applied. It is possible that there exists, for example, a combination of an untested clustering technique and a classifier that outperforms the settings used in this study. Also, although the hyperparameters were optimized using a state-of-the-art hyperparameter optimization technique, it is possible that using more iterations for optimization or a different optimization algorithm, other parameter settings would be found that outperform the settings used in the current evaluation. 
Furthermore, the generalizability of the findings is to some extent limited by the fact that the experiments were performed on a limited number of prediction tasks (24), constructed from nine event logs. Although these are all real-life event logs from different application fields that exhibit different characteristics, it may be possible that the results would be different using other datasets or different log preprocessing techniques for the same datasets. In order to mitigate these threats, we built an open-source software framework which allows the full replication of the experiments, and made this tool publicly available. Moreover, additional datasets, as well as new sequence classification and encoding methods can be plugged in. So the framework can be used for future experiments. Also, the preprocessed datasets constructed from the three publicly available event logs are included together with the tool implementation in order to enhance the reproducibility of the experiments.


\section{Conclusion}
\label{sec:conclusion}

This study provided a survey and comparative analysis and evaluation of existing outcome-oriented predictive business process monitoring techniques. The relevant existing studies were identified through a systematic literature review (SLR), which revealed 14 studies (some described across multiple papers) dealing with the problem of predicting case outcomes. Out of these, seven were considered to contain a distinct contribution (primary studies). Through further analysis of the primary studies, a taxonomy was proposed based on two main aspects, the trace bucketing approach and sequence encoding method employed. Combinations of these two aspects led to a total of 11 distinct methods.

The studies were characterized from different perspectives, resulting in a taxonomy of existing techniques. Finally, a comparative evaluation of the 11 identified techniques was performed using a unified experimental set-up and 24 predictive monitoring tasks constructed from 9 real-life event logs. To ensure a fair evaluation, all the selected techniques were implemented as a publicly available consolidated framework, which is designed to incorporate additional datasets and methods.

The results of the benchmark show that the most reliable and accurate results (in terms of AUC) are obtained using a lossy (aggregation) encoding of the sequence, e.g., the frequencies of performed activities rather than the ordered activities. One of the main benefits of this encoding is that it enables to represent all prefix traces, regardless of their length, in the same number of features. This way, a single classifier can be trained over all of the prefix traces, allowing the classifier to derive meaningful patterns by itself. These results disprove the existing opinion in the literature about the superiority of a lossless encoding of the trace (index-based encoding) that requires prefixes to be divided into buckets according to their length, while multiple classifiers are trained on each such subset of prefixes.

\added{The study also put into evidence the importance of checking for concept drifts when applying predictive monitoring methods. In the study, we found concept drifts in the data attributes extracted from two datasets, and in both cases, these drifts significantly affected the performance of all tested methods. This observation is aligned with previous studies in the field of process mining, which have shown that concept drifts are common in the control flow of business processes~\cite{Ostovar2016,Maaradji2017,Ostovar2017}. Techniques for automated detection and characterization of process control-flow drifts from event logs and event streams are available~\cite{Ostovar2016,Maaradji2017,Ostovar2017}. Researchers and practitioners using predictive monitoring methods should consider applying these detection methods, as well as standard statistical tests on the features extracted, to ensure that there is no drift present, which could affect the performance of the predictive models.}

The study paves the way to several directions of future work.
In Section~\ref{sec:bckg} we noted that case and event attributes can be of categorical, numeric or textual type. The systematic review showed that existing methods are focused on handling categorical and numeric attributes, to the exclusion of textual ones. Recent work has shown how text mining techniques can be used to extend the index-based encoding approach of~\cite{teinemaa2016predictive} in order to handle text attributes, however this latter work considered a reduced set of text mining techniques and has only been tested on two datasets of relatively small size and complexity.


Secondly, the methods identified in the survey are mainly focused on extracting features from one trace at a time (i.e., intra-case features), while only a single inter-case feature (the number of open cases) is included. However, due to the fact that the ongoing cases of a process share the same pool of resources, the outcome of a case may depend also on other aspects of the current state of the rest of ongoing cases in the process. Therefore, the accuracy of the models tested in this benchmark could be further improved by using a larger variety of inter-case features.

Lastly, as long-short term memory (LSTM) networks have recently gained attention in predicting remaining time and next activity of a running case of a business process~\cite{EvermannRF16,TaxVRD17}, another natural direction for future work is to study how LSTMs can be used for outcome prediction. In particular, could LSTMs automatically derive relevant features from collections of trace prefixes, and thus obviate the need for sophisticated feature engineering (aggregation functions), which has been so far the focus of predictive process monitoring research?

\begin{acks}
This research is funded by the~\grantsponsor{sponsor1}{Australian Research Council}{} (grant~\grantnum[]{sponsor1}{DP150103356}), the ~\grantsponsor{sponsor2}{Estonian Research Council}{} (grant~\grantnum[]{sponsor2}{IUT20-55}) and~\grantsponsor{sponsor3} Euuropean Regional Development Fund (Dora Plus Program)
\end{acks}

\bibliographystyle{ACM-Reference-Format}
\bibliography{bibliography}

\appendix
\section*{Appendix}
\label{appendix}

This Appendix reports the following:
\begin{itemize}
\item The distributions of case lengths in different outcome classes (Figures~\ref{fig:case_length_hist_3cols_1}-\ref{fig:case_length_hist_3cols_2});
\item The optimal number of clusters (Table~\ref{table:optimal_params_n_clusters}) and the optimal number of neighbors for KNN approaches (Table~\ref{table:optimal_params_n_neighbors}) found for each classifier;
\item The distributions of bucket sizes for the different bucketing methods (Figures~\ref{fig:bucket_size_density_plots_1}-\ref{fig:bucket_size_density_plots_2});
\item The overall AUC \added{and F-score} values for RF (Table~\ref{table:avg_results_rf}), logit (Table~\ref{table:avg_results_logit}), and SVM (Table~\ref{table:avg_results_svm});
\item The AUC scores across prefix lengths using XGBoost classifier and all of the compared methods (Figures~\ref{fig:results_all_xgboost_1}-\ref{fig:results_all_xgboost_2});
\item The AUC scores across prefix lengths, including long traces only, using the XGBoost classifier (Figure~\ref{fig:results_long_traces});
\item \added{Concept drift in the \emph{bpic2011\_4} log. (Figure~\ref{fig:bpic2011_f4_boxplots});}
\item \added{Concept drift in the \emph{sepsis\_1} log. (Figure~\ref{fig:sepsis_cases_1_boxplots});}
\item The execution times for RF (Tables~\ref{table:performance_results_rf_1}-\ref{table:performance_results_rf_2}), logit(Tables~\ref{table:performance_results_logit_1}-\ref{table:performance_results_logit_2}), and SVM(Tables~\ref{table:performance_results_svm_1}-\ref{table:performance_results_svm_2});
\item The offline (Figure~\ref{fig:offline_time_static_levels_xgboost}) and online (Figure~\ref{fig:online_time_static_levels_xgboost}) execution times and the overall AUC scores (Figure~\ref{fig:aucs_static_levels_xgboost}) when filtering the static categorical attribute domain, using the XGBoost classifier.
\end{itemize}

\begin{figure}[hbtp!]
\centering
\includegraphics[width=0.9\textwidth]{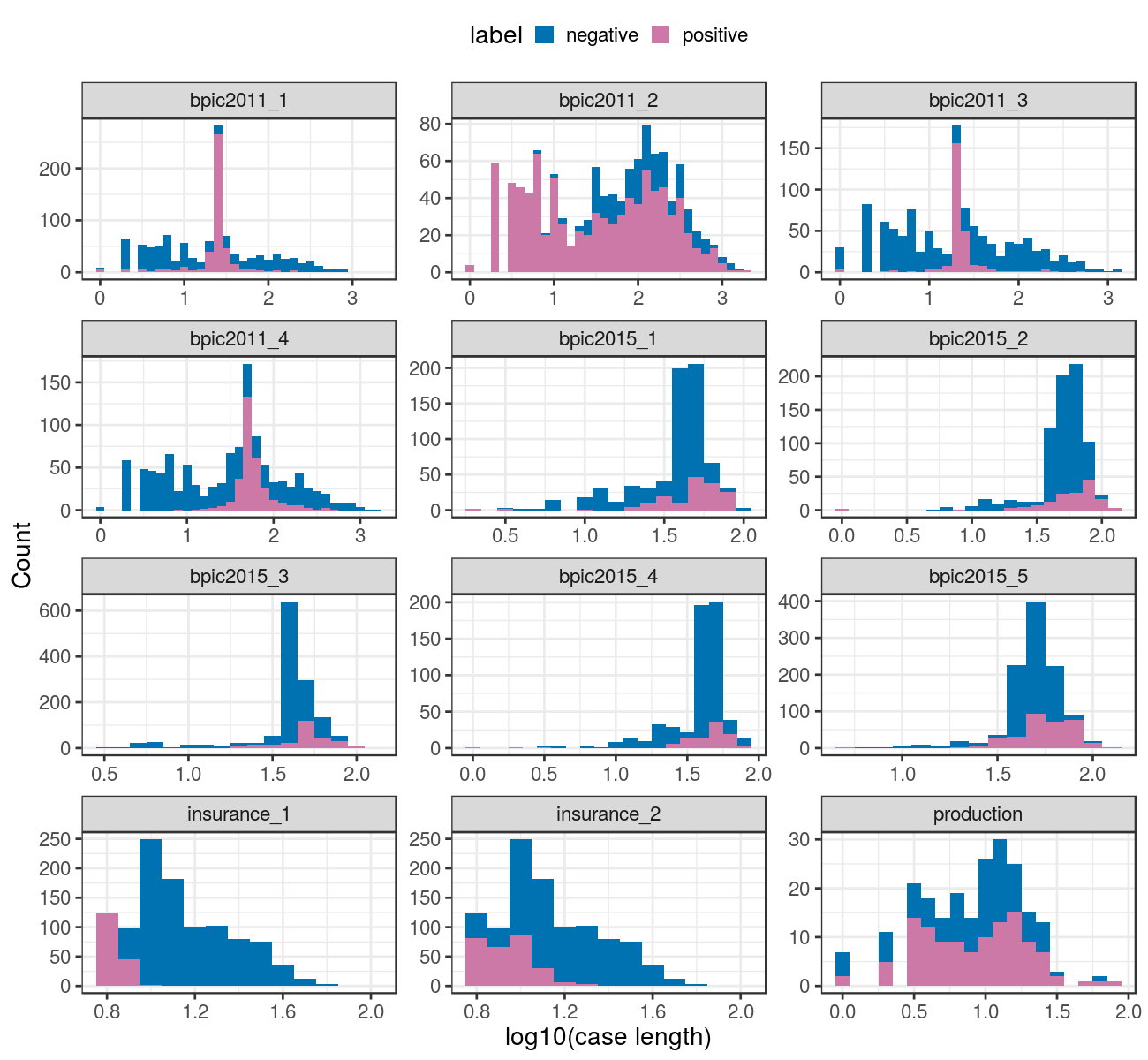}
\caption{Case length histograms for positive and negative classes \deleted{(1)}}
\label{fig:case_length_hist_3cols_1}
\end{figure}

\begin{figure}[hbtp!]
\centering
\includegraphics[width=0.9\textwidth]{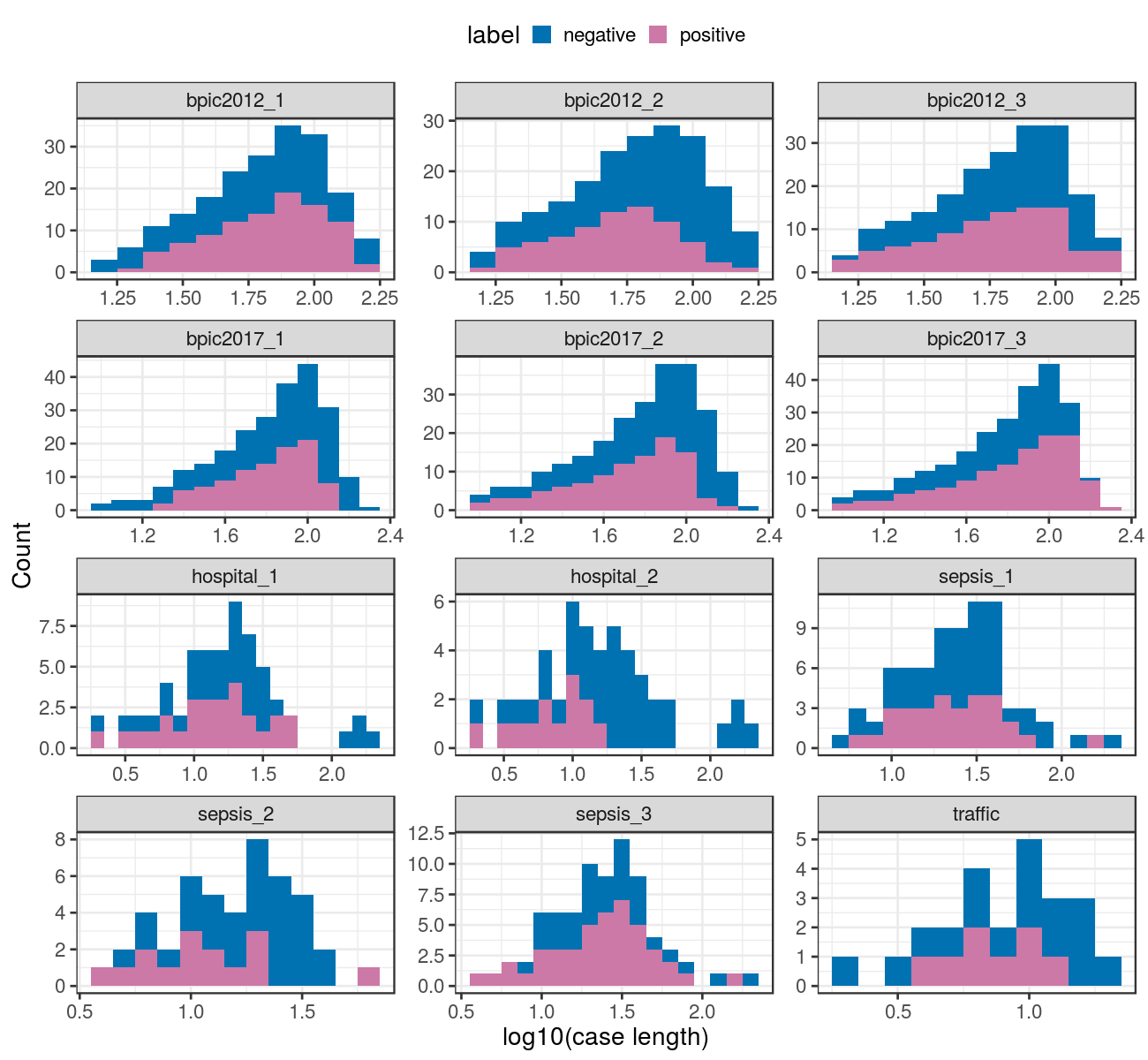}
\caption{Case length histograms for positive and negative classes (\replaced{continued}{2})}
\label{fig:case_length_hist_3cols_2}
\end{figure}

\begin{table}[hbtp!]
\caption{Best number of clusters}
\label{table:optimal_params_n_clusters}
\begin{center}
\begin{adjustbox}{max width=\textwidth}
\begin{tabular}{@{}lcccccccc@{}}
\toprule & \multicolumn{2}{c}{RF} & \multicolumn{2}{c}{XGBoost} & \multicolumn{2}{c}{Logit} & \multicolumn{2}{c}{SVM} \\ 
 dataset & cluster\_last & cluster\_agg & cluster\_last & cluster\_agg & cluster\_last & cluster\_agg & cluster\_last & cluster\_agg \\ \midrule
bpic2011\_1 & 10 & 8 & 10 & 6 & 24 & 23 & 15 & 43 \\
bpic2011\_2 & 28 & 4 & 3 & 6 & 20 & 13 & 27 & 24 \\
bpic2011\_3 & 30 & 4 & 28 & 4 & 33 & 13 & 32 & 44 \\
bpic2011\_4 & 2 & 21 & 2 & 2 & 16 & 2 & 24 & 36 \\
insurance\_2 & 8 & 12 & 2 & 2 & 4 & 3 & 30 & 25 \\
insurance\_1 & 6 & 18 & 3 & 2 & 10 & 47 & 45 & 3 \\
bpic2015\_1 & 39 & 10 & 37 & 4 & 21 & 2 & 13 & 7 \\
bpic2015\_2 & 32 & 6 & 31 & 5 & 42 & 7 & 9 & 13 \\
bpic2015\_3 & 44 & 12 & 36 & 10 & 41 & 11 & 11 & 13 \\
bpic2015\_4 & 45 & 3 & 47 & 5 & 47 & 40 & 19 & 8 \\
bpic2015\_5 & 43 & 4 & 49 & 19 & 32 & 4 & 8 & 4 \\
production & 44 & 21 & 18 & 2 & 38 & 44 & 10 & 7 \\
sepsis\_1 & 38 & 14 & 19 & 6 & 39 & 41 & 9 & 29 \\
sepsis\_2 & 3 & 8 & 4 & 2 & 7 & 3 & 10 & 21 \\
sepsis\_3 & 2 & 7 & 13 & 7 & 7 & 23 & 7 & 3 \\
bpic2012\_1 & 22 & 7 & 3 & 35 & 3 & 3 & 8 & 49 \\
bpic2012\_2 & 9 & 9 & 3 & 4 & 7 & 9 & 15 & 3 \\
bpic2012\_3 & 10 & 26 & 3 & 2 & 13 & 8 & 22 & 15 \\
bpic2017\_1 & 39 & 30 & 22 & 43 & 4 & 34 & 39 & 19 \\
bpic2017\_2 & 11 & 10 & 20 & 15 & 31 & 27 & 40 & 4 \\
bpic2017\_3 & 29 & 30 & 32 & 34 & 19 & 47 & 21 & 35 \\
traffic & 42 & 43 & 29 & 23 & 42 & 36 & 9 & 13 \\
hospital\_1 & 35 & 2 & 33 & 48 & 10 & 8 & 48 & 32 \\
hospital\_2 & 19 & 48 & 33 & 45 & 11 & 8 & 34 & 28 \\
\bottomrule
\end{tabular}
\end{adjustbox}
\end{center}
\end{table}

\begin{table}[hbtp!]
\caption{Best number of neighbors}
\label{table:optimal_params_n_neighbors}
\begin{center}
\begin{adjustbox}{max width=\textwidth}
\begin{tabular}{@{}lcccccccc@{}}
\toprule & \multicolumn{2}{c}{RF} & \multicolumn{2}{c}{XGBoost} & \multicolumn{2}{c}{Logit} & \multicolumn{2}{c}{SVM} \\ 
 dataset & knn\_last & knn\_agg & knn\_last & knn\_agg & knn\_last & knn\_agg & knn\_last & knn\_agg \\ \midrule
bpic2011\_1 & 47 & 45 & 50 & 50 & 46 & 48 & 49 & 39 \\
bpic2011\_2 & 45 & 47 & 50 & 46 & 26 & 21 & 42 & 40 \\
bpic2011\_3 & 50 & 46 & 50 & 46 & 45 & 32 & 44 & 42 \\
bpic2011\_4 & 40 & 41 & 43 & 46 & 44 & 50 & 16 & 32 \\
insurance\_2 & 46 & 47 & 50 & 45 & 48 & 49 & 32 & 44 \\
insurance\_1 & 45 & 49 & 44 & 50 & 29 & 36 & 16 & 12 \\
bpic2015\_1 & 31 & 49 & 50 & 45 & 32 & 17 & 12 & 3 \\
bpic2015\_2 & 48 & 50 & 46 & 46 & 41 & 12 & 11 & 2 \\
bpic2015\_3 & 29 & 48 & 46 & 46 & 40 & 49 & 2 & 3 \\
bpic2015\_4 & 30 & 43 & 50 & 36 & 13 & 9 & 3 & 38 \\
bpic2015\_5 & 30 & 37 & 46 & 50 & 27 & 47 & 2 & 2 \\
production & 10 & 19 & 19 & 16 & 46 & 14 & 15 & 21 \\
sepsis\_1 & 50 & 49 & 47 & 32 & 32 & 26 & 32 & 43 \\
sepsis\_2 & 50 & 41 & 47 & 49 & 47 & 49 & 46 & 39 \\
sepsis\_3 & 47 & 48 & 50 & 50 & 32 & 50 & 49 & 29 \\
bpic2012\_1 & 2 & 50 & 9 & 17 & 6 & 45 & 37 & 33 \\
bpic2012\_2 & 50 & 50 & 14 & 50 & 3 & 42 & 32 & 39 \\
bpic2012\_3 & 50 & 50 & 19 & 3 & 9 & 25 & 22 & 22 \\
bpic2017\_1 & 50 & 50 & 50 & 50 & 50 & 50 & 4 & 50 \\
bpic2017\_2 & 50 & 50 & 50 & 50 & 50 & 50 & 50 & 50 \\
bpic2017\_3 & 50 & 50 & 50 & 50 & 50 & 50 & 50 & 50 \\
traffic & 50 & 50 & 14 & 25 & 10 & 10 & 31 & 42 \\
hospital\_1 & 50 & 50 & 26 & 22 & 29 & 50 & 38 & 3 \\
hospital\_2 & 50 & 50 & 50 & 50 & 36 & 6 & 31 & 24 \\
\bottomrule
\end{tabular}
\end{adjustbox}
\end{center}
\end{table}

\begin{figure}[hbtp!]
\centering
\includegraphics[width=1\textwidth]{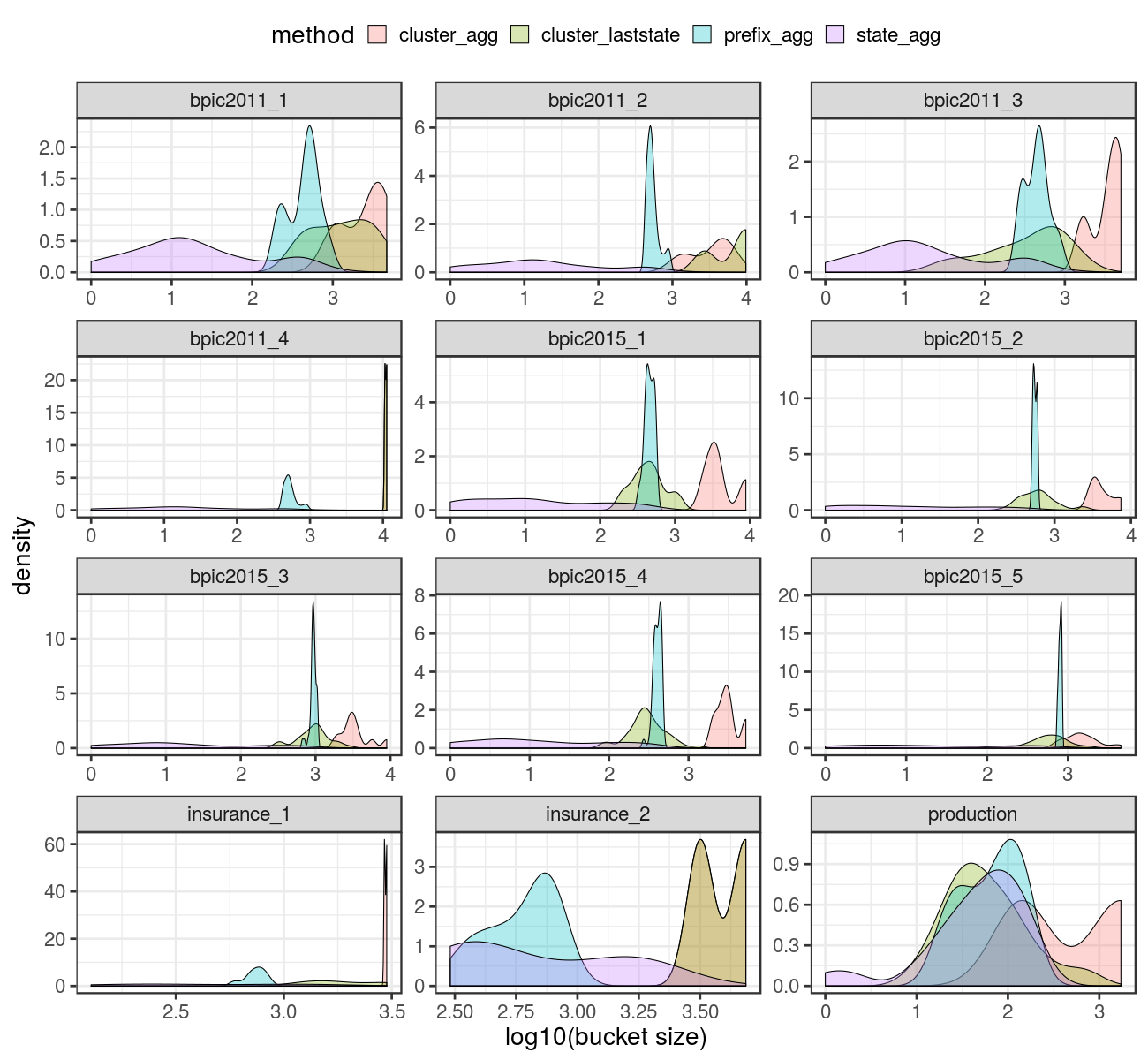}
\caption{Bucket size distributions \deleted{(1)}}
\label{fig:bucket_size_density_plots_1}
\end{figure}

\begin{figure}[hbtp!]
\centering
\includegraphics[width=1\textwidth]{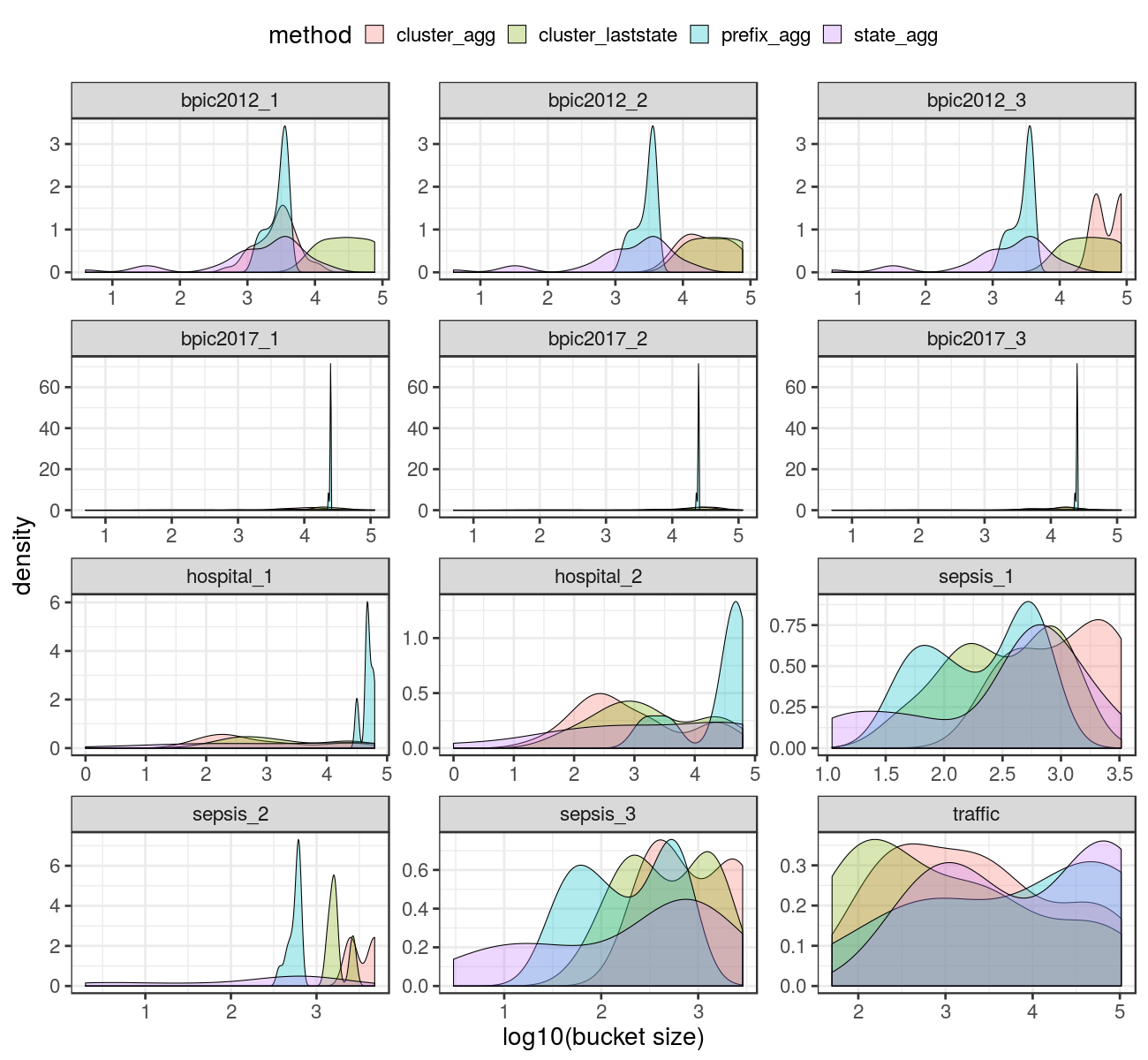}
\caption{Bucket size distributions (\replaced{continued}{2})}
\label{fig:bucket_size_density_plots_2}
\end{figure}

\begin{table}[hbtp]
\caption{Overall AUC \added{(F-score)} for {\bfseries random forest}}
\vspace{-\baselineskip}
\label{table:avg_results_rf}
\begin{center}
\begin{adjustbox}{max width=\textwidth}
\begin{tabular}{@{}lcccccc@{}}
\toprule
 & {\bfseries bpic2011\_1} & {\bfseries bpic2011\_2} & {\bfseries bpic2011\_3} & {\bfseries bpic2011\_4} & {\bfseries insurance\_1} & {\bfseries insurance\_2}
\\ \midrule
single\_laststate & $0.87$ $(0.73)$ & $0.92$ $(0.83)$ & $0.94$ $(0.79)$ & $\bm{0.9}$ $(\bm{0.82})$ & $0.88$ $(0.53)$ & $0.83$ $(0.47)$ \\
single\_agg & $\bm{0.94}$ $(0.86)$ & $\bm{0.98}$ $(\bm{0.95})$ & $\bm{0.98}$ $(\bm{0.94})$ & $0.89$ $(0.8)$ & $\bm{0.91}$ $(\bm{0.65})$ & $0.82$ $(0.48)$ \\
knn\_laststate & $0.92$ $(0.85)$ & $0.96$ $(0.92)$ & $0.92$ $(0.85)$ & $0.79$ $(0.7)$ & $0.85$ $(0.55)$ & $0.77$ $(0.5)$ \\
knn\_agg & $0.87$ $(0.8)$ & $0.94$ $(0.9)$ & $0.9$ $(0.81)$ & $0.78$ $(0.7)$ & $0.87$ $(0.63)$ & $0.79$ $(0.54)$ \\
state\_laststate & $0.88$ $(0.74)$ & $0.92$ $(0.86)$ & $0.94$ $(0.76)$ & $0.88$ $(0.8)$ & $0.88$ $(0.55)$ & $0.84$ $(\bm{0.61})$ \\
state\_agg & $0.92$ $(0.85)$ & $0.96$ $(0.93)$ & $0.97$ $(0.87)$ & $0.87$ $(0.77)$ & $0.9$ $(0.63)$ & $\bm{0.85}$ $(0.59)$ \\
cluster\_laststate & $0.9$ $(0.75)$ & $0.91$ $(0.87)$ & $\bm{0.98}$ $(0.92)$ & $0.88$ $(0.79)$ & $0.89$ $(0.54)$ & $0.82$ $(0.46)$ \\
cluster\_agg & $0.91$ $(0.82)$ & $0.96$ $(0.94)$ & $\bm{0.98}$ $(0.93)$ & $0.89$ $(0.79)$ & $0.9$ $(0.61)$ & $0.82$ $(0.58)$ \\
prefix\_index & $0.93$ $(0.8)$ & $0.96$ $(0.88)$ & $0.97$ $(0.73)$ & $0.85$ $(0.76)$ & $0.88$ $(0.56)$ & $0.79$ $(0.37)$ \\
prefix\_laststate & $0.9$ $(0.75)$ & $0.94$ $(0.87)$ & $0.97$ $(0.72)$ & $0.88$ $(0.78)$ & $0.87$ $(0.5)$ & $0.84$ $(0.57)$ \\
prefix\_agg & $\bm{0.94}$ $(\bm{0.88})$ & $0.97$ $(0.94)$ & $\bm{0.98}$ $(0.78)$ & $0.88$ $(0.78)$ & $0.9$ $(0.59)$ & $0.83$ $(0.58)$ \\
\midrule
 & {\bfseries bpic2015\_1} & {\bfseries bpic2015\_2} & {\bfseries bpic2015\_3} & {\bfseries bpic2015\_4} & {\bfseries bpic2015\_5} & {\bfseries production}
\\ \midrule
single\_laststate & $0.83$ $(0.43)$ & $0.84$ $(0.5)$ & $0.76$ $(0.49)$ & $0.83$ $(0.51)$ & $0.84$ $(0.59)$ & $0.65$ $(0.55)$ \\
single\_agg & $\bm{0.88}$ $(\bm{0.73})$ & $\bm{0.91}$ $(\bm{0.74})$ & $\bm{0.91}$ $(0.72)$ & $0.86$ $(0.62)$ & $0.88$ $(0.76)$ & $0.63$ $(0.54)$ \\
knn\_laststate & $0.8$ $(0.44)$ & $0.87$ $(0.63)$ & $0.87$ $(0.67)$ & $\bm{0.88}$ $(0.56)$ & $0.85$ $(0.71)$ & $0.62$ $(0.55)$ \\
knn\_agg & $0.78$ $(0.53)$ & $0.85$ $(0.63)$ & $0.87$ $(0.73)$ & $0.87$ $(\bm{0.67})$ & $0.85$ $(0.69)$ & $0.66$ $(0.58)$ \\
state\_laststate & $0.75$ $(0.52)$ & $0.84$ $(0.57)$ & $0.84$ $(0.58)$ & $0.85$ $(0.57)$ & $0.86$ $(0.68)$ & $0.62$ $(0.56)$ \\
state\_agg & $0.79$ $(0.64)$ & $0.87$ $(0.69)$ & $0.89$ $(\bm{0.74})$ & $0.87$ $(0.66)$ & $0.88$ $(0.75)$ & $0.68$ $(0.58)$ \\
cluster\_laststate & $0.75$ $(0.43)$ & $0.86$ $(0.61)$ & $0.85$ $(0.67)$ & $0.87$ $(0.63)$ & $\bm{0.89}$ $(0.74)$ & $0.6$ $(0.53)$ \\
cluster\_agg & $0.85$ $(0.69)$ & $0.89$ $(0.73)$ & $\bm{0.91}$ $(0.71)$ & $0.87$ $(0.63)$ & $0.88$ $(0.76)$ & $0.74$ $(\bm{0.66})$ \\
prefix\_index & $0.82$ $(0.52)$ & $0.85$ $(0.46)$ & $0.89$ $(0.67)$ & $0.85$ $(0.59)$ & $0.86$ $(0.68)$ & $\bm{0.76}$ $(0.57)$ \\
prefix\_laststate & $0.75$ $(0.37)$ & $0.84$ $(0.4)$ & $0.79$ $(0.44)$ & $0.85$ $(0.38)$ & $0.84$ $(0.61)$ & $0.72$ $(0.58)$ \\
prefix\_agg & $0.82$ $(0.65)$ & $0.87$ $(0.7)$ & $0.9$ $(\bm{0.74})$ & $0.87$ $(\bm{0.67})$ & $0.88$ $(\bm{0.77})$ & $0.69$ $(0.59)$ \\
\midrule
 & {\bfseries sepsis\_1} & {\bfseries sepsis\_2} & {\bfseries sepsis\_3} & {\bfseries bpic2012\_1} & {\bfseries bpic2012\_2} & {\bfseries bpic2012\_3}
\\ \midrule
single\_laststate & $\bm{0.49}$ $(0.01)$ & $0.78$ $(\bm{0.45})$ & $0.67$ $(0.3)$ & $0.67$ $(0.63)$ & $0.6$ $(0.11)$ & $0.69$ $(\bm{0.42})$ \\
single\_agg & $0.41$ $(0.0)$ & $0.79$ $(0.39)$ & $0.66$ $(0.34)$ & $\bm{0.69}$ $(\bm{0.64})$ & $\bm{0.61}$ $(0.17)$ & $\bm{0.7}$ $(\bm{0.42})$ \\
knn\_laststate & $0.46$ $(\bm{0.05})$ & $0.77$ $(0.26)$ & $0.64$ $(0.16)$ & $0.59$ $(0.59)$ & $0.58$ $(0.11)$ & $0.66$ $(0.41)$ \\
knn\_agg & $0.48$ $(0.04)$ & $0.74$ $(0.2)$ & $0.61$ $(0.17)$ & $0.64$ $(0.59)$ & $0.57$ $(0.14)$ & $0.67$ $(\bm{0.42})$ \\
state\_laststate & $0.46$ $(0.0)$ & $0.79$ $(0.41)$ & $\bm{0.71}$ $(0.26)$ & $0.68$ $(0.63)$ & $0.6$ $(0.16)$ & $0.69$ $(0.4)$ \\
state\_agg & $0.43$ $(0.0)$ & $0.79$ $(0.44)$ & $0.7$ $(0.3)$ & $0.68$ $(\bm{0.64})$ & $0.59$ $(0.17)$ & $\bm{0.7}$ $(\bm{0.42})$ \\
cluster\_laststate & $0.48$ $(0.01)$ & $\bm{0.8}$ $(0.41)$ & $\bm{0.71}$ $(\bm{0.35})$ & $0.66$ $(0.61)$ & $\bm{0.61}$ $(0.09)$ & $0.68$ $(0.39)$ \\
cluster\_agg & $0.43$ $(0.0)$ & $\bm{0.8}$ $(0.44)$ & $0.69$ $(0.3)$ & $0.67$ $(\bm{0.64})$ & $0.59$ $(0.16)$ & $0.68$ $(0.41)$ \\
prefix\_index & $0.47$ $(0.0)$ & $0.75$ $(0.41)$ & $\bm{0.71}$ $(0.19)$ & $0.67$ $(0.61)$ & $0.6$ $(\bm{0.22})$ & $0.67$ $(0.39)$ \\
prefix\_laststate & $0.46$ $(0.01)$ & $\bm{0.8}$ $(0.43)$ & $\bm{0.71}$ $(0.21)$ & $0.66$ $(0.62)$ & $0.59$ $(0.14)$ & $0.68$ $(0.4)$ \\
prefix\_agg & $0.46$ $(0.02)$ & $0.76$ $(0.41)$ & $0.7$ $(0.26)$ & $0.68$ $(\bm{0.64})$ & $0.59$ $(0.16)$ & $\bm{0.7}$ $(0.41)$ \\
\midrule
 & {\bfseries bpic2017\_1} & {\bfseries bpic2017\_2} & {\bfseries bpic2017\_3} & {\bfseries traffic} & {\bfseries hospital\_1} & {\bfseries hospital\_2}
\\ \midrule
single\_laststate & $\bm{0.83}$ $(0.7)$ & $\bm{0.81}$ $(0.47)$ & $0.79$ $(0.72)$ & $0.66$ $(0.67)$ & $\bm{0.88}$ $(\bm{0.66})$ & $\bm{0.72}$ $(0.11)$ \\
single\_agg & $\bm{0.83}$ $(0.71)$ & $0.8$ $(\bm{0.48})$ & $\bm{0.8}$ $(\bm{0.73})$ & $0.65$ $(0.66)$ & $\bm{0.88}$ $(0.65)$ & $0.7$ $(0.11)$ \\
knn\_laststate & $0.78$ $(0.65)$ & $0.61$ $(0.09)$ & $0.78$ $(0.67)$ & $\bm{0.67}$ $(\bm{0.7})$ & $0.81$ $(0.51)$ & $0.58$ $(0.05)$ \\
knn\_agg & $0.78$ $(0.65)$ & $0.57$ $(0.13)$ & $0.76$ $(0.66)$ & $0.66$ $(\bm{0.7})$ & $0.81$ $(0.44)$ & $0.56$ $(0.04)$ \\
state\_laststate & $\bm{0.83}$ $(0.7)$ & $0.8$ $(0.46)$ & $\bm{0.8}$ $(0.72)$ & $0.65$ $(0.66)$ & $\bm{0.88}$ $(0.65)$ & $0.71$ $(\bm{0.12})$ \\
state\_agg & $\bm{0.83}$ $(\bm{0.72})$ & $0.8$ $(0.47)$ & $\bm{0.8}$ $(\bm{0.73})$ & $0.66$ $(0.66)$ & $\bm{0.88}$ $(0.64)$ & $0.7$ $(0.07)$ \\
cluster\_laststate & $\bm{0.83}$ $(0.71)$ & $0.8$ $(0.46)$ & $\bm{0.8}$ $(0.72)$ & $0.66$ $(0.66)$ & $\bm{0.88}$ $(0.65)$ & $0.7$ $(\bm{0.12})$ \\
cluster\_agg & $\bm{0.83}$ $(0.71)$ & $0.79$ $(0.46)$ & $\bm{0.8}$ $(\bm{0.73})$ & $0.66$ $(0.66)$ & $\bm{0.88}$ $(0.65)$ & $0.69$ $(0.07)$ \\
prefix\_index & $\bm{0.83}$ $(\bm{0.72})$ & $0.8$ $(0.46)$ & $0.79$ $(\bm{0.73})$ & $0.66$ $(0.66)$ & $\bm{0.88}$ $(0.64)$ & $0.69$ $(0.1)$ \\
prefix\_laststate & $\bm{0.83}$ $(0.7)$ & $0.8$ $(0.46)$ & $0.79$ $(0.72)$ & $0.65$ $(0.66)$ & $\bm{0.88}$ $(0.64)$ & $0.71$ $(0.11)$ \\
prefix\_agg & $\bm{0.83}$ $(0.71)$ & $0.8$ $(0.47)$ & $\bm{0.8}$ $(\bm{0.73})$ & $0.66$ $(0.66)$ & $\bm{0.88}$ $(0.63)$ & $0.7$ $(0.09)$ \\
\bottomrule
\end{tabular}
\end{adjustbox}
\end{center}
\end{table}

\begin{table}[hbtp]
\caption{Overall AUC \added{(F-score)} for {\bfseries logistic regression}}
\vspace{-\baselineskip}
\label{table:avg_results_logit}
\begin{center}
\begin{adjustbox}{max width=\textwidth}
\begin{tabular}{@{}lcccccc@{}}
\toprule
 & {\bfseries bpic2011\_1} & {\bfseries bpic2011\_2} & {\bfseries bpic2011\_3} & {\bfseries bpic2011\_4} & {\bfseries insurance\_1} & {\bfseries insurance\_2}
\\ \midrule
single\_laststate & $0.9$ $(0.82)$ & $0.9$ $(0.83)$ & $0.92$ $(0.75)$ & $0.88$ $(0.76)$ & $0.84$ $(0.47)$ & $\bm{0.83}$ $(0.56)$ \\
single\_agg & $0.92$ $(0.83)$ & $0.94$ $(0.9)$ & $0.96$ $(0.86)$ & $0.87$ $(0.75)$ & $0.79$ $(0.46)$ & $0.77$ $(0.49)$ \\
knn\_laststate & $0.91$ $(0.79)$ & $0.94$ $(\bm{0.92})$ & $0.92$ $(0.87)$ & $0.81$ $(\bm{0.82})$ & $0.77$ $(0.49)$ & $0.65$ $(0.46)$ \\
knn\_agg & $0.82$ $(0.73)$ & $0.86$ $(0.86)$ & $0.87$ $(0.77)$ & $0.75$ $(0.72)$ & $0.79$ $(0.5)$ & $0.7$ $(0.57)$ \\
state\_laststate & $0.91$ $(0.8)$ & $0.89$ $(0.86)$ & $0.91$ $(0.76)$ & $0.87$ $(0.79)$ & $0.82$ $(0.48)$ & $0.8$ $(\bm{0.62})$ \\
state\_agg & $0.91$ $(0.82)$ & $0.94$ $(0.91)$ & $0.96$ $(0.86)$ & $0.85$ $(0.77)$ & $0.82$ $(0.49)$ & $0.75$ $(0.51)$ \\
cluster\_laststate & $0.91$ $(0.8)$ & $0.9$ $(0.88)$ & $0.95$ $(0.86)$ & $\bm{0.89}$ $(0.8)$ & $0.82$ $(0.53)$ & $0.77$ $(0.53)$ \\
cluster\_agg & $\bm{0.94}$ $(\bm{0.84})$ & $\bm{0.95}$ $(\bm{0.92})$ & $\bm{0.97}$ $(\bm{0.88})$ & $0.87$ $(0.77)$ & $0.82$ $(\bm{0.59})$ & $0.76$ $(0.58)$ \\
prefix\_index & $0.91$ $(0.81)$ & $0.92$ $(0.88)$ & $0.94$ $(0.83)$ & $0.8$ $(0.73)$ & $0.82$ $(0.55)$ & $0.67$ $(0.53)$ \\
prefix\_laststate & $0.9$ $(0.79)$ & $0.89$ $(0.83)$ & $0.95$ $(0.82)$ & $0.86$ $(0.77)$ & $0.83$ $(0.5)$ & $0.74$ $(0.55)$ \\
prefix\_agg & $0.92$ $(0.83)$ & $\bm{0.95}$ $(0.91)$ & $0.96$ $(0.79)$ & $0.86$ $(0.77)$ & $\bm{0.85}$ $(0.57)$ & $0.75$ $(0.6)$ \\
\midrule
 & {\bfseries bpic2015\_1} & {\bfseries bpic2015\_2} & {\bfseries bpic2015\_3} & {\bfseries bpic2015\_4} & {\bfseries bpic2015\_5} & {\bfseries production}
\\ \midrule
single\_laststate & $0.8$ $(0.5)$ & $0.87$ $(0.51)$ & $0.79$ $(0.45)$ & $0.86$ $(0.5)$ & $0.81$ $(0.58)$ & $0.68$ $(0.59)$ \\
single\_agg & $0.81$ $(0.61)$ & $0.9$ $(0.68)$ & $\bm{0.89}$ $(0.7)$ & $0.87$ $(0.51)$ & $\bm{0.86}$ $(0.74)$ & $0.67$ $(0.58)$ \\
knn\_laststate & $0.76$ $(0.52)$ & $0.84$ $(\bm{0.73})$ & $0.85$ $(\bm{0.72})$ & $0.85$ $(0.57)$ & $0.81$ $(0.68)$ & $\bm{0.71}$ $(0.58)$ \\
knn\_agg & $0.75$ $(0.47)$ & $0.81$ $(0.67)$ & $0.84$ $(\bm{0.72})$ & $0.81$ $(0.51)$ & $0.82$ $(0.7)$ & $0.57$ $(0.46)$ \\
state\_laststate & $0.75$ $(0.48)$ & $0.83$ $(0.53)$ & $0.81$ $(0.53)$ & $0.87$ $(0.52)$ & $0.82$ $(0.62)$ & $0.65$ $(0.56)$ \\
state\_agg & $0.79$ $(0.57)$ & $0.88$ $(0.67)$ & $0.87$ $(0.7)$ & $0.89$ $(0.64)$ & $0.84$ $(0.72)$ & $0.58$ $(0.38)$ \\
cluster\_laststate & $0.46$ $(0.19)$ & $0.81$ $(0.58)$ & $0.84$ $(0.61)$ & $0.85$ $(0.63)$ & $0.85$ $(0.7)$ & $0.66$ $(0.59)$ \\
cluster\_agg & $\bm{0.86}$ $(0.61)$ & $\bm{0.91}$ $(0.69)$ & $0.88$ $(0.7)$ & $0.86$ $(0.63)$ & $0.85$ $(0.74)$ & $\bm{0.71}$ $(0.59)$ \\
prefix\_index & $0.84$ $(0.51)$ & $0.87$ $(0.64)$ & $0.86$ $(0.69)$ & $0.89$ $(0.57)$ & $0.83$ $(0.69)$ & $\bm{0.71}$ $(\bm{0.6})$ \\
prefix\_laststate & $0.73$ $(0.43)$ & $0.81$ $(0.45)$ & $0.79$ $(0.49)$ & $0.82$ $(0.45)$ & $0.83$ $(0.62)$ & $0.68$ $(0.56)$ \\
prefix\_agg & $0.85$ $(\bm{0.63})$ & $0.89$ $(0.69)$ & $\bm{0.89}$ $(0.71)$ & $\bm{0.9}$ $(\bm{0.66})$ & $\bm{0.86}$ $(\bm{0.76})$ & $0.67$ $(\bm{0.6})$ \\
\midrule
 & {\bfseries sepsis\_1} & {\bfseries sepsis\_2} & {\bfseries sepsis\_3} & {\bfseries bpic2012\_1} & {\bfseries bpic2012\_2} & {\bfseries bpic2012\_3}
\\ \midrule
single\_laststate & $0.43$ $(0.0)$ & $0.88$ $(0.44)$ & $\bm{0.74}$ $(0.34)$ & $0.65$ $(0.57)$ & $0.58$ $(0.09)$ & $\bm{0.7}$ $(0.32)$ \\
single\_agg & $\bm{0.57}$ $(0.09)$ & $0.86$ $(0.47)$ & $0.73$ $(0.37)$ & $0.65$ $(0.53)$ & $\bm{0.59}$ $(0.13)$ & $0.69$ $(0.3)$ \\
knn\_laststate & $0.43$ $(0.12)$ & $0.76$ $(0.35)$ & $0.58$ $(0.27)$ & $0.6$ $(0.49)$ & $0.54$ $(\bm{0.17})$ & $0.64$ $(\bm{0.52})$ \\
knn\_agg & $0.5$ $(\bm{0.13})$ & $0.76$ $(0.35)$ & $0.59$ $(0.27)$ & $0.57$ $(0.47)$ & $0.55$ $(\bm{0.17})$ & $0.6$ $(0.45)$ \\
state\_laststate & $0.47$ $(0.01)$ & $\bm{0.9}$ $(0.43)$ & $0.71$ $(0.34)$ & $0.65$ $(0.56)$ & $\bm{0.59}$ $(0.09)$ & $0.69$ $(0.34)$ \\
state\_agg & $0.55$ $(\bm{0.13})$ & $\bm{0.9}$ $(0.43)$ & $0.7$ $(0.38)$ & $0.66$ $(0.58)$ & $\bm{0.59}$ $(0.13)$ & $\bm{0.7}$ $(0.36)$ \\
cluster\_laststate & $0.47$ $(0.05)$ & $0.86$ $(0.43)$ & $0.67$ $(0.32)$ & $0.64$ $(0.55)$ & $0.58$ $(0.1)$ & $0.68$ $(0.33)$ \\
cluster\_agg & $0.51$ $(0.11)$ & $0.87$ $(0.44)$ & $0.7$ $(0.35)$ & $0.65$ $(0.55)$ & $0.58$ $(0.13)$ & $0.69$ $(0.34)$ \\
prefix\_index & $0.5$ $(0.12)$ & $0.88$ $(0.48)$ & $0.7$ $(0.84)$ & $0.66$ $(\bm{0.59})$ & $0.56$ $(0.16)$ & $0.67$ $(0.39)$ \\
prefix\_laststate & $0.42$ $(0.11)$ & $0.88$ $(0.45)$ & $0.7$ $(0.86)$ & $0.65$ $(0.56)$ & $0.58$ $(0.09)$ & $0.69$ $(0.34)$ \\
prefix\_agg & $0.55$ $(0.12)$ & $0.87$ $(\bm{0.49})$ & $0.73$ $(\bm{0.87})$ & $\bm{0.67}$ $(0.58)$ & $\bm{0.59}$ $(0.13)$ & $0.69$ $(0.34)$ \\
\midrule
 & {\bfseries bpic2017\_1} & {\bfseries bpic2017\_2} & {\bfseries bpic2017\_3} & {\bfseries traffic} & {\bfseries hospital\_1} & {\bfseries hospital\_2}
\\ \midrule
single\_laststate & $0.82$ $(0.67)$ & $\bm{0.81}$ $(\bm{0.46})$ & $0.79$ $(0.73)$ & $0.65$ $(0.64)$ & $\bm{0.88}$ $(0.58)$ & $0.73$ $(0.05)$ \\
single\_agg & $\bm{0.83}$ $(0.67)$ & $0.79$ $(0.23)$ & $0.79$ $(\bm{0.74})$ & $0.66$ $(\bm{0.65})$ & $0.87$ $(0.6)$ & $0.72$ $(0.04)$ \\
knn\_laststate & $0.7$ $(0.59)$ & $0.62$ $(0.3)$ & $0.77$ $(0.7)$ & $0.6$ $(\bm{0.65})$ & $0.78$ $(0.45)$ & $0.59$ $(0.07)$ \\
knn\_agg & $0.73$ $(0.61)$ & $0.63$ $(0.28)$ & $0.75$ $(0.67)$ & $0.63$ $(0.63)$ & $0.8$ $(0.48)$ & $0.57$ $(0.02)$ \\
state\_laststate & $0.82$ $(0.67)$ & $0.72$ $(0.3)$ & $0.79$ $(0.73)$ & $0.67$ $(0.64)$ & $\bm{0.88}$ $(\bm{0.64})$ & $0.73$ $(0.09)$ \\
state\_agg & $0.82$ $(0.68)$ & $0.8$ $(0.45)$ & $\bm{0.8}$ $(\bm{0.74})$ & $\bm{0.68}$ $(0.64)$ & $\bm{0.88}$ $(0.63)$ & $0.71$ $(0.09)$ \\
cluster\_laststate & $0.82$ $(0.67)$ & $0.78$ $(0.41)$ & $0.79$ $(\bm{0.74})$ & $\bm{0.68}$ $(0.64)$ & $\bm{0.88}$ $(0.63)$ & $0.72$ $(0.08)$ \\
cluster\_agg & $0.81$ $(0.67)$ & $0.77$ $(0.39)$ & $0.79$ $(0.73)$ & $\bm{0.68}$ $(\bm{0.65})$ & $\bm{0.88}$ $(0.62)$ & $0.7$ $(0.09)$ \\
prefix\_index & $0.82$ $(0.68)$ & $0.78$ $(0.41)$ & $0.79$ $(0.73)$ & $\bm{0.68}$ $(0.64)$ & $0.87$ $(0.57)$ & $0.73$ $(\bm{0.1})$ \\
prefix\_laststate & $0.82$ $(0.67)$ & $0.8$ $(0.45)$ & $0.79$ $(\bm{0.74})$ & $0.67$ $(0.64)$ & $\bm{0.88}$ $(0.59)$ & $\bm{0.74}$ $(0.08)$ \\
prefix\_agg & $\bm{0.83}$ $(\bm{0.69})$ & $0.8$ $(0.41)$ & $0.79$ $(\bm{0.74})$ & $\bm{0.68}$ $(0.64)$ & $\bm{0.88}$ $(0.55)$ & $0.73$ $(0.07)$ \\
\bottomrule
\end{tabular}
\end{adjustbox}
\end{center}
\end{table}

\begin{table}[hbtp]
\caption{Overall AUC \added{(F-score)} for {\bfseries SVM}}
\vspace{-\baselineskip}
\label{table:avg_results_svm}
\begin{center}
\begin{adjustbox}{max width=\textwidth}
\begin{tabular}{@{}lcccccc@{}}
\toprule
 & {\bfseries bpic2011\_1} & {\bfseries bpic2011\_2} & {\bfseries bpic2011\_3} & {\bfseries bpic2011\_4} & {\bfseries insurance\_1} & {\bfseries insurance\_2}
\\ \midrule
single\_laststate & $0.89$ $(0.65)$ & $0.9$ $(0.76)$ & $0.92$ $(0.0)$ & $0.86$ $(0.0)$ & $0.78$ $(0.38)$ & $\bm{0.82}$ $(0.39)$ \\
single\_agg & $0.87$ $(0.73)$ & $\bm{0.95}$ $(\bm{0.91})$ & $\bm{0.96}$ $(0.0)$ & $0.87$ $(0.35)$ & $0.81$ $(0.24)$ & $0.78$ $(0.42)$ \\
knn\_laststate & $\bm{0.92}$ $(0.83)$ & $\bm{0.95}$ $(\bm{0.91})$ & $0.93$ $(\bm{0.81})$ & $0.76$ $(0.46)$ & $0.74$ $(\bm{0.46})$ & $0.63$ $(0.35)$ \\
knn\_agg & $0.88$ $(\bm{0.85})$ & $0.94$ $(0.9)$ & $0.92$ $(0.66)$ & $0.72$ $(0.29)$ & $0.77$ $(\bm{0.46})$ & $0.7$ $(0.15)$ \\
state\_laststate & $0.88$ $(0.54)$ & $0.89$ $(0.82)$ & $0.86$ $(0.55)$ & $0.8$ $(\bm{0.62})$ & $0.79$ $(\bm{0.46})$ & $0.77$ $(\bm{0.46})$ \\
state\_agg & $0.91$ $(0.75)$ & $\bm{0.95}$ $(0.87)$ & $0.94$ $(0.61)$ & $0.82$ $(\bm{0.62})$ & $0.83$ $(0.42)$ & $0.73$ $(0.0)$ \\
cluster\_laststate & $0.9$ $(0.78)$ & $0.89$ $(0.79)$ & $0.94$ $(0.58)$ & $\bm{0.88}$ $(0.5)$ & $0.68$ $(0.23)$ & $0.68$ $(0.37)$ \\
cluster\_agg & $\bm{0.92}$ $(0.75)$ & $0.94$ $(0.84)$ & $0.93$ $(0.78)$ & $0.86$ $(0.49)$ & $0.74$ $(0.27)$ & $0.76$ $(0.35)$ \\
prefix\_index & $0.91$ $(0.66)$ & $0.92$ $(0.77)$ & $0.93$ $(0.52)$ & $0.82$ $(0.36)$ & $0.83$ $(0.26)$ & $0.65$ $(0.0)$ \\
prefix\_laststate & $0.87$ $(0.59)$ & $0.9$ $(0.81)$ & $0.93$ $(0.38)$ & $0.84$ $(0.0)$ & $0.8$ $(0.09)$ & $0.74$ $(0.41)$ \\
prefix\_agg & $\bm{0.92}$ $(0.64)$ & $0.94$ $(0.89)$ & $0.95$ $(0.12)$ & $0.86$ $(0.0)$ & $\bm{0.85}$ $(0.13)$ & $0.77$ $(0.33)$ \\
\midrule
 & {\bfseries bpic2015\_1} & {\bfseries bpic2015\_2} & {\bfseries bpic2015\_3} & {\bfseries bpic2015\_4} & {\bfseries bpic2015\_5} & {\bfseries production}
\\ \midrule
single\_laststate & $0.78$ $(0.54)$ & $0.8$ $(0.5)$ & $0.82$ $(0.43)$ & $0.86$ $(0.41)$ & $0.81$ $(0.55)$ & $0.71$ $(0.6)$ \\
single\_agg & $\bm{0.82}$ $(\bm{0.56})$ & $\bm{0.88}$ $(0.57)$ & $\bm{0.88}$ $(0.67)$ & $0.85$ $(\bm{0.56})$ & $0.85$ $(0.64)$ & $0.66$ $(0.54)$ \\
knn\_laststate & $0.72$ $(0.48)$ & $0.76$ $(0.6)$ & $0.8$ $(0.66)$ & $0.76$ $(\bm{0.56})$ & $0.78$ $(0.6)$ & $0.6$ $(0.41)$ \\
knn\_agg & $0.7$ $(0.49)$ & $0.78$ $(\bm{0.62})$ & $0.81$ $(\bm{0.68})$ & $0.82$ $(\bm{0.56})$ & $0.78$ $(0.63)$ & $0.63$ $(0.51)$ \\
state\_laststate & $0.68$ $(0.44)$ & $0.76$ $(0.5)$ & $0.75$ $(0.42)$ & $0.8$ $(0.42)$ & $0.8$ $(0.57)$ & $0.67$ $(0.54)$ \\
state\_agg & $0.63$ $(0.34)$ & $0.85$ $(\bm{0.62})$ & $0.85$ $(0.53)$ & $0.86$ $(0.52)$ & $0.85$ $(0.57)$ & $0.63$ $(0.56)$ \\
cluster\_laststate & $0.78$ $(0.37)$ & $0.8$ $(0.46)$ & $0.82$ $(0.61)$ & $0.85$ $(0.47)$ & $0.85$ $(0.57)$ & $\bm{0.74}$ $(\bm{0.62})$ \\
cluster\_agg & $0.79$ $(0.0)$ & $0.84$ $(0.0)$ & $\bm{0.88}$ $(0.49)$ & $\bm{0.88}$ $(0.5)$ & $0.83$ $(0.17)$ & $\bm{0.74}$ $(0.59)$ \\
prefix\_index & $0.77$ $(0.15)$ & $0.83$ $(0.23)$ & $0.85$ $(0.51)$ & $0.86$ $(0.22)$ & $0.83$ $(0.52)$ & $0.72$ $(0.57)$ \\
prefix\_laststate & $0.67$ $(0.07)$ & $0.69$ $(0.04)$ & $0.75$ $(0.0)$ & $0.75$ $(0.09)$ & $0.82$ $(0.55)$ & $0.67$ $(0.49)$ \\
prefix\_agg & $0.81$ $(0.01)$ & $0.86$ $(0.53)$ & $\bm{0.88}$ $(0.54)$ & $0.86$ $(0.34)$ & $\bm{0.86}$ $(\bm{0.66})$ & $0.66$ $(0.53)$ \\
\midrule
 & {\bfseries sepsis\_1} & {\bfseries sepsis\_2} & {\bfseries sepsis\_3} & {\bfseries bpic2012\_1} & {\bfseries bpic2012\_2} & {\bfseries bpic2012\_3}
\\ \midrule
single\_laststate & $0.5$ $(0.0)$ & $0.78$ $(\bm{0.41})$ & $\bm{0.72}$ $(0.26)$ & $0.63$ $(0.52)$ & $0.56$ $(0.09)$ & $0.68$ $(0.18)$ \\
single\_agg & $0.49$ $(0.0)$ & $0.82$ $(0.0)$ & $\bm{0.72}$ $(0.0)$ & $0.63$ $(0.38)$ & $0.55$ $(0.0)$ & $\bm{0.7}$ $(0.2)$ \\
knn\_laststate & $0.47$ $(\bm{0.09})$ & $0.7$ $(0.11)$ & $0.62$ $(0.0)$ & $0.57$ $(0.5)$ & $0.51$ $(0.06)$ & $0.59$ $(\bm{0.3})$ \\
knn\_agg & $0.48$ $(0.0)$ & $0.71$ $(0.01)$ & $0.59$ $(0.0)$ & $0.63$ $(\bm{0.56})$ & $0.53$ $(0.06)$ & $0.58$ $(0.28)$ \\
state\_laststate & $0.5$ $(0.0)$ & $0.75$ $(0.0)$ & $0.69$ $(0.05)$ & $0.63$ $(0.42)$ & $0.52$ $(0.09)$ & $0.61$ $(0.13)$ \\
state\_agg & $\bm{0.54}$ $(0.0)$ & $0.81$ $(0.0)$ & $0.65$ $(\bm{0.27})$ & $\bm{0.64}$ $(0.45)$ & $0.53$ $(0.09)$ & $0.65$ $(0.27)$ \\
cluster\_laststate & $0.49$ $(0.0)$ & $0.81$ $(0.0)$ & $0.67$ $(0.0)$ & $\bm{0.64}$ $(0.52)$ & $0.54$ $(0.08)$ & $0.65$ $(\bm{0.3})$ \\
cluster\_agg & $0.5$ $(0.0)$ & $0.79$ $(0.4)$ & $0.68$ $(0.0)$ & $0.6$ $(0.28)$ & $0.55$ $(\bm{0.12})$ & $0.67$ $(0.25)$ \\
prefix\_index & $\bm{0.54}$ $(0.0)$ & $\bm{0.84}$ $(0.33)$ & $0.65$ $(0.0)$ & $0.61$ $(0.39)$ & $\bm{0.57}$ $(0.06)$ & $0.66$ $(0.05)$ \\
prefix\_laststate & $0.46$ $(0.0)$ & $0.8$ $(0.22)$ & $0.66$ $(0.18)$ & $\bm{0.64}$ $(0.46)$ & $0.56$ $(0.04)$ & $0.67$ $(0.23)$ \\
prefix\_agg & $0.51$ $(0.01)$ & $0.81$ $(0.0)$ & $0.68$ $(0.01)$ & $\bm{0.64}$ $(0.31)$ & $\bm{0.57}$ $(0.02)$ & $0.67$ $(0.0)$ \\
\midrule
 & {\bfseries bpic2017\_1} & {\bfseries bpic2017\_2} & {\bfseries bpic2017\_3} & {\bfseries traffic} & {\bfseries hospital\_1} & {\bfseries hospital\_2}
\\ \midrule
single\_laststate & $0.75$ $(0.14)$ & $0.59$ $(0.0)$ & $0.71$ $(0.67)$ & $0.64$ $(0.63)$ & $\bm{0.85}$ $(0.49)$ & $0.51$ $(0.0)$ \\
single\_agg & $\bm{0.79}$ $(\bm{0.67})$ & $0.71$ $(0.0)$ & $0.64$ $(0.0)$ & $0.67$ $(0.5)$ & $0.73$ $(0.0)$ & $\bm{0.78}$ $(0.0)$ \\
knn\_laststate & $0.55$ $(0.28)$ & $0.54$ $(0.0)$ & $0.68$ $(0.51)$ & $0.62$ $(\bm{0.69})$ & $0.78$ $(0.48)$ & $0.5$ $(0.0)$ \\
knn\_agg & $0.63$ $(0.06)$ & $0.57$ $(0.0)$ & $0.71$ $(0.54)$ & $0.63$ $(0.17)$ & $0.61$ $(0.43)$ & $0.5$ $(0.04)$ \\
state\_laststate & $0.57$ $(0.11)$ & $0.52$ $(0.0)$ & $0.62$ $(0.06)$ & $0.65$ $(0.6)$ & $0.84$ $(0.56)$ & $0.57$ $(0.03)$ \\
state\_agg & $\bm{0.79}$ $(0.58)$ & $0.57$ $(0.03)$ & $0.55$ $(0.4)$ & $0.66$ $(0.59)$ & $0.84$ $(\bm{0.59})$ & $0.57$ $(0.01)$ \\
cluster\_laststate & $0.59$ $(0.07)$ & $0.56$ $(0.0)$ & $\bm{0.77}$ $(\bm{0.68})$ & $0.66$ $(0.58)$ & $0.82$ $(0.57)$ & $0.61$ $(0.07)$ \\
cluster\_agg & $0.64$ $(0.13)$ & $0.73$ $(0.0)$ & $0.73$ $(0.58)$ & $\bm{0.73}$ $(0.58)$ & $0.82$ $(0.54)$ & $0.59$ $(0.0)$ \\
prefix\_index & $0.68$ $(0.0)$ & $0.73$ $(0.28)$ & $0.68$ $(0.0)$ & $0.59$ $(0.27)$ & $0.82$ $(0.46)$ & $0.59$ $(0.0)$ \\
prefix\_laststate & $0.76$ $(0.34)$ & $\bm{0.77}$ $(\bm{0.46})$ & $0.74$ $(0.58)$ & $0.65$ $(0.6)$ & $\bm{0.85}$ $(0.52)$ & $0.54$ $(0.02)$ \\
prefix\_agg & $0.6$ $(0.01)$ & $0.73$ $(0.28)$ & $0.73$ $(0.0)$ & $0.65$ $(0.6)$ & $0.84$ $(0.52)$ & $0.66$ $(\bm{0.09})$ \\
\bottomrule
\end{tabular}
\end{adjustbox}
\end{center}
\end{table}

\begin{figure}[hbtp!]
\centering
\includegraphics[width=1\textwidth]{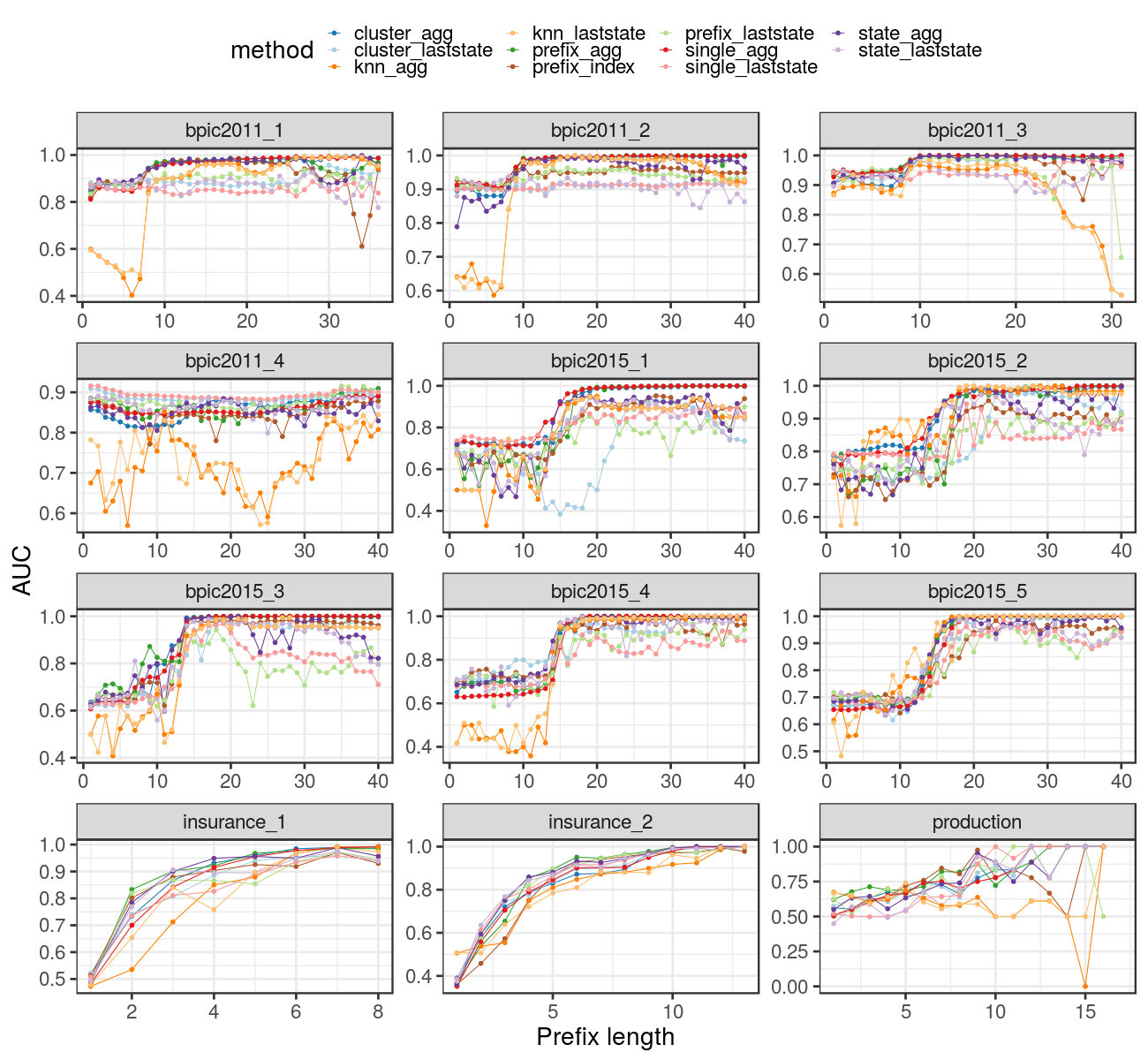}
\caption{AUC across prefix lengths using {\bfseries XGBoost}, all methods \deleted{(1)}}
\label{fig:results_all_xgboost_1}
\end{figure}

\begin{figure}[hbtp!]
\centering
\includegraphics[width=1\textwidth]{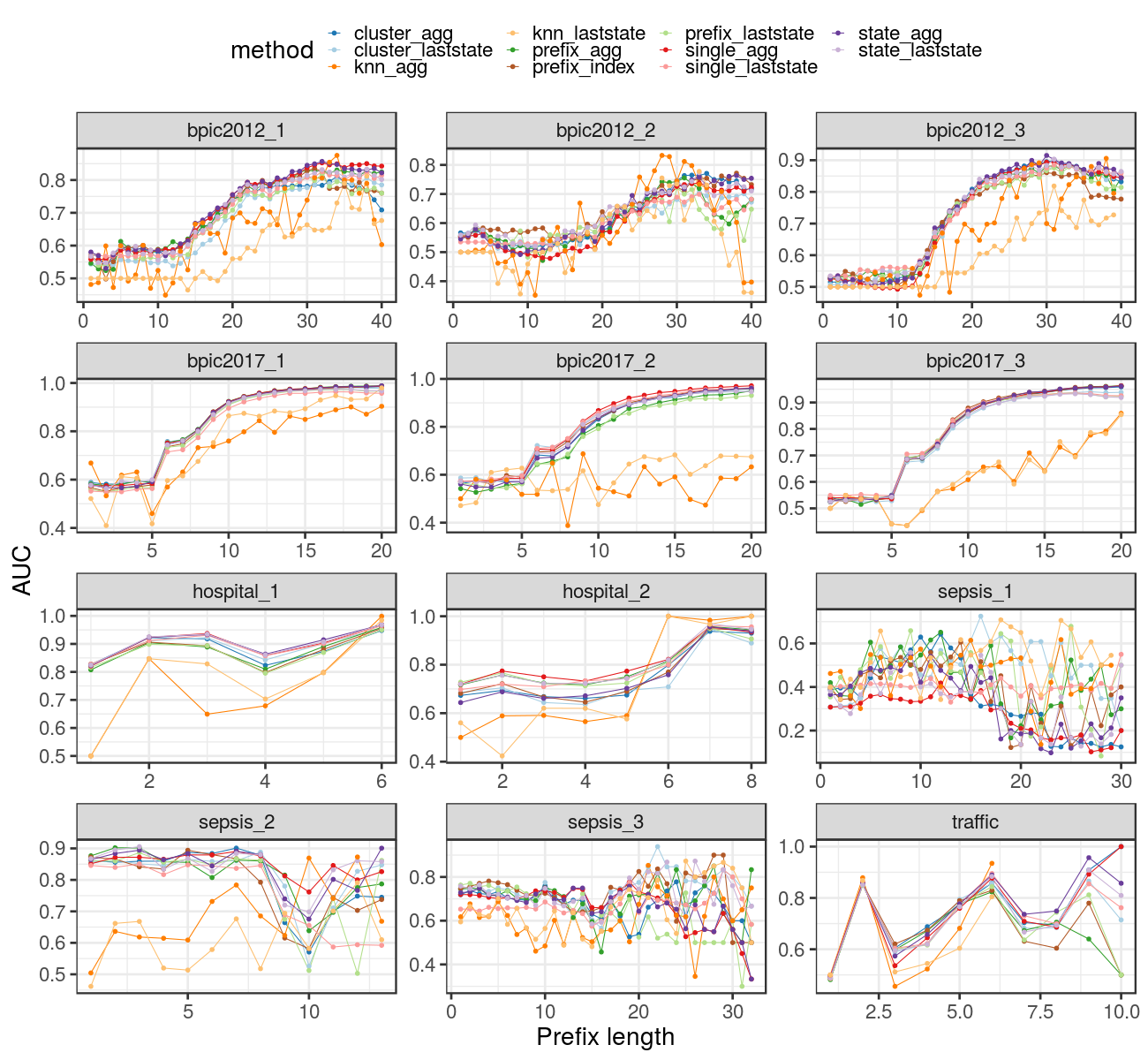}
\caption{AUC across prefix lengths using {\bfseries XGBoost}, all methods (\replaced{continued}{2})}
\label{fig:results_all_xgboost_2}
\end{figure}

\begin{figure}[hbtp!]
\centering
\includegraphics[width=1\textwidth]{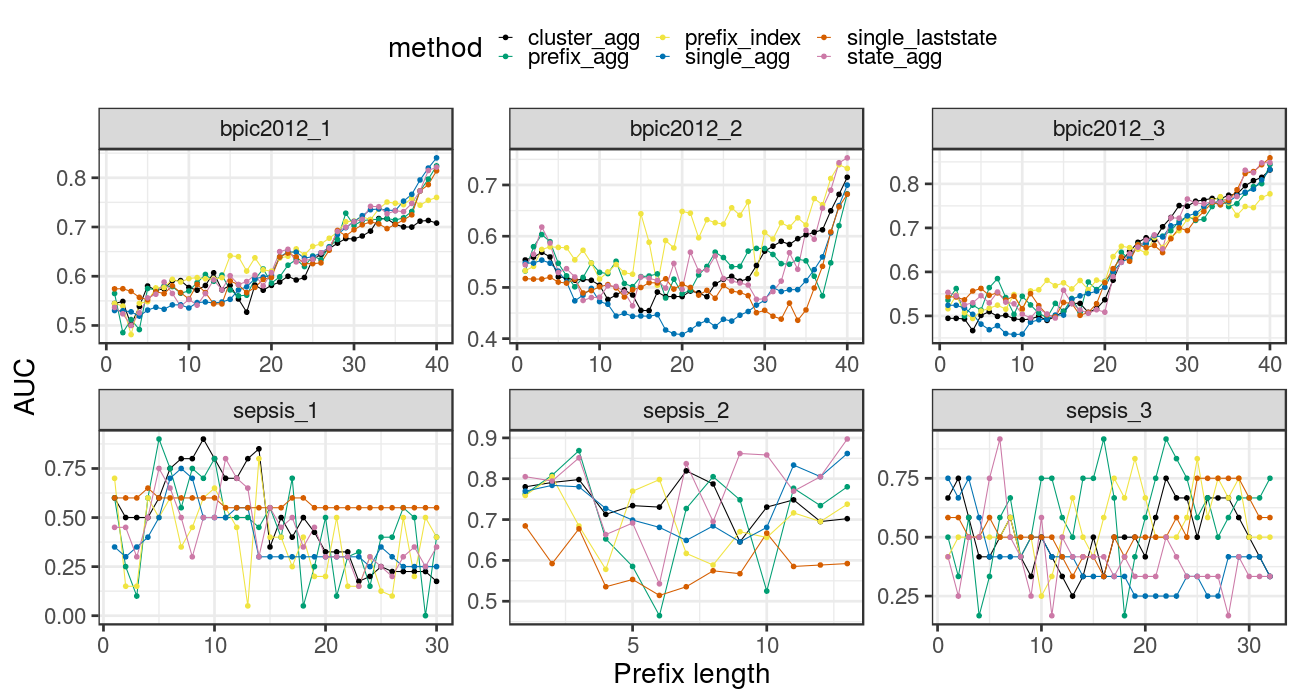}
\caption{AUC across prefix lengths using {\bfseries XGBoost}, long traces only}
\label{fig:results_long_traces}
\end{figure}


\begin{figure}[hbtp!]
	\hspace{0cm}
	\begin{subfigure}[b]{0.48\linewidth}
     \includegraphics[width=\textwidth]{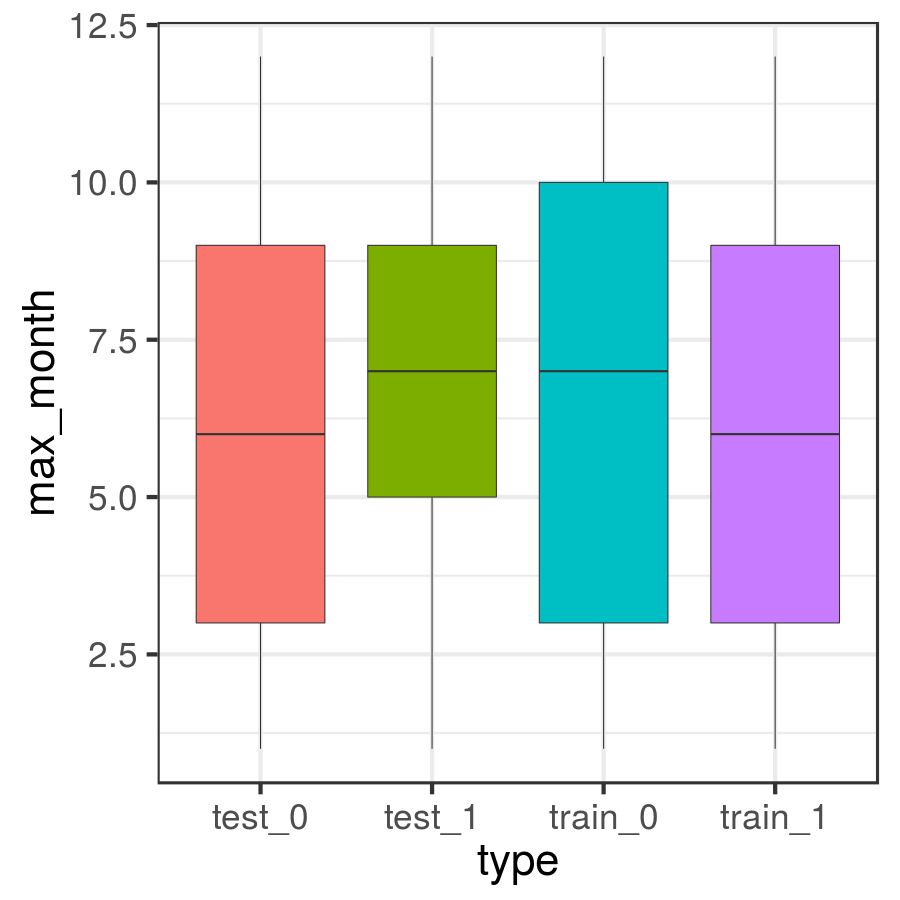}
     \caption{Attribute = \emph{max\_month}. Significant differences in means between train\_0--test\_0 ($Z=6.077$, $p<.001$) and train\_1--test\_1 ($Z=7.972$, $p<.001$).}
     \label{fig:bpic2011_4_boxplot_max_month}
 	\end{subfigure}
 	\hspace{0.2cm}
   \begin{subfigure}[b]{0.48\linewidth}
    \includegraphics[width=\textwidth]{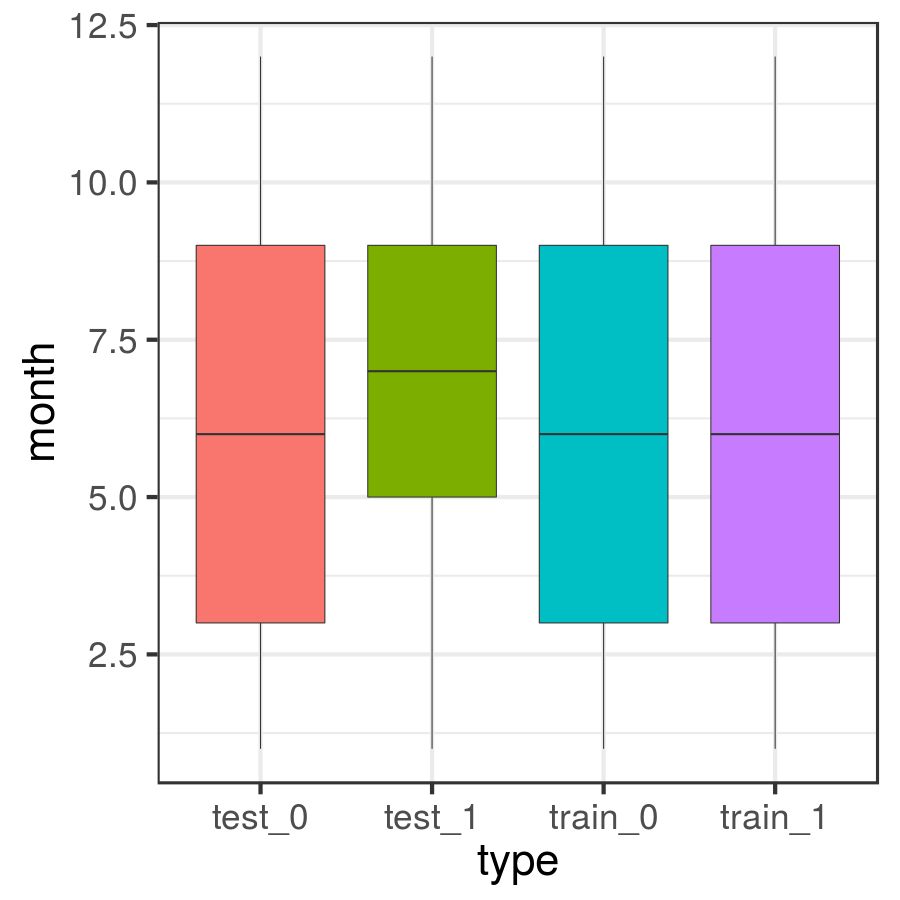}
    \caption{Attribute = \emph{month}. The difference is significant between train\_1--test\_1 ($Z=8.754$, $p<.001$), but not between train\_0--test\_0 ($Z=.028$, $p=.978$).}
    \label{fig:bpic2011_4_boxplot_last_month}
   \end{subfigure}  
    \caption{Concept drift in the \emph{bpic2011\_4} log. The distributions of the variables are different across the two classes in the train and the test set. The drift becomes more evident in the \emph{max\_month} feature used by the aggregation encoding, while it is not so severe in the original \emph{month} feature used by the last state encoding. Statistical significance of the differences is assessed using Wilcoxon signed-rank test.}
    \label{fig:bpic2011_f4_boxplots}
\end{figure}

\begin{figure}[hbtp!]
	\hspace{-0.2cm}
	\begin{subfigure}[b]{0.3\linewidth}
     \includegraphics[width=\textwidth]{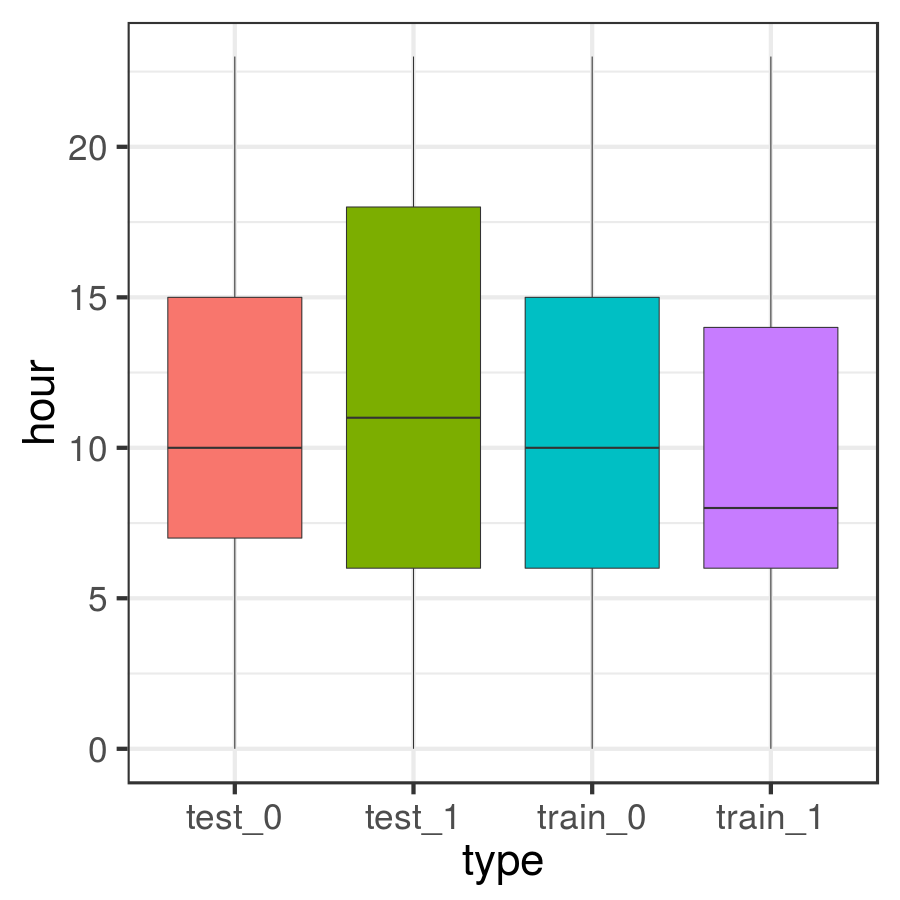}
     \caption{Attribute = \emph{hour}. Significant difference between train\_1--test\_1 ($Z=3.651$, $p<.001$), but not between train\_0--test\_0 ($Z=1.373$, $p=.17$).}
     \label{fig:sepsis_cases_1_boxplot_hour}
 	\end{subfigure}
 	\hspace{0.2cm}
   \begin{subfigure}[b]{0.3\linewidth}
    \includegraphics[width=\textwidth]{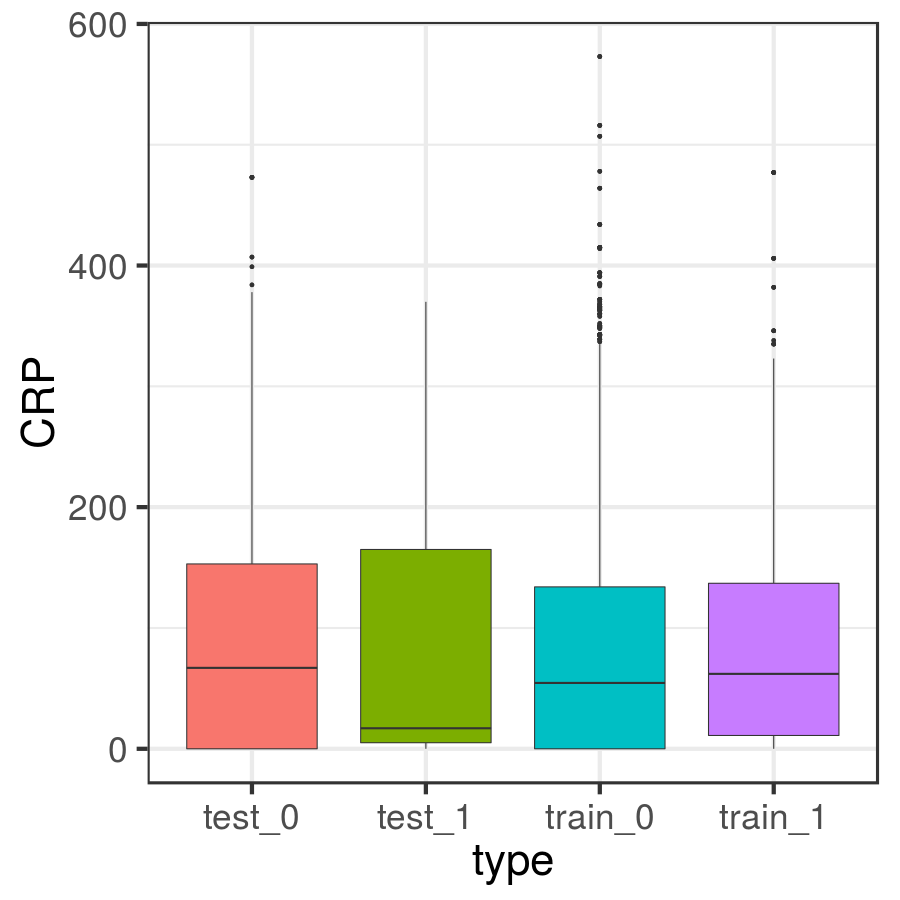}
    \caption{Attribute = \emph{CRP}. Significant differences between train\_0--test\_0 ($Z=3.61$, $p<.001$) and train\_1--test\_1 ($Z=2.492$, $p=.013$).}
    \label{fig:sepsis_cases_1_boxplot_crp}
   \end{subfigure}  
    	\hspace{0.2cm}      
   \begin{subfigure}[b]{0.3\linewidth}
     \includegraphics[width=\textwidth]{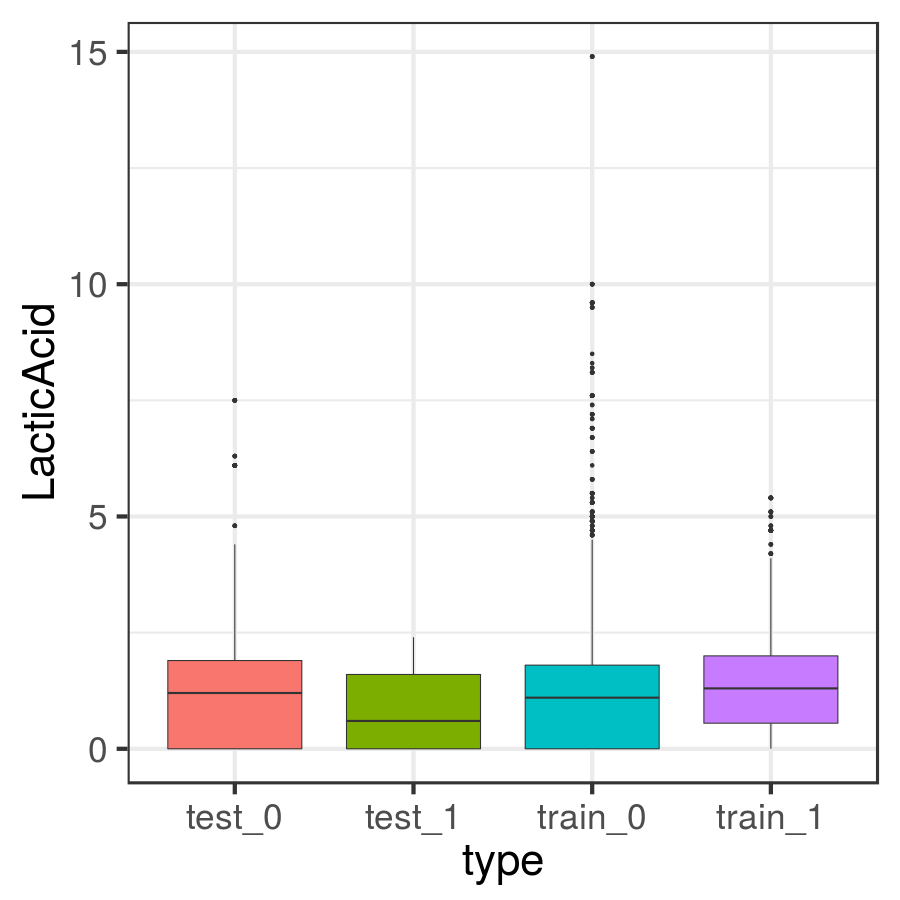}
     \caption{Attribute = \emph{LacticAcid}. Significant differences in train\_0--test\_0 ($Z=3.064$, $p=.002$) and train\_1--test\_1 ($Z=7.337$, $p<.001$).}
     \label{fig:sepsis_cases_1_boxplot_lacticacid}
 	\end{subfigure}
    \caption{Concept drift in data attributes in \emph{sepsis\_1} log. The distributions of the variables are different across the two classes in the train and the test set. Statistical significance of the differences is assessed using Wilcoxon signed-rank test.}
     \label{fig:sepsis_cases_1_boxplots}
\end{figure}

\begin{table}[hbtp]
\caption{Execution times for {\bfseries random forest}\deleted{ (1)}}
\label{table:performance_results_rf_1}
\begin{center}
\begin{adjustbox}{max width=\textwidth}
\begin{tabular}{@{}lcccccccc@{}}
\toprule
 & \multicolumn{2}{c}{{\bfseries bpic2011\_1}} & \multicolumn{2}{c}{{\bfseries bpic2011\_2}} & \multicolumn{2}{c}{{\bfseries bpic2011\_3}} \\ \cmidrule(lr){2-3} \cmidrule(lr){4-5} \cmidrule(lr){6-7}
method & offline\_total (s) & online\_avg (ms) & offline\_total (s) & online\_avg (ms) & offline\_total (s) & online\_avg (ms) \\ \midrule
single\_laststate & $42.88 \pm 0.78$ & $76 \pm 108$ & $38.81 \pm 0.27$ & $67 \pm 106$ & $34.12 \pm 0.17$ & $78 \pm 107$ \\ 
single\_agg & $51.92 \pm 0.85$ & $76 \pm 109$ & $369.75 \pm 2.56$ & $68 \pm 107$ & $57.62 \pm 1.04$ & $79 \pm 109$ \\ 
knn\_laststate & $6.24 \pm 0.24$ & $168 \pm 251$ & $9.75 \pm 0.44$ & $143 \pm 228$ & $4.15 \pm 0.09$ & $169 \pm 257$ \\ 
knn\_agg & $6.14 \pm 0.12$ & $176 \pm 264$ & $10.2 \pm 0.35$ & $151 \pm 240$ & $4.22 \pm 0.22$ & $190 \pm 288$ \\ 
state\_laststate & $112.57 \pm 0.39$ & $\bm{58 \pm 80}$ & $145.86 \pm 0.49$ & $\bm{53 \pm 82}$ & $81.7 \pm 0.19$ & $\bm{60 \pm 79}$ \\ 
state\_agg & $265.71 \pm 0.58$ & $69 \pm 95$ & $226.93 \pm 0.82$ & $64 \pm 100$ & $153.58 \pm 0.14$ & $72 \pm 95$ \\ 
cluster\_laststate & $35.88 \pm 0.23$ & $73 \pm 122$ & $73.3 \pm 0.18$ & $56 \pm 105$ & $42.61 \pm 0.12$ & $64 \pm 114$ \\ 
cluster\_agg & $211.72 \pm 0.86$ & $76 \pm 124$ & $439.57 \pm 0.68$ & $73 \pm 122$ & $58.2 \pm 0.1$ & $80 \pm 122$ \\ 
prefix\_index & $410.74 \pm 5.03$ & $126 \pm 79$ & $465.87 \pm 10.13$ & $127 \pm 71$ & $331.3 \pm 12.36$ & $121 \pm 75$ \\ 
prefix\_laststate & $98.34 \pm 0.22$ & $66 \pm 98$ & $129.95 \pm 0.5$ & $57 \pm 93$ & $61.51 \pm 0.07$ & $69 \pm 97$ \\ 
prefix\_agg & $173.76 \pm 0.29$ & $70 \pm 101$ & $189.64 \pm 0.57$ & $61 \pm 95$ & $122.13 \pm 0.15$ & $73 \pm 100$ \\ 
\bottomrule
 & \multicolumn{2}{c}{{\bfseries bpic2011\_4}} & \multicolumn{2}{c}{{\bfseries bpic2015\_1}} & \multicolumn{2}{c}{{\bfseries bpic2015\_2}} \\ \cmidrule(lr){2-3} \cmidrule(lr){4-5} \cmidrule(lr){6-7}
method & offline\_total (s) & online\_avg (ms) & offline\_total (s) & online\_avg (ms) & offline\_total (s) & online\_avg (ms) \\ \midrule
single\_laststate & $50.65 \pm 0.21$ & $68 \pm 106$ & $22.29 \pm 0.54$ & $26 \pm 40$ & $60.39 \pm 0.53$ & $24 \pm 38$ \\ 
single\_agg & $135.39 \pm 7.21$ & $68 \pm 108$ & $102.42 \pm 0.65$ & $26 \pm 41$ & $86.86 \pm 0.26$ & $24 \pm 40$ \\ 
knn\_laststate & $11.42 \pm 2.19$ & $151 \pm 236$ & $8.35 \pm 0.48$ & $126 \pm 226$ & $11.01 \pm 0.16$ & $116 \pm 225$ \\ 
knn\_agg & $9.3 \pm 0.2$ & $152 \pm 237$ & $9.33 \pm 0.13$ & $138 \pm 240$ & $11.24 \pm 0.54$ & $115 \pm 225$ \\ 
state\_laststate & $144.21 \pm 4.36$ & $\bm{54 \pm 83}$ & $107.37 \pm 0.18$ & $31 \pm 54$ & $100.87 \pm 0.27$ & $33 \pm 55$ \\ 
state\_agg & $227.05 \pm 4.78$ & $65 \pm 100$ & $135.72 \pm 0.47$ & $34 \pm 57$ & $132.69 \pm 0.32$ & $37 \pm 58$ \\ 
cluster\_laststate & $428.27 \pm 8.7$ & $75 \pm 121$ & $40.05 \pm 0.58$ & $43 \pm 69$ & $55.96 \pm 2.26$ & $40 \pm 66$ \\ 
cluster\_agg & $125.97 \pm 0.29$ & $\bm{54 \pm 87}$ & $52.37 \pm 0.83$ & $30 \pm 46$ & $73.58 \pm 2.67$ & $32 \pm 52$ \\ 
prefix\_index & $1121.44 \pm 25.31$ & $126 \pm 71$ & $504.84 \pm 0.61$ & $62 \pm 12$ & $1137.08 \pm 2.07$ & $60 \pm 14$ \\ 
prefix\_laststate & $196.68 \pm 0.02$ & $57 \pm 93$ & $70.06 \pm 0.24$ & $\bm{14 \pm 19}$ & $75.49 \pm 0.13$ & $\bm{12 \pm 16}$ \\ 
prefix\_agg & $306.88 \pm 0.64$ & $61 \pm 95$ & $97.72 \pm 0.08$ & $17 \pm 21$ & $159.75 \pm 0.25$ & $15 \pm 18$ \\ 
\bottomrule
 & \multicolumn{2}{c}{{\bfseries bpic2015\_3}} & \multicolumn{2}{c}{{\bfseries bpic2015\_4}} & \multicolumn{2}{c}{{\bfseries bpic2015\_5}} \\ \cmidrule(lr){2-3} \cmidrule(lr){4-5} \cmidrule(lr){6-7}
method & offline\_total (s) & online\_avg (ms) & offline\_total (s) & online\_avg (ms) & offline\_total (s) & online\_avg (ms) \\ \midrule
single\_laststate & $138.37 \pm 2.95$ & $26 \pm 42$ & $20.86 \pm 0.6$ & $24 \pm 36$ & $37.74 \pm 0.9$ & $22 \pm 35$ \\ 
single\_agg & $95.39 \pm 0.2$ & $27 \pm 43$ & $70.51 \pm 0.32$ & $25 \pm 38$ & $105.67 \pm 0.64$ & $23 \pm 36$ \\ 
knn\_laststate & $19.94 \pm 0.64$ & $113 \pm 226$ & $7.66 \pm 0.02$ & $120 \pm 228$ & $15.2 \pm 0.33$ & $115 \pm 221$ \\ 
knn\_agg & $19.15 \pm 0.68$ & $124 \pm 239$ & $7.77 \pm 0.35$ & $127 \pm 237$ & $18.78 \pm 0.24$ & $122 \pm 231$ \\ 
state\_laststate & $147.65 \pm 0.1$ & $35 \pm 59$ & $89.11 \pm 0.16$ & $31 \pm 51$ & $114.35 \pm 0.4$ & $29 \pm 48$ \\ 
state\_agg & $192.76 \pm 0.73$ & $39 \pm 62$ & $93.86 \pm 0.31$ & $34 \pm 54$ & $157.16 \pm 6.82$ & $32 \pm 52$ \\ 
cluster\_laststate & $121.76 \pm 0.41$ & $43 \pm 73$ & $41.47 \pm 0.19$ & $37 \pm 72$ & $72.07 \pm 0.35$ & $38 \pm 66$ \\ 
cluster\_agg & $73.09 \pm 0.4$ & $31 \pm 49$ & $31.47 \pm 0.27$ & $35 \pm 55$ & $64.21 \pm 0.58$ & $31 \pm 49$ \\ 
prefix\_index & $1531.4 \pm 15.22$ & $71 \pm 17$ & $176.56 \pm 0.22$ & $51 \pm 10$ & $366.0 \pm 2.63$ & $60 \pm 13$ \\ 
prefix\_laststate & $100.27 \pm 0.28$ & $\bm{13 \pm 18}$ & $62.28 \pm 0.02$ & $\bm{12 \pm 17}$ & $99.77 \pm 0.71$ & $\bm{12 \pm 17}$ \\ 
prefix\_agg & $172.12 \pm 4.21$ & $16 \pm 19$ & $94.56 \pm 0.16$ & $15 \pm 18$ & $156.44 \pm 3.28$ & $14 \pm 18$ \\ 
\bottomrule
 & \multicolumn{2}{c}{{\bfseries production}} & \multicolumn{2}{c}{{\bfseries insurance\_1}} & \multicolumn{2}{c}{{\bfseries insurance\_2}} \\ \cmidrule(lr){2-3} \cmidrule(lr){4-5} \cmidrule(lr){6-7}
method & offline\_total (s) & online\_avg (ms) & offline\_total (s) & online\_avg (ms) & offline\_total (s) & online\_avg (ms) \\ \midrule
single\_laststate & $2.99 \pm 0.06$ & $43 \pm 37$ & $12.51 \pm 0.09$ & $55 \pm 44$ & $9.65 \pm 0.09$ & $47 \pm 42$ \\ 
single\_agg & $8.5 \pm 0.05$ & $47 \pm 40$ & $17.82 \pm 0.12$ & $57 \pm 47$ & $11.85 \pm 0.08$ & $48 \pm 45$ \\ 
knn\_laststate & $0.89 \pm 0.05$ & $307 \pm 338$ & $0.61 \pm 0.03$ & $204 \pm 206$ & $0.89 \pm 0.04$ & $217 \pm 173$ \\ 
knn\_agg & $1.04 \pm 0.01$ & $338 \pm 333$ & $0.65 \pm 0.04$ & $223 \pm 228$ & $0.94 \pm 0.04$ & $224 \pm 178$ \\ 
state\_laststate & $17.34 \pm 0.07$ & $\bm{41 \pm 34}$ & $14.36 \pm 0.04$ & $\bm{48 \pm 35}$ & $16.63 \pm 0.01$ & $\bm{40 \pm 34}$ \\ 
state\_agg & $17.2 \pm 0.21$ & $47 \pm 40$ & $16.88 \pm 0.12$ & $57 \pm 44$ & $15.36 \pm 0.18$ & $49 \pm 43$ \\ 
cluster\_laststate & $18.89 \pm 0.26$ & $45 \pm 43$ & $10.9 \pm 0.1$ & $50 \pm 50$ & $19.69 \pm 0.06$ & $48 \pm 44$ \\ 
cluster\_agg & $15.18 \pm 0.13$ & $49 \pm 47$ & $24.89 \pm 0.09$ & $50 \pm 49$ & $31.68 \pm 0.1$ & $51 \pm 48$ \\ 
prefix\_index & $24.93 \pm 0.11$ & $70 \pm 27$ & $27.69 \pm 0.22$ & $107 \pm 10$ & $28.91 \pm 0.4$ & $103 \pm 11$ \\ 
prefix\_laststate & $18.32 \pm 0.89$ & $\bm{41 \pm 36}$ & $15.16 \pm 0.0$ & $51 \pm 37$ & $22.93 \pm 0.1$ & $43 \pm 37$ \\ 
prefix\_agg & $26.86 \pm 0.08$ & $57 \pm 48$ & $16.36 \pm 0.03$ & $\bm{48 \pm 35}$ & $24.44 \pm 0.38$ & $\bm{40 \pm 33}$ \\ 
\bottomrule
\end{tabular}
\end{adjustbox}
\end{center}
\end{table} 

\begin{table}[hbtp]
\caption{Execution times for {\bfseries random forest} (\replaced{continued}{2})}
\label{table:performance_results_rf_2}
\begin{center}
\begin{adjustbox}{max width=\textwidth}
\begin{tabular}{@{}lcccccccc@{}}
\toprule
 & \multicolumn{2}{c}{{\bfseries sepsis\_1}} & \multicolumn{2}{c}{{\bfseries sepsis\_2}} & \multicolumn{2}{c}{{\bfseries sepsis\_3}} \\ \cmidrule(lr){2-3} \cmidrule(lr){4-5} \cmidrule(lr){6-7}
method & offline\_total (s) & online\_avg (ms) & offline\_total (s) & online\_avg (ms) & offline\_total (s) & online\_avg (ms) \\ \midrule
single\_laststate & $87.17 \pm 2.69$ & $38 \pm 43$ & $10.85 \pm 0.19$ & $\bm{46 \pm 45}$ & $8.81 \pm 0.03$ & $40 \pm 44$ \\ 
single\_agg & $43.23 \pm 0.31$ & $39 \pm 45$ & $14.38 \pm 0.14$ & $49 \pm 48$ & $46.89 \pm 0.22$ & $43 \pm 46$ \\ 
knn\_laststate & $2.78 \pm 0.06$ & $242 \pm 271$ & $0.99 \pm 0.01$ & $305 \pm 301$ & $1.9 \pm 0.05$ & $261 \pm 279$ \\ 
knn\_agg & $2.79 \pm 0.06$ & $255 \pm 282$ & $1.04 \pm 0.05$ & $307 \pm 305$ & $1.89 \pm 0.04$ & $272 \pm 288$ \\ 
state\_laststate & $71.76 \pm 0.18$ & $41 \pm 46$ & $23.6 \pm 0.06$ & $50 \pm 50$ & $22.51 \pm 0.33$ & $43 \pm 46$ \\ 
state\_agg & $140.28 \pm 0.2$ & $42 \pm 47$ & $19.23 \pm 0.03$ & $51 \pm 51$ & $123.09 \pm 0.17$ & $45 \pm 48$ \\ 
cluster\_laststate & $41.22 \pm 0.15$ & $\bm{36 \pm 44}$ & $25.36 \pm 0.21$ & $49 \pm 48$ & $36.94 \pm 0.26$ & $41 \pm 45$ \\ 
cluster\_agg & $77.48 \pm 0.11$ & $38 \pm 45$ & $81.31 \pm 0.18$ & $50 \pm 50$ & $47.18 \pm 0.08$ & $43 \pm 46$ \\ 
prefix\_index & $165.12 \pm 0.03$ & $51 \pm 39$ & $114.1 \pm 0.09$ & $58 \pm 42$ & $137.86 \pm 0.18$ & $52 \pm 39$ \\ 
prefix\_laststate & $78.81 \pm 0.18$ & $39 \pm 44$ & $44.0 \pm 0.02$ & $48 \pm 46$ & $51.21 \pm 0.25$ & $41 \pm 44$ \\ 
prefix\_agg & $141.0 \pm 0.06$ & $37 \pm 41$ & $57.12 \pm 0.03$ & $\bm{46 \pm 44}$ & $73.75 \pm 0.01$ & $\bm{39 \pm 42}$ \\ 
\bottomrule
 & \multicolumn{2}{c}{{\bfseries bpic2012\_1}} & \multicolumn{2}{c}{{\bfseries bpic2012\_2}} & \multicolumn{2}{c}{{\bfseries bpic2012\_3}} \\ \cmidrule(lr){2-3} \cmidrule(lr){4-5} \cmidrule(lr){6-7}
method & offline\_total (s) & online\_avg (ms) & offline\_total (s) & online\_avg (ms) & offline\_total (s) & online\_avg (ms) \\ \midrule
single\_laststate & $258.83 \pm 4.05$ & $13 \pm 20$ & $113.69 \pm 1.3$ & $13 \pm 20$ & $221.25 \pm 0.69$ & $13 \pm 20$ \\ 
single\_agg & $395.73 \pm 2.98$ & $14 \pm 21$ & $358.35 \pm 1.11$ & $13 \pm 21$ & $389.96 \pm 7.22$ & $14 \pm 21$ \\ 
knn\_laststate & $29.02 \pm 4.34$ & $107 \pm 213$ & $30.21 \pm 0.59$ & $501 \pm 593$ & $27.96 \pm 0.55$ & $491 \pm 577$ \\ 
knn\_agg & $29.96 \pm 0.59$ & $506 \pm 592$ & $30.54 \pm 0.6$ & $509 \pm 599$ & $29.9 \pm 0.59$ & $513 \pm 605$ \\ 
state\_laststate & $139.79 \pm 1.06$ & $14 \pm 19$ & $215.16 \pm 0.66$ & $14 \pm 19$ & $129.01 \pm 0.82$ & $14 \pm 19$ \\ 
state\_agg & $697.14 \pm 6.11$ & $15 \pm 21$ & $547.46 \pm 4.22$ & $15 \pm 20$ & $394.35 \pm 6.53$ & $15 \pm 21$ \\ 
cluster\_laststate & $147.92 \pm 5.76$ & $15 \pm 23$ & $125.69 \pm 0.15$ & $16 \pm 24$ & $175.72 \pm 2.12$ & $15 \pm 22$ \\ 
cluster\_agg & $535.28 \pm 10.75$ & $16 \pm 25$ & $196.31 \pm 2.84$ & $16 \pm 24$ & $276.72 \pm 3.67$ & $15 \pm 23$ \\ 
prefix\_index & $7198.38 \pm 70.16$ & $43 \pm 10$ & $8059.82 \pm 147.53$ & $41 \pm 9$ & $4076.35 \pm 70.84$ & $43 \pm 10$ \\ 
prefix\_laststate & $277.58 \pm 0.87$ & $\bm{11 \pm 13}$ & $253.66 \pm 0.28$ & $\bm{10 \pm 12}$ & $265.56 \pm 0.7$ & $\bm{10 \pm 12}$ \\ 
prefix\_agg & $1326.15 \pm 4.76$ & $\bm{11 \pm 14}$ & $548.0 \pm 1.36$ & $11 \pm 14$ & $833.38 \pm 3.9$ & $12 \pm 15$ \\ 
\bottomrule
 & \multicolumn{2}{c}{{\bfseries bpic2017\_1}} & \multicolumn{2}{c}{{\bfseries bpic2017\_2}} & \multicolumn{2}{c}{{\bfseries bpic2017\_3}} \\ \cmidrule(lr){2-3} \cmidrule(lr){4-5} \cmidrule(lr){6-7}
method & offline\_total (s) & online\_avg (ms) & offline\_total (s) & online\_avg (ms) & offline\_total (s) & online\_avg (ms) \\ \midrule
single\_laststate & $1289.72 \pm 25.31$ & $26 \pm 32$ & $1625.87 \pm 31.91$ & $30 \pm 36$ & $3172.95 \pm 62.27$ & $26 \pm 31$ \\ 
single\_agg & $2880.68 \pm 56.53$ & $31 \pm 38$ & $4371.05 \pm 85.78$ & $27 \pm 33$ & $8649.74 \pm 169.75$ & $28 \pm 34$ \\ 
knn\_laststate & $143.14 \pm 2.81$ & $1757 \pm 1700$ & $138.09 \pm 2.71$ & $1792 \pm 1727$ & $117.39 \pm 2.3$ & $1532 \pm 1475$ \\ 
knn\_agg & $136.12 \pm 2.67$ & $1784 \pm 1715$ & $132.3 \pm 2.6$ & $1642 \pm 1582$ & $130.03 \pm 2.55$ & $1679 \pm 1612$ \\ 
state\_laststate & $1239.01 \pm 24.32$ & $25 \pm 30$ & $1247.27 \pm 24.48$ & $29 \pm 35$ & $795.83 \pm 15.62$ & $25 \pm 30$ \\ 
state\_agg & $12366.43 \pm 242.69$ & $31 \pm 34$ & $5671.91 \pm 111.31$ & $33 \pm 39$ & $13165.58 \pm 258.37$ & $27 \pm 31$ \\ 
cluster\_laststate & $2325.38 \pm 45.64$ & $\bm{24 \pm 31}$ & $1018.82 \pm 19.99$ & $25 \pm 29$ & $1367.58 \pm 26.84$ & $\bm{23 \pm 28}$ \\ 
cluster\_agg & $1535.54 \pm 30.14$ & $26 \pm 34$ & $3979.78 \pm 78.1$ & $27 \pm 32$ & $1728.18 \pm 33.92$ & $25 \pm 31$ \\ 
prefix\_index & $25283.44 \pm 496.19$ & $88 \pm 12$ & $22481.59 \pm 441.2$ & $91 \pm 12$ & $19949.41 \pm 391.51$ & $86 \pm 14$ \\ 
prefix\_laststate & $2270.56 \pm 29.0$ & $25 \pm 28$ & $789.43 \pm 15.49$ & $\bm{22 \pm 24}$ & $4245.86 \pm 60.81$ & $26 \pm 30$ \\ 
prefix\_agg & $5933.72 \pm 36.32$ & $27 \pm 32$ & $5003.87 \pm 98.2$ & $25 \pm 28$ & $10723.16 \pm 41.97$ & $\bm{23 \pm 27}$ \\ 
\bottomrule
 & \multicolumn{2}{c}{{\bfseries traffic}} & \multicolumn{2}{c}{{\bfseries hospital\_1}} & \multicolumn{2}{c}{{\bfseries hospital\_2}} \\ \cmidrule(lr){2-3} \cmidrule(lr){4-5} \cmidrule(lr){6-7}
method & offline\_total (s) & online\_avg (ms) & offline\_total (s) & online\_avg (ms) & offline\_total (s) & online\_avg (ms) \\ \midrule
single\_laststate & $2553.51 \pm 175.77$ & $\bm{86 \pm 48}$ & $5508.58 \pm 108.11$ & $401 \pm 263$ & $2974.02 \pm 58.37$ & $456 \pm 300$ \\ 
single\_agg & $2253.99 \pm 12.52$ & $94 \pm 53$ & $11667.12 \pm 228.97$ & $470 \pm 309$ & $131453.83 \pm 2579.78$ & $411 \pm 271$ \\ 
knn\_laststate & $424.08 \pm 56.11$ & $543 \pm 392$ & $123.42 \pm 2.42$ & $463 \pm 362$ & $368.05 \pm 7.22$ & $435 \pm 401$ \\ 
knn\_agg & $439.44 \pm 59.76$ & $560 \pm 404$ & $115.82 \pm 2.27$ & $497 \pm 387$ & $362.69 \pm 7.12$ & $453 \pm 419$ \\ 
state\_laststate & $1222.57 \pm 25.28$ & $94 \pm 53$ & $1329.31 \pm 26.09$ & $371 \pm 298$ & $1491.67 \pm 29.27$ & $\bm{309 \pm 255}$ \\ 
state\_agg & $1362.58 \pm 48.94$ & $97 \pm 54$ & $959.94 \pm 18.84$ & $414 \pm 270$ & $1663.75 \pm 32.65$ & $370 \pm 236$ \\ 
cluster\_laststate & $1129.09 \pm 44.03$ & $90 \pm 52$ & $2402.26 \pm 47.14$ & $\bm{341 \pm 305}$ & $1794.1 \pm 35.21$ & $332 \pm 324$ \\ 
cluster\_agg & $1261.5 \pm 3.47$ & $96 \pm 55$ & $5956.26 \pm 116.89$ & $465 \pm 305$ & $1267.46 \pm 24.87$ & $393 \pm 295$ \\ 
prefix\_index & $2051.23 \pm 27.11$ & $116 \pm 26$ & $3566.4 \pm 69.99$ & $890 \pm 90$ & $6309.37 \pm 123.82$ & $930 \pm 174$ \\ 
prefix\_laststate & $1365.6 \pm 8.31$ & $91 \pm 50$ & $1950.52 \pm 38.28$ & $402 \pm 281$ & $2085.55 \pm 16.28$ & $337 \pm 235$ \\ 
prefix\_agg & $1601.71 \pm 11.23$ & $99 \pm 55$ & $17359.15 \pm 340.67$ & $431 \pm 251$ & $7652.09 \pm 2.19$ & $374 \pm 219$ \\ 
\bottomrule
\end{tabular}
\end{adjustbox}
\end{center}
\end{table}

\begin{table}[hbtp]
\caption{Execution times for {\bfseries logistic regression}\deleted{ (1)}}
\label{table:performance_results_logit_1}
\begin{center}
\begin{adjustbox}{max width=\textwidth}
\begin{tabular}{@{}lcccccccc@{}}
\toprule
 & \multicolumn{2}{c}{{\bfseries bpic2011\_1}} & \multicolumn{2}{c}{{\bfseries bpic2011\_2}} & \multicolumn{2}{c}{{\bfseries bpic2011\_3}} \\ \cmidrule(lr){2-3} \cmidrule(lr){4-5} \cmidrule(lr){6-7}
method & offline\_total (s) & online\_avg (ms) & offline\_total (s) & online\_avg (ms) & offline\_total (s) & online\_avg (ms) \\ \midrule
single\_laststate & $8.14 \pm 0.29$ & $68 \pm 97$ & $11.09 \pm 0.4$ & $61 \pm 96$ & $5.55 \pm 0.13$ & $70 \pm 96$ \\ 
single\_agg & $11.85 \pm 0.39$ & $69 \pm 99$ & $16.89 \pm 0.17$ & $62 \pm 98$ & $6.58 \pm 0.05$ & $71 \pm 97$ \\ 
knn\_laststate & $5.69 \pm 0.03$ & $\bm{29 \pm 42}$ & $9.86 \pm 0.22$ & $\bm{23 \pm 37}$ & $4.12 \pm 0.12$ & $\bm{29 \pm 41}$ \\ 
knn\_agg & $5.93 \pm 0.14$ & $32 \pm 46$ & $9.69 \pm 0.05$ & $26 \pm 42$ & $4.28 \pm 0.13$ & $\bm{29 \pm 43}$ \\ 
state\_laststate & $8.64 \pm 0.1$ & $51 \pm 70$ & $11.97 \pm 0.15$ & $47 \pm 73$ & $6.57 \pm 0.01$ & $52 \pm 68$ \\ 
state\_agg & $11.49 \pm 0.05$ & $66 \pm 91$ & $15.76 \pm 0.59$ & $57 \pm 88$ & $7.77 \pm 0.01$ & $61 \pm 80$ \\ 
cluster\_laststate & $19.61 \pm 0.1$ & $56 \pm 102$ & $27.12 \pm 0.79$ & $53 \pm 100$ & $16.11 \pm 0.07$ & $55 \pm 102$ \\ 
cluster\_agg & $19.1 \pm 0.7$ & $53 \pm 76$ & $27.15 \pm 0.2$ & $58 \pm 109$ & $13.86 \pm 0.21$ & $65 \pm 107$ \\ 
prefix\_index & $28.51 \pm 0.71$ & $122 \pm 71$ & $44.66 \pm 2.52$ & $124 \pm 65$ & $18.37 \pm 0.5$ & $106 \pm 62$ \\ 
prefix\_laststate & $7.52 \pm 0.13$ & $58 \pm 87$ & $10.34 \pm 0.12$ & $51 \pm 83$ & $5.32 \pm 0.07$ & $60 \pm 85$ \\ 
prefix\_agg & $9.28 \pm 0.11$ & $61 \pm 88$ & $13.22 \pm 0.14$ & $54 \pm 83$ & $7.34 \pm 0.08$ & $64 \pm 86$ \\ 
\bottomrule
 & \multicolumn{2}{c}{{\bfseries bpic2011\_4}} & \multicolumn{2}{c}{{\bfseries bpic2015\_1}} & \multicolumn{2}{c}{{\bfseries bpic2015\_2}} \\ \cmidrule(lr){2-3} \cmidrule(lr){4-5} \cmidrule(lr){6-7}
method & offline\_total (s) & online\_avg (ms) & offline\_total (s) & online\_avg (ms) & offline\_total (s) & online\_avg (ms) \\ \midrule
single\_laststate & $11.13 \pm 0.71$ & $62 \pm 97$ & $7.97 \pm 0.29$ & $20 \pm 30$ & $7.04 \pm 0.05$ & $18 \pm 29$ \\ 
single\_agg & $20.21 \pm 2.74$ & $62 \pm 98$ & $16.57 \pm 0.41$ & $22 \pm 34$ & $48.14 \pm 0.66$ & $20 \pm 32$ \\ 
knn\_laststate & $8.55 \pm 0.43$ & $\bm{26 \pm 40}$ & $8.98 \pm 0.07$ & $23 \pm 37$ & $10.59 \pm 0.04$ & $24 \pm 39$ \\ 
knn\_agg & $8.56 \pm 0.36$ & $\bm{26 \pm 41}$ & $8.16 \pm 0.27$ & $24 \pm 39$ & $11.9 \pm 0.11$ & $26 \pm 45$ \\ 
state\_laststate & $11.71 \pm 0.08$ & $47 \pm 72$ & $8.52 \pm 0.2$ & $26 \pm 45$ & $9.65 \pm 0.2$ & $27 \pm 46$ \\ 
state\_agg & $15.1 \pm 0.04$ & $55 \pm 85$ & $9.8 \pm 0.36$ & $29 \pm 48$ & $11.07 \pm 0.08$ & $31 \pm 49$ \\ 
cluster\_laststate & $25.99 \pm 0.08$ & $54 \pm 106$ & $24.07 \pm 0.07$ & $22 \pm 35$ & $37.74 \pm 0.67$ & $36 \pm 61$ \\ 
cluster\_agg & $26.49 \pm 0.23$ & $68 \pm 112$ & $19.51 \pm 0.27$ & $33 \pm 52$ & $27.18 \pm 0.34$ & $27 \pm 45$ \\ 
prefix\_index & $41.41 \pm 0.35$ & $118 \pm 60$ & $27.99 \pm 0.6$ & $56 \pm 12$ & $38.47 \pm 0.45$ & $54 \pm 15$ \\ 
prefix\_laststate & $10.81 \pm 0.11$ & $51 \pm 82$ & $7.09 \pm 0.06$ & $\bm{7 \pm 9}$ & $7.67 \pm 0.28$ & $\bm{6 \pm 6}$ \\ 
prefix\_agg & $14.14 \pm 0.08$ & $54 \pm 84$ & $7.53 \pm 0.03$ & $8 \pm 9$ & $9.77 \pm 0.07$ & $7 \pm 7$ \\ 
\bottomrule
 & \multicolumn{2}{c}{{\bfseries bpic2015\_3}} & \multicolumn{2}{c}{{\bfseries bpic2015\_4}} & \multicolumn{2}{c}{{\bfseries bpic2015\_5}} \\ \cmidrule(lr){2-3} \cmidrule(lr){4-5} \cmidrule(lr){6-7}
method & offline\_total (s) & online\_avg (ms) & offline\_total (s) & online\_avg (ms) & offline\_total (s) & online\_avg (ms) \\ \midrule
single\_laststate & $38.41 \pm 0.18$ & $21 \pm 33$ & $6.57 \pm 0.09$ & $18 \pm 27$ & $18.67 \pm 0.14$ & $17 \pm 26$ \\ 
single\_agg & $29.69 \pm 0.72$ & $21 \pm 34$ & $9.17 \pm 0.24$ & $18 \pm 29$ & $24.9 \pm 0.24$ & $17 \pm 27$ \\ 
knn\_laststate & $19.6 \pm 0.62$ & $26 \pm 42$ & $7.6 \pm 0.34$ & $21 \pm 35$ & $14.0 \pm 0.14$ & $20 \pm 33$ \\ 
knn\_agg & $18.5 \pm 0.67$ & $31 \pm 49$ & $7.76 \pm 0.36$ & $23 \pm 38$ & $14.56 \pm 0.54$ & $26 \pm 42$ \\ 
state\_laststate & $14.56 \pm 0.3$ & $30 \pm 50$ & $7.17 \pm 0.08$ & $26 \pm 42$ & $12.4 \pm 0.09$ & $23 \pm 39$ \\ 
state\_agg & $18.28 \pm 0.07$ & $33 \pm 53$ & $8.37 \pm 0.09$ & $29 \pm 45$ & $14.39 \pm 0.3$ & $26 \pm 42$ \\ 
cluster\_laststate & $60.56 \pm 1.47$ & $36 \pm 63$ & $27.27 \pm 0.18$ & $34 \pm 61$ & $38.51 \pm 0.45$ & $18 \pm 31$ \\ 
cluster\_agg & $44.96 \pm 0.59$ & $26 \pm 41$ & $26.35 \pm 0.24$ & $33 \pm 60$ & $48.48 \pm 0.84$ & $25 \pm 40$ \\ 
prefix\_index & $65.17 \pm 0.35$ & $64 \pm 19$ & $25.3 \pm 0.3$ & $43 \pm 8$ & $46.49 \pm 0.74$ & $53 \pm 13$ \\ 
prefix\_laststate & $12.02 \pm 0.08$ & $\bm{7 \pm 8}$ & $6.37 \pm 0.07$ & $\bm{6 \pm 7}$ & $11.19 \pm 0.02$ & $\bm{6 \pm 7}$ \\ 
prefix\_agg & $17.16 \pm 0.1$ & $8 \pm 8$ & $6.65 \pm 0.04$ & $7 \pm 7$ & $12.4 \pm 0.13$ & $7 \pm 7$ \\ 
\bottomrule
 & \multicolumn{2}{c}{{\bfseries production}} & \multicolumn{2}{c}{{\bfseries insurance\_1}} & \multicolumn{2}{c}{{\bfseries insurance\_2}} \\ \cmidrule(lr){2-3} \cmidrule(lr){4-5} \cmidrule(lr){6-7}
method & offline\_total (s) & online\_avg (ms) & offline\_total (s) & online\_avg (ms) & offline\_total (s) & online\_avg (ms) \\ \midrule
single\_laststate & $0.89 \pm 0.01$ & $25 \pm 21$ & $1.41 \pm 0.01$ & $37 \pm 30$ & $2.88 \pm 0.01$ & $32 \pm 29$ \\ 
single\_agg & $1.11 \pm 0.13$ & $28 \pm 25$ & $1.04 \pm 0.05$ & $39 \pm 33$ & $2.68 \pm 0.09$ & $34 \pm 32$ \\ 
knn\_laststate & $0.87 \pm 0.03$ & $31 \pm 28$ & $0.62 \pm 0.03$ & $\bm{20 \pm 18}$ & $0.92 \pm 0.01$ & $\bm{24 \pm 16}$ \\ 
knn\_agg & $0.93 \pm 0.05$ & $31 \pm 31$ & $0.63 \pm 0.01$ & $24 \pm 21$ & $0.89 \pm 0.01$ & $27 \pm 18$ \\ 
state\_laststate & $1.47 \pm 0.12$ & $\bm{23 \pm 19}$ & $1.29 \pm 0.04$ & $31 \pm 22$ & $2.43 \pm 0.01$ & $27 \pm 21$ \\ 
state\_agg & $1.68 \pm 0.06$ & $30 \pm 26$ & $1.54 \pm 0.03$ & $43 \pm 32$ & $2.03 \pm 0.02$ & $37 \pm 32$ \\ 
cluster\_laststate & $5.16 \pm 0.08$ & $28 \pm 29$ & $4.96 \pm 0.08$ & $32 \pm 31$ & $5.34 \pm 0.13$ & $34 \pm 31$ \\ 
cluster\_agg & $5.72 \pm 0.05$ & $33 \pm 34$ & $5.83 \pm 0.11$ & $27 \pm 28$ & $8.5 \pm 0.11$ & $36 \pm 34$ \\ 
prefix\_index & $3.03 \pm 0.08$ & $51 \pm 10$ & $2.89 \pm 0.02$ & $89 \pm 5$ & $3.86 \pm 0.03$ & $90 \pm 4$ \\ 
prefix\_laststate & $1.15 \pm 0.01$ & $\bm{23 \pm 19}$ & $1.26 \pm 0.0$ & $33 \pm 23$ & $2.33 \pm 0.01$ & $28 \pm 22$ \\ 
prefix\_agg & $1.67 \pm 0.0$ & $34 \pm 28$ & $1.29 \pm 0.01$ & $32 \pm 22$ & $1.75 \pm 0.01$ & $27 \pm 21$ \\ 
\bottomrule
\end{tabular}
\end{adjustbox}
\end{center}
\end{table} 

\begin{table}[hbtp]
\caption{Execution times for {\bfseries logistic regression} (\replaced{continued}{2})}
\label{table:performance_results_logit_2}
\begin{center}
\begin{adjustbox}{max width=\textwidth}
\begin{tabular}{@{}lcccccccc@{}}
\toprule
 & \multicolumn{2}{c}{{\bfseries sepsis\_1}} & \multicolumn{2}{c}{{\bfseries sepsis\_2}} & \multicolumn{2}{c}{{\bfseries sepsis\_3}} \\ \cmidrule(lr){2-3} \cmidrule(lr){4-5} \cmidrule(lr){6-7}
method & offline\_total (s) & online\_avg (ms) & offline\_total (s) & online\_avg (ms) & offline\_total (s) & online\_avg (ms) \\ \midrule
single\_laststate & $3.21 \pm 0.15$ & $27 \pm 31$ & $1.31 \pm 0.03$ & $\bm{33 \pm 33}$ & $2.3 \pm 0.04$ & $29 \pm 31$ \\ 
single\_agg & $21.44 \pm 0.28$ & $29 \pm 33$ & $1.48 \pm 0.03$ & $35 \pm 35$ & $2.54 \pm 0.04$ & $31 \pm 33$ \\ 
knn\_laststate & $2.82 \pm 0.06$ & $28 \pm 32$ & $0.98 \pm 0.04$ & $35 \pm 35$ & $1.97 \pm 0.05$ & $31 \pm 33$ \\ 
knn\_agg & $2.88 \pm 0.11$ & $31 \pm 34$ & $1.05 \pm 0.06$ & $40 \pm 39$ & $1.85 \pm 0.06$ & $34 \pm 36$ \\ 
state\_laststate & $5.74 \pm 0.17$ & $29 \pm 33$ & $1.88 \pm 0.03$ & $36 \pm 36$ & $3.04 \pm 0.02$ & $31 \pm 34$ \\ 
state\_agg & $22.54 \pm 0.11$ & $32 \pm 36$ & $2.1 \pm 0.02$ & $39 \pm 39$ & $3.18 \pm 0.02$ & $35 \pm 37$ \\ 
cluster\_laststate & $9.51 \pm 0.11$ & $\bm{26 \pm 32}$ & $5.52 \pm 0.18$ & $35 \pm 34$ & $6.52 \pm 0.15$ & $31 \pm 34$ \\ 
cluster\_agg & $11.73 \pm 0.04$ & $27 \pm 33$ & $4.98 \pm 0.09$ & $38 \pm 38$ & $7.58 \pm 0.03$ & $30 \pm 34$ \\ 
prefix\_index & $16.74 \pm 0.07$ & $39 \pm 26$ & $3.14 \pm 0.02$ & $45 \pm 28$ & $6.1 \pm 0.02$ & $40 \pm 26$ \\ 
prefix\_laststate & $4.8 \pm 0.07$ & $27 \pm 30$ & $2.1 \pm 0.07$ & $\bm{33 \pm 32}$ & $3.25 \pm 0.03$ & $\bm{28 \pm 30}$ \\ 
prefix\_agg & $8.01 \pm 0.07$ & $28 \pm 32$ & $2.27 \pm 0.05$ & $35 \pm 34$ & $3.64 \pm 0.06$ & $30 \pm 32$ \\ 
\bottomrule
 & \multicolumn{2}{c}{{\bfseries bpic2012\_1}} & \multicolumn{2}{c}{{\bfseries bpic2012\_2}} & \multicolumn{2}{c}{{\bfseries bpic2012\_3}} \\ \cmidrule(lr){2-3} \cmidrule(lr){4-5} \cmidrule(lr){6-7}
method & offline\_total (s) & online\_avg (ms) & offline\_total (s) & online\_avg (ms) & offline\_total (s) & online\_avg (ms) \\ \midrule
single\_laststate & $30.51 \pm 0.38$ & $7 \pm 10$ & $29.15 \pm 0.79$ & $7 \pm 10$ & $30.31 \pm 1.76$ & $7 \pm 10$ \\ 
single\_agg & $31.1 \pm 0.27$ & $8 \pm 12$ & $60.4 \pm 3.11$ & $8 \pm 12$ & $74.28 \pm 0.31$ & $8 \pm 12$ \\ 
knn\_laststate & $30.24 \pm 0.38$ & $55 \pm 62$ & $28.85 \pm 0.08$ & $27 \pm 30$ & $27.84 \pm 0.83$ & $74 \pm 82$ \\ 
knn\_agg & $29.06 \pm 0.46$ & $350 \pm 374$ & $28.32 \pm 0.08$ & $310 \pm 333$ & $31.05 \pm 2.0$ & $192 \pm 208$ \\ 
state\_laststate & $24.21 \pm 0.29$ & $8 \pm 10$ & $24.41 \pm 0.1$ & $8 \pm 10$ & $23.68 \pm 0.37$ & $8 \pm 10$ \\ 
state\_agg & $29.84 \pm 0.39$ & $10 \pm 12$ & $28.8 \pm 1.11$ & $9 \pm 11$ & $30.25 \pm 0.17$ & $9 \pm 12$ \\ 
cluster\_laststate & $38.2 \pm 0.11$ & $10 \pm 15$ & $37.28 \pm 0.57$ & $9 \pm 13$ & $39.4 \pm 0.72$ & $9 \pm 13$ \\ 
cluster\_agg & $108.75 \pm 1.51$ & $10 \pm 17$ & $64.86 \pm 2.16$ & $9 \pm 14$ & $79.46 \pm 6.91$ & $9 \pm 14$ \\ 
prefix\_index & $67.45 \pm 2.49$ & $36 \pm 12$ & $65.61 \pm 1.1$ & $34 \pm 11$ & $62.86 \pm 1.03$ & $36 \pm 12$ \\ 
prefix\_laststate & $26.0 \pm 0.87$ & $\bm{4 \pm 3}$ & $25.84 \pm 0.39$ & $\bm{4 \pm 3}$ & $23.42 \pm 0.3$ & $\bm{4 \pm 3}$ \\ 
prefix\_agg & $38.74 \pm 0.91$ & $5 \pm 5$ & $42.84 \pm 0.41$ & $5 \pm 5$ & $47.5 \pm 0.28$ & $5 \pm 5$ \\ 
\bottomrule
 & \multicolumn{2}{c}{{\bfseries bpic2017\_1}} & \multicolumn{2}{c}{{\bfseries bpic2017\_2}} & \multicolumn{2}{c}{{\bfseries bpic2017\_3}} \\ \cmidrule(lr){2-3} \cmidrule(lr){4-5} \cmidrule(lr){6-7}
method & offline\_total (s) & online\_avg (ms) & offline\_total (s) & online\_avg (ms) & offline\_total (s) & online\_avg (ms) \\ \midrule
single\_laststate & $88.86 \pm 1.74$ & $19 \pm 23$ & $155.13 \pm 3.04$ & $19 \pm 23$ & $107.82 \pm 2.12$ & $18 \pm 21$ \\ 
single\_agg & $213.25 \pm 4.18$ & $21 \pm 26$ & $143.2 \pm 2.81$ & $21 \pm 27$ & $209.74 \pm 4.12$ & $19 \pm 23$ \\ 
knn\_laststate & $137.49 \pm 2.7$ & $1575 \pm 1484$ & $129.35 \pm 2.54$ & $1518 \pm 1426$ & $135.54 \pm 2.66$ & $1567 \pm 1473$ \\ 
knn\_agg & $127.0 \pm 2.49$ & $1474 \pm 1387$ & $129.12 \pm 2.53$ & $1524 \pm 1432$ & $121.37 \pm 2.38$ & $1413 \pm 1326$ \\ 
state\_laststate & $92.96 \pm 1.82$ & $19 \pm 20$ & $74.44 \pm 1.46$ & $18 \pm 24$ & $84.76 \pm 1.66$ & $17 \pm 20$ \\ 
state\_agg & $472.57 \pm 9.27$ & $20 \pm 22$ & $244.76 \pm 4.8$ & $22 \pm 24$ & $114.88 \pm 2.25$ & $19 \pm 22$ \\ 
cluster\_laststate & $154.87 \pm 3.04$ & $21 \pm 25$ & $172.13 \pm 3.38$ & $\bm{15 \pm 18}$ & $104.51 \pm 2.05$ & $16 \pm 19$ \\ 
cluster\_agg & $368.51 \pm 7.23$ & $19 \pm 24$ & $431.13 \pm 8.46$ & $17 \pm 21$ & $163.87 \pm 3.22$ & $16 \pm 20$ \\ 
prefix\_index & $539.24 \pm 10.58$ & $72 \pm 9$ & $512.71 \pm 10.06$ & $72 \pm 9$ & $340.34 \pm 6.68$ & $72 \pm 9$ \\ 
prefix\_laststate & $95.98 \pm 0.95$ & $\bm{15 \pm 16}$ & $112.1 \pm 5.26$ & $\bm{15 \pm 16}$ & $75.28 \pm 0.25$ & $\bm{14 \pm 15}$ \\ 
prefix\_agg & $281.04 \pm 0.84$ & $\bm{15 \pm 17}$ & $551.86 \pm 26.98$ & $17 \pm 20$ & $325.93 \pm 0.41$ & $16 \pm 17$ \\ 
\bottomrule
 & \multicolumn{2}{c}{{\bfseries traffic}} & \multicolumn{2}{c}{{\bfseries hospital\_1}} & \multicolumn{2}{c}{{\bfseries hospital\_2}} \\ \cmidrule(lr){2-3} \cmidrule(lr){4-5} \cmidrule(lr){6-7}
method & offline\_total (s) & online\_avg (ms) & offline\_total (s) & online\_avg (ms) & offline\_total (s) & online\_avg (ms) \\ \midrule
single\_laststate & $95.22 \pm 0.54$ & $\bm{62 \pm 34}$ & $171.88 \pm 3.37$ & $381 \pm 250$ & $105.77 \pm 2.08$ & $395 \pm 259$ \\ 
single\_agg & $273.55 \pm 1.81$ & $66 \pm 37$ & $361.41 \pm 7.09$ & $405 \pm 267$ & $269.08 \pm 5.28$ & $397 \pm 262$ \\ 
knn\_laststate & $384.89 \pm 5.26$ & $75 \pm 42$ & $108.0 \pm 0.13$ & $\bm{43 \pm 24}$ & $361.86 \pm 7.1$ & $\bm{88 \pm 64}$ \\ 
knn\_agg & $361.35 \pm 8.72$ & $81 \pm 51$ & $110.14 \pm 1.45$ & $78 \pm 54$ & $352.76 \pm 6.92$ & $94 \pm 65$ \\ 
state\_laststate & $68.11 \pm 0.53$ & $66 \pm 37$ & $274.97 \pm 5.4$ & $349 \pm 287$ & $144.61 \pm 2.84$ & $334 \pm 282$ \\ 
state\_agg & $151.29 \pm 4.43$ & $75 \pm 42$ & $274.89 \pm 5.39$ & $368 \pm 230$ & $308.15 \pm 6.05$ & $402 \pm 249$ \\ 
cluster\_laststate & $93.99 \pm 6.84$ & $66 \pm 39$ & $225.33 \pm 4.42$ & $330 \pm 284$ & $124.98 \pm 2.45$ & $280 \pm 244$ \\ 
cluster\_agg & $111.74 \pm 2.5$ & $68 \pm 39$ & $259.91 \pm 5.1$ & $386 \pm 264$ & $681.42 \pm 13.37$ & $369 \pm 246$ \\ 
prefix\_index & $331.76 \pm 1.14$ & $89 \pm 11$ & $263.88 \pm 5.18$ & $964 \pm 112$ & $229.74 \pm 4.51$ & $848 \pm 100$ \\ 
prefix\_laststate & $80.76 \pm 0.95$ & $\bm{62 \pm 34}$ & $122.85 \pm 1.04$ & $363 \pm 262$ & $112.29 \pm 0.47$ & $313 \pm 223$ \\ 
prefix\_agg & $186.94 \pm 2.77$ & $63 \pm 35$ & $166.22 \pm 5.34$ & $394 \pm 232$ & $173.75 \pm 0.52$ & $353 \pm 204$ \\ 
\bottomrule
\end{tabular}
\end{adjustbox}
\end{center}
\end{table}

\begin{table}[hbtp]
\caption{Execution times for {\bfseries SVM}\deleted{ (1)}}
\label{table:performance_results_svm_1}
\begin{center}
\begin{adjustbox}{max width=\textwidth}
\begin{tabular}{@{}lcccccccc@{}}
\toprule
 & \multicolumn{2}{c}{{\bfseries bpic2011\_1}} & \multicolumn{2}{c}{{\bfseries bpic2011\_2}} & \multicolumn{2}{c}{{\bfseries bpic2011\_3}} \\ \cmidrule(lr){2-3} \cmidrule(lr){4-5} \cmidrule(lr){6-7}
method & offline\_total (s) & online\_avg (ms) & offline\_total (s) & online\_avg (ms) & offline\_total (s) & online\_avg (ms) \\ \midrule
single\_laststate & $503.86 \pm 1.75$ & $70 \pm 100$ & $457.58 \pm 4.1$ & $64 \pm 100$ & $173.03 \pm 1.95$ & $73 \pm 99$ \\ 
single\_agg & $40.4 \pm 0.35$ & $69 \pm 100$ & $381.13 \pm 0.35$ & $63 \pm 100$ & $156.46 \pm 0.44$ & $73 \pm 100$ \\ 
knn\_laststate & $6.0 \pm 0.22$ & $\bm{29 \pm 42}$ & $9.79 \pm 0.44$ & $\bm{27 \pm 42}$ & $4.26 \pm 0.23$ & $\bm{29 \pm 41}$ \\ 
knn\_agg & $5.91 \pm 0.26$ & $\bm{29 \pm 42}$ & $9.98 \pm 0.28$ & $28 \pm 44$ & $4.25 \pm 0.25$ & $31 \pm 44$ \\ 
state\_laststate & $12.44 \pm 0.07$ & $52 \pm 72$ & $19.77 \pm 0.13$ & $49 \pm 75$ & $9.52 \pm 0.04$ & $53 \pm 70$ \\ 
state\_agg & $16.44 \pm 0.08$ & $61 \pm 84$ & $21.67 \pm 0.04$ & $58 \pm 90$ & $10.44 \pm 0.08$ & $63 \pm 83$ \\ 
cluster\_laststate & $35.72 \pm 0.22$ & $62 \pm 110$ & $44.74 \pm 0.14$ & $45 \pm 75$ & $18.37 \pm 0.28$ & $48 \pm 76$ \\ 
cluster\_agg & $30.12 \pm 0.1$ & $57 \pm 99$ & $46.53 \pm 0.29$ & $56 \pm 106$ & $21.49 \pm 0.11$ & $50 \pm 74$ \\ 
prefix\_index & $39.94 \pm 0.49$ & $123 \pm 72$ & $64.98 \pm 3.13$ & $125 \pm 65$ & $25.33 \pm 0.46$ & $107 \pm 62$ \\ 
prefix\_laststate & $16.31 \pm 0.17$ & $60 \pm 90$ & $23.17 \pm 0.1$ & $52 \pm 84$ & $11.99 \pm 0.04$ & $62 \pm 87$ \\ 
prefix\_agg & $17.82 \pm 0.08$ & $62 \pm 89$ & $24.89 \pm 0.04$ & $54 \pm 84$ & $11.54 \pm 0.1$ & $65 \pm 88$ \\ 
\bottomrule
 & \multicolumn{2}{c}{{\bfseries bpic2011\_4}} & \multicolumn{2}{c}{{\bfseries bpic2015\_1}} & \multicolumn{2}{c}{{\bfseries bpic2015\_2}} \\ \cmidrule(lr){2-3} \cmidrule(lr){4-5} \cmidrule(lr){6-7}
method & offline\_total (s) & online\_avg (ms) & offline\_total (s) & online\_avg (ms) & offline\_total (s) & online\_avg (ms) \\ \midrule
single\_laststate & $549.47 \pm 23.94$ & $65 \pm 102$ & $86.12 \pm 0.31$ & $21 \pm 33$ & $111.39 \pm 0.15$ & $20 \pm 32$ \\ 
single\_agg & $130.1 \pm 15.04$ & $63 \pm 99$ & $68.34 \pm 6.06$ & $22 \pm 34$ & $88.97 \pm 1.6$ & $21 \pm 33$ \\ 
knn\_laststate & $8.93 \pm 0.43$ & $\bm{23 \pm 35}$ & $8.32 \pm 0.43$ & $23 \pm 37$ & $11.31 \pm 0.37$ & $23 \pm 39$ \\ 
knn\_agg & $8.84 \pm 0.41$ & $26 \pm 40$ & $8.4 \pm 0.33$ & $21 \pm 38$ & $11.4 \pm 0.34$ & $18 \pm 28$ \\ 
state\_laststate & $18.98 \pm 0.03$ & $48 \pm 74$ & $8.87 \pm 0.11$ & $26 \pm 45$ & $9.74 \pm 0.07$ & $28 \pm 46$ \\ 
state\_agg & $25.72 \pm 0.83$ & $58 \pm 89$ & $10.33 \pm 0.21$ & $28 \pm 47$ & $11.49 \pm 0.15$ & $31 \pm 48$ \\ 
cluster\_laststate & $46.62 \pm 0.44$ & $51 \pm 103$ & $22.02 \pm 0.1$ & $22 \pm 34$ & $29.91 \pm 0.11$ & $25 \pm 41$ \\ 
cluster\_agg & $43.22 \pm 0.12$ & $49 \pm 92$ & $25.47 \pm 0.13$ & $25 \pm 39$ & $33.13 \pm 0.22$ & $25 \pm 40$ \\ 
prefix\_index & $60.85 \pm 0.68$ & $119 \pm 60$ & $34.55 \pm 0.4$ & $52 \pm 12$ & $41.09 \pm 0.26$ & $52 \pm 14$ \\ 
prefix\_laststate & $21.82 \pm 0.04$ & $52 \pm 85$ & $7.45 \pm 0.05$ & $\bm{7 \pm 9}$ & $9.34 \pm 0.1$ & $\bm{6 \pm 6}$ \\ 
prefix\_agg & $22.69 \pm 0.19$ & $54 \pm 85$ & $9.16 \pm 0.12$ & $9 \pm 10$ & $10.71 \pm 0.04$ & $8 \pm 7$ \\ 
\bottomrule
 & \multicolumn{2}{c}{{\bfseries bpic2015\_3}} & \multicolumn{2}{c}{{\bfseries bpic2015\_4}} & \multicolumn{2}{c}{{\bfseries bpic2015\_5}} \\ \cmidrule(lr){2-3} \cmidrule(lr){4-5} \cmidrule(lr){6-7}
method & offline\_total (s) & online\_avg (ms) & offline\_total (s) & online\_avg (ms) & offline\_total (s) & online\_avg (ms) \\ \midrule
single\_laststate & $363.33 \pm 16.46$ & $21 \pm 32$ & $45.01 \pm 0.31$ & $17 \pm 27$ & $567.21 \pm 11.07$ & $16 \pm 25$ \\ 
single\_agg & $178.69 \pm 12.55$ & $21 \pm 34$ & $28.26 \pm 0.59$ & $18 \pm 28$ & $144.11 \pm 1.39$ & $17 \pm 27$ \\ 
knn\_laststate & $18.82 \pm 0.62$ & $21 \pm 37$ & $7.6 \pm 0.05$ & $20 \pm 35$ & $14.34 \pm 0.32$ & $19 \pm 32$ \\ 
knn\_agg & $19.6 \pm 0.81$ & $26 \pm 45$ & $7.07 \pm 0.11$ & $23 \pm 38$ & $14.37 \pm 0.04$ & $19 \pm 34$ \\ 
state\_laststate & $15.56 \pm 0.21$ & $30 \pm 50$ & $7.3 \pm 0.05$ & $26 \pm 43$ & $13.02 \pm 0.16$ & $23 \pm 39$ \\ 
state\_agg & $19.63 \pm 0.09$ & $34 \pm 54$ & $9.29 \pm 0.07$ & $29 \pm 46$ & $16.39 \pm 0.11$ & $27 \pm 43$ \\ 
cluster\_laststate & $51.08 \pm 0.33$ & $25 \pm 39$ & $20.86 \pm 0.1$ & $21 \pm 33$ & $43.79 \pm 0.08$ & $21 \pm 35$ \\ 
cluster\_agg & $52.3 \pm 1.32$ & $25 \pm 40$ & $19.61 \pm 0.3$ & $24 \pm 37$ & $140.02 \pm 1.08$ & $25 \pm 41$ \\ 
prefix\_index & $77.94 \pm 0.32$ & $61 \pm 18$ & $27.1 \pm 0.09$ & $42 \pm 8$ & $66.81 \pm 1.85$ & $50 \pm 12$ \\ 
prefix\_laststate & $16.14 \pm 0.1$ & $\bm{7 \pm 8}$ & $6.4 \pm 0.12$ & $\bm{6 \pm 7}$ & $13.36 \pm 0.18$ & $\bm{6 \pm 7}$ \\ 
prefix\_agg & $22.12 \pm 0.07$ & $9 \pm 9$ & $9.14 \pm 0.03$ & $10 \pm 10$ & $22.26 \pm 0.07$ & $10 \pm 10$ \\ 
\bottomrule
 & \multicolumn{2}{c}{{\bfseries production}} & \multicolumn{2}{c}{{\bfseries insurance\_1}} & \multicolumn{2}{c}{{\bfseries insurance\_2}} \\ \cmidrule(lr){2-3} \cmidrule(lr){4-5} \cmidrule(lr){6-7}
method & offline\_total (s) & online\_avg (ms) & offline\_total (s) & online\_avg (ms) & offline\_total (s) & online\_avg (ms) \\ \midrule
single\_laststate & $1.46 \pm 0.05$ & $26 \pm 22$ & $4.42 \pm 0.05$ & $38 \pm 30$ & $8.06 \pm 0.05$ & $33 \pm 30$ \\ 
single\_agg & $1.48 \pm 0.05$ & $28 \pm 24$ & $3.63 \pm 0.05$ & $39 \pm 33$ & $7.16 \pm 0.09$ & $34 \pm 32$ \\ 
knn\_laststate & $0.88 \pm 0.05$ & $26 \pm 25$ & $0.63 \pm 0.02$ & $\bm{19 \pm 17}$ & $0.9 \pm 0.02$ & $\bm{22 \pm 15}$ \\ 
knn\_agg & $0.93 \pm 0.05$ & $33 \pm 31$ & $0.64 \pm 0.01$ & $22 \pm 44$ & $1.02 \pm 0.07$ & $27 \pm 18$ \\ 
state\_laststate & $1.32 \pm 0.06$ & $\bm{22 \pm 19}$ & $1.64 \pm 0.03$ & $31 \pm 22$ & $2.37 \pm 0.01$ & $27 \pm 21$ \\ 
state\_agg & $1.67 \pm 0.1$ & $30 \pm 26$ & $1.78 \pm 0.03$ & $43 \pm 33$ & $3.01 \pm 0.02$ & $37 \pm 32$ \\ 
cluster\_laststate & $4.84 \pm 0.14$ & $31 \pm 29$ & $5.48 \pm 0.09$ & $24 \pm 25$ & $6.56 \pm 0.07$ & $27 \pm 26$ \\ 
cluster\_agg & $5.04 \pm 0.13$ & $37 \pm 34$ & $5.37 \pm 0.1$ & $39 \pm 37$ & $6.83 \pm 0.03$ & $30 \pm 26$ \\ 
prefix\_index & $3.5 \pm 0.01$ & $61 \pm 12$ & $2.85 \pm 0.03$ & $89 \pm 4$ & $5.98 \pm 0.04$ & $89 \pm 4$ \\ 
prefix\_laststate & $1.17 \pm 0.01$ & $23 \pm 19$ & $1.8 \pm 0.0$ & $33 \pm 23$ & $2.45 \pm 0.01$ & $28 \pm 22$ \\ 
prefix\_agg & $1.56 \pm 0.18$ & $31 \pm 26$ & $1.55 \pm 0.23$ & $32 \pm 23$ & $2.12 \pm 0.02$ & $26 \pm 21$ \\ 
\bottomrule
\end{tabular}
\end{adjustbox}
\end{center}
\end{table} 

\begin{table}[hbtp]
\caption{Execution times for {\bfseries SVM} (\replaced{continued}{2})}
\label{table:performance_results_svm_2}
\begin{center}
\begin{adjustbox}{max width=\textwidth}
\begin{tabular}{@{}lcccccccc@{}}
\toprule
 & \multicolumn{2}{c}{{\bfseries sepsis\_1}} & \multicolumn{2}{c}{{\bfseries sepsis\_2}} & \multicolumn{2}{c}{{\bfseries sepsis\_3}} \\ \cmidrule(lr){2-3} \cmidrule(lr){4-5} \cmidrule(lr){6-7}
method & offline\_total (s) & online\_avg (ms) & offline\_total (s) & online\_avg (ms) & offline\_total (s) & online\_avg (ms) \\ \midrule
single\_laststate & $62.48 \pm 0.54$ & $28 \pm 32$ & $6.46 \pm 0.02$ & $34 \pm 34$ & $12.88 \pm 0.09$ & $30 \pm 32$ \\ 
single\_agg & $15.52 \pm 0.22$ & $29 \pm 33$ & $5.81 \pm 0.06$ & $36 \pm 36$ & $14.88 \pm 0.08$ & $31 \pm 34$ \\ 
knn\_laststate & $3.14 \pm 0.07$ & $29 \pm 33$ & $1.06 \pm 0.01$ & $38 \pm 37$ & $1.91 \pm 0.06$ & $31 \pm 33$ \\ 
knn\_agg & $2.72 \pm 0.12$ & $30 \pm 34$ & $1.02 \pm 0.02$ & $37 \pm 37$ & $1.82 \pm 0.03$ & $33 \pm 35$ \\ 
state\_laststate & $5.14 \pm 0.09$ & $29 \pm 33$ & $2.08 \pm 0.02$ & $35 \pm 36$ & $4.23 \pm 0.02$ & $31 \pm 34$ \\ 
state\_agg & $9.5 \pm 0.16$ & $33 \pm 37$ & $2.61 \pm 0.03$ & $40 \pm 40$ & $5.08 \pm 0.02$ & $35 \pm 38$ \\ 
cluster\_laststate & $10.03 \pm 0.09$ & $\bm{26 \pm 32}$ & $6.02 \pm 0.11$ & $34 \pm 34$ & $7.7 \pm 0.07$ & $\bm{29 \pm 32}$ \\ 
cluster\_agg & $10.37 \pm 0.09$ & $28 \pm 34$ & $6.84 \pm 0.03$ & $\bm{33 \pm 35}$ & $7.95 \pm 0.06$ & $31 \pm 34$ \\ 
prefix\_index & $10.45 \pm 0.03$ & $36 \pm 24$ & $3.16 \pm 0.01$ & $41 \pm 26$ & $6.67 \pm 0.05$ & $38 \pm 25$ \\ 
prefix\_laststate & $6.63 \pm 0.02$ & $28 \pm 31$ & $2.49 \pm 0.03$ & $34 \pm 33$ & $4.78 \pm 0.04$ & $30 \pm 32$ \\ 
prefix\_agg & $6.61 \pm 0.09$ & $30 \pm 33$ & $2.85 \pm 0.04$ & $36 \pm 35$ & $4.96 \pm 0.02$ & $31 \pm 34$ \\ 
\bottomrule
 & \multicolumn{2}{c}{{\bfseries bpic2012\_1}} & \multicolumn{2}{c}{{\bfseries bpic2012\_2}} & \multicolumn{2}{c}{{\bfseries bpic2012\_3}} \\ \cmidrule(lr){2-3} \cmidrule(lr){4-5} \cmidrule(lr){6-7}
method & offline\_total (s) & online\_avg (ms) & offline\_total (s) & online\_avg (ms) & offline\_total (s) & online\_avg (ms) \\ \midrule
single\_laststate & $12067.81 \pm 399.76$ & $10 \pm 14$ & $14557.12 \pm 780.58$ & $8 \pm 11$ & $2746.74 \pm 133.03$ & $9 \pm 13$ \\ 
single\_agg & $3143.14 \pm 24.75$ & $10 \pm 16$ & $1336.26 \pm 5.35$ & $9 \pm 14$ & $3940.96 \pm 508.12$ & $10 \pm 15$ \\ 
knn\_laststate & $28.92 \pm 0.47$ & $279 \pm 299$ & $31.47 \pm 0.55$ & $254 \pm 274$ & $30.7 \pm 2.47$ & $174 \pm 188$ \\ 
knn\_agg & $33.21 \pm 1.1$ & $258 \pm 277$ & $29.0 \pm 0.93$ & $297 \pm 319$ & $31.11 \pm 0.58$ & $171 \pm 189$ \\ 
state\_laststate & $108.84 \pm 0.29$ & $8 \pm 10$ & $156.77 \pm 2.77$ & $8 \pm 10$ & $128.66 \pm 0.22$ & $8 \pm 10$ \\ 
state\_agg & $168.48 \pm 1.15$ & $10 \pm 12$ & $88.35 \pm 1.29$ & $10 \pm 12$ & $191.19 \pm 3.22$ & $10 \pm 13$ \\ 
cluster\_laststate & $414.04 \pm 3.83$ & $9 \pm 14$ & $139.3 \pm 0.17$ & $8 \pm 13$ & $181.63 \pm 2.76$ & $8 \pm 12$ \\ 
cluster\_agg & $186.63 \pm 2.21$ & $8 \pm 9$ & $2409.21 \pm 27.71$ & $11 \pm 18$ & $273.16 \pm 6.91$ & $10 \pm 15$ \\ 
prefix\_index & $538.77 \pm 6.5$ & $37 \pm 12$ & $420.91 \pm 13.0$ & $34 \pm 11$ & $344.04 \pm 8.18$ & $34 \pm 11$ \\ 
prefix\_laststate & $77.87 \pm 0.94$ & $\bm{4 \pm 3}$ & $211.0 \pm 9.02$ & $\bm{4 \pm 3}$ & $100.44 \pm 2.24$ & $\bm{5 \pm 4}$ \\ 
prefix\_agg & $100.43 \pm 0.25$ & $5 \pm 5$ & $288.14 \pm 4.56$ & $5 \pm 5$ & $66.23 \pm 0.75$ & $\bm{5 \pm 4}$ \\ 
\bottomrule
 & \multicolumn{2}{c}{{\bfseries bpic2017\_1}} & \multicolumn{2}{c}{{\bfseries bpic2017\_2}} & \multicolumn{2}{c}{{\bfseries bpic2017\_3}} \\ \cmidrule(lr){2-3} \cmidrule(lr){4-5} \cmidrule(lr){6-7}
method & offline\_total (s) & online\_avg (ms) & offline\_total (s) & online\_avg (ms) & offline\_total (s) & online\_avg (ms) \\ \midrule
single\_laststate & $86469.6 \pm 1696.97$ & $38 \pm 46$ & $91457.8 \pm 1923.76$ & $42 \pm 56$ & $64770.83 \pm 1271.13$ & $32 \pm 39$ \\ 
single\_agg & $89619.4 \pm 1736.87$ & $40 \pm 47$ & $23877.21 \pm 468.59$ & $30 \pm 38$ & $68407.03 \pm 1323.42$ & $35 \pm 42$ \\ 
knn\_laststate & $117.67 \pm 2.31$ & $118 \pm 108$ & $133.43 \pm 2.62$ & $1515 \pm 1424$ & $119.49 \pm 2.35$ & $1375 \pm 1292$ \\ 
knn\_agg & $122.36 \pm 2.4$ & $1408 \pm 1332$ & $118.44 \pm 2.32$ & $1412 \pm 1329$ & $139.45 \pm 2.74$ & $1587 \pm 1497$ \\ 
state\_laststate & $18343.13 \pm 359.98$ & $23 \pm 26$ & $1905.91 \pm 37.4$ & $20 \pm 24$ & $37822.2 \pm 742.26$ & $17 \pm 23$ \\
state\_agg & $10257.06 \pm 201.29$ & $25 \pm 28$ & $58203.39 \pm 1142.24$ & $23 \pm 26$ & $10121.8 \pm 198.64$ & $24 \pm 26$ \\ 
cluster\_laststate & $8744.81 \pm 171.62$ & $17 \pm 22$ & $7006.73 \pm 137.51$ & $17 \pm 21$ & $3026.05 \pm 59.39$ & $18 \pm 22$ \\ 
cluster\_agg & $74691.13 \pm 1465.81$ & $22 \pm 27$ & $8903.74 \pm 174.74$ & $23 \pm 28$ & $5264.25 \pm 103.31$ & $18 \pm 21$ \\ 
prefix\_index & $17933.49 \pm 351.94$ & $76 \pm 9$ & $39670.6 \pm 778.54$ & $87 \pm 10$ & $24417.21 \pm 479.19$ & $89 \pm 11$ \\ 
prefix\_laststate & $16596.88 \pm 1364.46$ & $\bm{16 \pm 17}$ & $3553.0 \pm 3.84$ & $\bm{14 \pm 15}$ & $7667.05 \pm 53.16$ & $\bm{15 \pm 16}$ \\ 
prefix\_agg & $21941.34 \pm 37.13$ & $\bm{16 \pm 18}$ & $5395.36 \pm 587.31$ & $16 \pm 18$ & $4509.12 \pm 372.24$ & $18 \pm 20$ \\ 
\bottomrule
 & \multicolumn{2}{c}{{\bfseries traffic}} & \multicolumn{2}{c}{{\bfseries hospital\_1}} & \multicolumn{2}{c}{{\bfseries hospital\_2}} \\ \cmidrule(lr){2-3} \cmidrule(lr){4-5} \cmidrule(lr){6-7}
method & offline\_total (s) & online\_avg (ms) & offline\_total (s) & online\_avg (ms) & offline\_total (s) & online\_avg (ms) \\ \midrule
single\_laststate & $52218.27 \pm 1024.78$ & $103 \pm 57$ & $163070.85 \pm 3200.27$ & $487 \pm 319$ & $100195.84 \pm 1966.34$ & $482 \pm 311$ \\ 
single\_agg & $58867.16 \pm 1155.27$ & $114 \pm 62$ & $65398.88 \pm 1283.45$ & $436 \pm 281$ & $291765.77 \pm 5725.9$ & $545 \pm 353$  \\ 
knn\_laststate & $394.55 \pm 7.74$ & $69 \pm 40$ & $109.75 \pm 1.32$ & $81 \pm 55$ & $312.82 \pm 6.14$ & $\bm{62 \pm 49}$ \\ 
knn\_agg & $482.25 \pm 9.46$ & $96 \pm 54$ & $102.69 \pm 2.02$ & $\bm{42 \pm 23}$ & $321.04 \pm 6.3$ & $85 \pm 68$ \\ 
state\_laststate & $31579.01 \pm 1040.68$ & $78 \pm 48$ & $33549.22 \pm 658.4$ & $320 \pm 268$ & $110178.73 \pm 2162.26$ & $365 \pm 307$ \\ 
state\_agg & $71510.15 \pm 1403.39$ & $87 \pm 53$ & $31458.64 \pm 617.38$ & $359 \pm 222$ & $18274.1 \pm 358.63$ & $385 \pm 236$ \\ 
cluster\_laststate & $7466.95 \pm 149.95$ & $72 \pm 46$ & $23090.77 \pm 453.16$ & $313 \pm 292$ & $19699.97 \pm 386.61$ & $285 \pm 255$ \\ 
cluster\_agg & $97876.03 \pm 1920.82$ & $\bm{45 \pm 40}$ & $46246.86 \pm 907.59$ & $324 \pm 239$ & $22553.25 \pm 442.61$ & $325 \pm 242$ \\ 
prefix\_index & $28902.06 \pm 145.18$ & $125 \pm 20$ & $53322.05 \pm 1046.45$ & $1093 \pm 132$ & $82296.52 \pm 1615.07$ & $1013 \pm 214$ \\ 
prefix\_laststate & $12316.56 \pm 14.1$ & $72 \pm 43$ & $39203.83 \pm 1030.35$ & $360 \pm 255$ & $76378.26 \pm 118.46$ & $344 \pm 244$ \\ 
prefix\_agg & $24735.14 \pm 425.56$ & $75 \pm 45$ & $52526.59 \pm 857.39$ & $393 \pm 234$ & $57426.99 \pm 22.74$ & $364 \pm 215$ \\ 
\bottomrule
\end{tabular}
\end{adjustbox}
\end{center}
\end{table}

\begin{figure}[hbtp]
\centering
\includegraphics[width=1\textwidth]{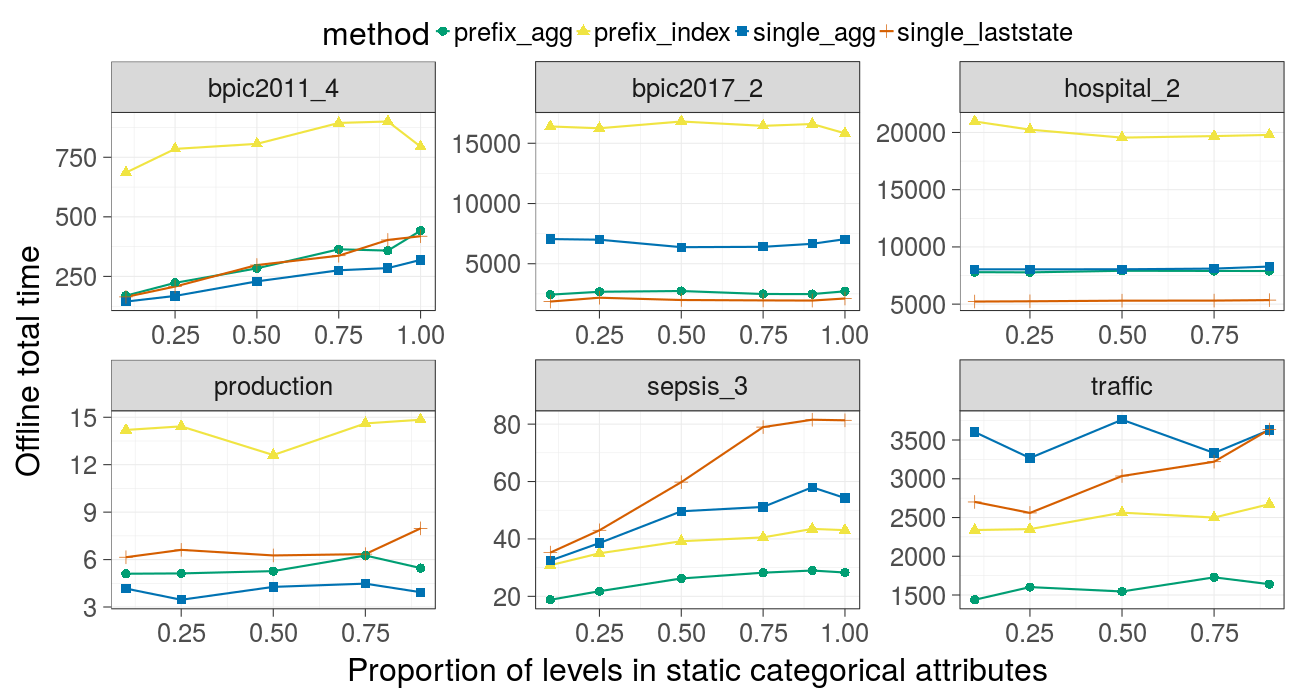}
\caption{Offline times across different filtering proportions of {\bfseries static} categorical attribute levels (XGBoost)}
\label{fig:offline_time_static_levels_xgboost}
\end{figure}

\begin{figure}[hbtp]
\centering
\includegraphics[width=1\textwidth]{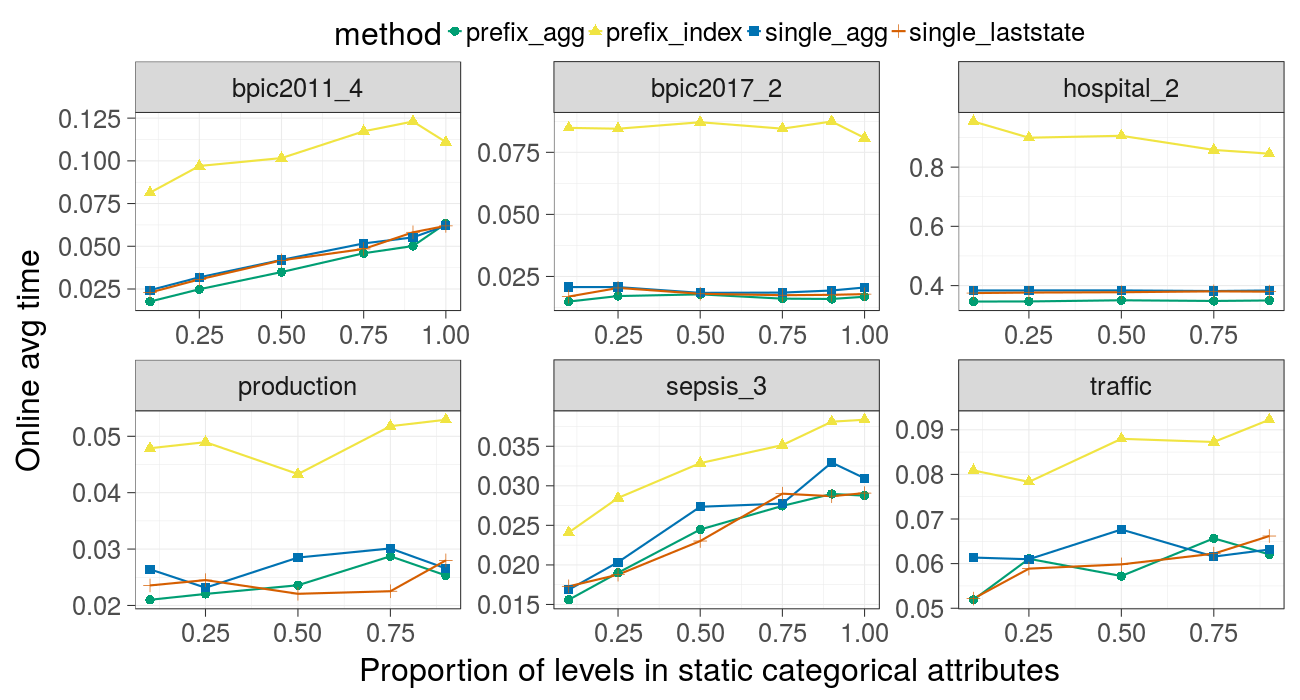}
\caption{Online times across different filtering proportions of {\bfseries static}  categorical attribute levels (XGBoost)}
\label{fig:online_time_static_levels_xgboost}
\end{figure}

\begin{figure}[hbtp]
\centering
\includegraphics[width=1\textwidth]{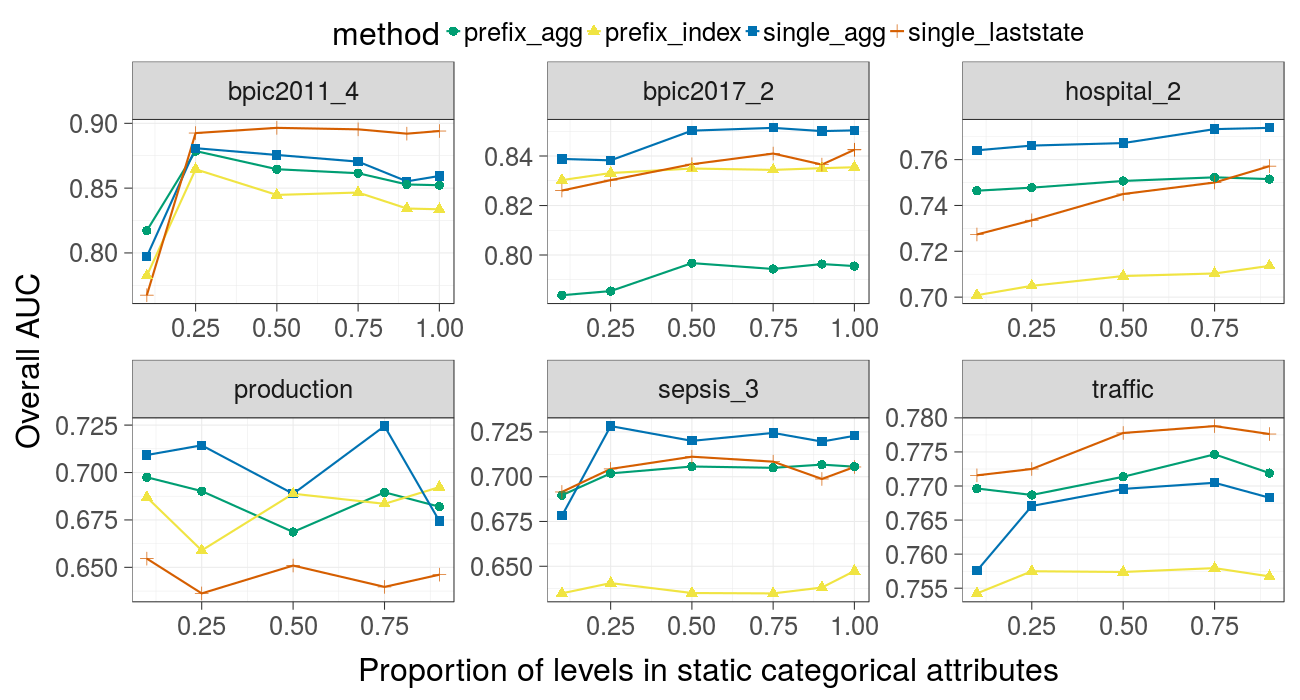}
\caption{AUC across different filtering proportions of {\bfseries static} categorical attribute levels (XGBoost)}
\label{fig:aucs_static_levels_xgboost}
\end{figure}

\end{document}